\documentclass[journal]{IEEEtran}
\usepackage{amsmath}
\usepackage{amssymb}
\usepackage{amsfonts}
\usepackage{amsthm}
\usepackage{mathtools}
\usepackage{balance}
\usepackage{bm}
\usepackage{subfloat}
\usepackage{subfig}
\usepackage{colortbl}
\usepackage{lipsum}      
\usepackage{changepage}
\usepackage{caption}
\usepackage{color}
\usepackage{pgf,tikz}
\usepackage{balance}
\usepackage{mathtools}
\usepackage{bbm}
\usepackage{array}
\usepackage{relsize}
\usepackage{cite}
\usepackage{amsthm}
\usepackage{verbatim}
\usepackage{epstopdf}
\usepackage{relsize}
\usepackage{array}
\usepackage{url}
\usepackage{stfloats}
\usepackage[hidelinks]{hyperref}
\usepackage{url}
\usepackage{stfloats}
\usepackage[hidelinks]{hyperref}
\usepackage{url}
\usepackage[linesnumbered,ruled]{algorithm2e}
\usepackage{algpseudocode}
\newtheorem{theorem}{Theorem}
\newtheorem{corollary}{Corollary}
\newtheorem{proposition}{Proposition}
\newtheorem{lemma}{Lemma}
\newtheorem{remark}{Remark}
\newtheorem{definition}{Definition}

\newcommand{\eqdef}{\mathrel{\mathop:}=}
\newcommand*{\QEDclosed}{\hfill\ensuremath{\blacksquare}}

\newcommand{\Cross}{\mathbin{\tikz [x=1.0ex,y=1.0ex,line width=.1ex] \draw (0,0) -- (1,1) (0,1) -- (1,0);}}%
\usepackage{pdfpages}

\begin{document}

\title{Error Bounds for a Matrix-Vector Product Approximation with Deep ReLU Neural Networks }

\author{Tilahun~M. Getu,~\IEEEmembership{Member,~IEEE}   
  \thanks{\IEEEcompsocthanksitem T. M. Getu is with the National Institute of Standards and Technology (NIST), 100 Bureau Drive, Gaithersburg, MD 20899, USA and also with the \'Ecole de Technologie Sup\'erieure (\'ETS), Montr\'eal, QC H3C 1K3, Canada (e-mail: tilahun.getu@nist.gov).}
}

\maketitle

\begin{abstract}
Among the several paradigms of artificial intelligence (AI)/machine learning (ML), a remarkably successful paradigm is deep learning. Deep learning's phenomenal success has been hoped to be interpreted via fundamental research on the theory of deep learning. Accordingly, applied research on deep learning has spurred the theory of deep learning oriented depth and breadth of developments. Inspired by such developments, we pose these fundamental questions: can we accurately approximate an arbitrary matrix-vector product using deep rectified linear unit (ReLU) feedforward neural networks (FNNs)? If so, can we bound the resulting approximation error? In light of these questions, we derive error bounds in Lebesgue and Sobolev norms that comprise our developed deep approximation theory. Guided by this theory, we have successfully trained deep ReLU FNNs whose test results justify our developed theory. The developed theory is also applicable for guiding and easing the training of teacher deep ReLU FNNs in view of the emerging teacher-student AI/ML paradigms that are essential for solving several AI/ML problems in wireless communications and signal processing; network science and graph signal processing; and network neuroscience and brain physics.            
\end{abstract}

\IEEEpeerreviewmaketitle

\section{Introduction}
\label{sec: ReLU_representation_introduction}
\subsection{Related Works}
\label{subsec: intro_related_works}
Understanding natural and artificial intelligence have been mankind's prominent inquiry that has led to compelling advancements toward various promising technologies. The promising technologies continually investigated by the neuroscience and artificial intelligence (AI)/machine learning (ML) research communities include brain emulation, human-machine interfaces, and AI \cite[Ch. 2]{Storm_Sup_Intel_14}. In the quest for AI \cite{Nilsson_quest_for_AI'09}, feedforward neural networks (FNNs) have been workhorse models since Rosenblatt’s perceptron of the 1950s \cite[Ch. 1]{Haykin_NNs_09}. Rosenblatt’s perceptron was the first algorithmically described neural network \cite[Ch. 1]{Haykin_NNs_09} and it inspired multilayer perceptrons. Multilayer perceptrons are FNNs with one or more hidden layers and trained using the back-propagation algorithm \cite{Rumelhart_BackPro_86}. Following the back-propagation algorithm's discovery in around 1986 \cite{Rumelhart_BackPro_86}, research on the representation and approximation power of (shallow) FNNs were intensified in the late 1980s and 1990s \cite{Pinkus99approximationtheory}. During this era, several works established that FNNs with one hidden layer are universal approximators \cite{Cybenko_Approx_Superpos_89,Hornik_Univ_Approx_89,Funahashi_NNs_89,Barron_ANN_Approx_94,Mhaskar_NN_Opt_Approx_96,Pinkus99approximationtheory}.  

Propelled by the advancements of research on FNNs and the availability of enormous computing power as well as a deluge of data, deep learning \cite{Lecun_DL_Nature_15,IGYAC16,DL_methods_14} has recently emerged as a notably successful AI/ML paradigm. In such an AI/ML paradigm, the various implicit discriminative features of the training data are disentangled by deep representation learning \cite{Bengio_RL_13} through non-linear (also linear\footnote{Deep FNNs with linear layers can also learn discriminative features of the input data successfully. In this respect, their convergence was characterized in \cite{NIPS_Kawaguchi_16,Arora2018_Convergence} and their nonlinear learning dynamics were analyzed in \cite{Saxe2013_exact_NonL_Dynamics}.}) deep transformations. Deep transformations are at the heart of the exponential expressivity power of deep networks \cite{Rolnick_ICLR_18,NIPS_Expo_Expre_2016,Lin_2017,PMLR-v49-Eldan_16,Mhaskar_Deep_vs_Shallow_16,Poggio_IJAC_17} that have become the norm of AI/ML research, especially after the superhuman performance of \textit{AlexNet} \cite{AlexNet_12}. AlexNet has sparked waves of research on deep learning (and deep reinforcement learning) that are now applied in numerous research fields as diverse as clustering, information retrieval, dimensionality reduction, and natural language processing \cite{DL_methods_14,IGYAC16}; computer vision, speech recognition, image processing, and object recognition \cite{IGYAC16,Lecun_DL_Nature_15,LWXPJLM18}; wireless communications, wireless networking, and signal processing \cite{QFQH_18,Chen_Saaf_ANN_ML_19,DRL_Applcs_19,Huang_DL_5G'20}; and gaming, finance, energy, and healthcare \cite{mnih-DQN-2015,Li2018_DRL}.         
 
Along the lines of the state-of-the-art deep learning techniques' striking performance demonstrated for various learning problems, the most important fundamental question is interpreting the \textquotedblleft black box of AI'' \cite{Castelvecchi_Black_Box_16}. This undertaking has been motivated by the theories on (shallow) FNNs, the rapid advancements and broad adoption of deep learning, and the need to devise an interpretable/explainable AI. Toward interpretable/explainable AI, the theory of deep learning (ToDL) has recently emerged as a promising and vibrant research field. This research field, specifically, attempts to demystify the various hidden transformations of deep architectures in order to provide a fundamental theoretical guarantee and explanation on the success of the existing deep networks. Deep networks under ToDL studies include FNNs, convolutional neural networks (CNNs) \cite{Gu_CNN_Advances_2017,SJ_3DCNN_13}, recurrent neural networks (RNNs) and their popular variants named long short-term memory (LSTM) \cite{conf/icml/JozefowiczZS15,Greff_LSTM_17}, autoencoders (AEs) \cite{Tschannen_2018_AE_Advance,Bank2020_AEs,Yu2019_understanding_AEs}, generative adversarial networks (GANs) \cite{Wang2019GenerativeAN,Creswell_GANs_18,Hu2017unifying_GMs}, \textit{ResNet} \cite{He_ResNet_CVPR_16}, and \textit{DenseNet} \cite{Huang_DenseNet_CVPR_17,Huang_DenseNet_Jour_19}. These deep networks' ToDL exposition has matured into three fundamental research themes: approximation (representation)\footnote{The mathematics and AI/ML research communities pursue research, respectively, on deep approximation and deep representation which are the same AI/ML problems.}, optimization, and generalization \cite{Sun_SPM_global_landscape'20,Poggio_Theo_Issues_Dnets_2020}. 

\subsection{Motivation and Context}
\label{subsec: intro_motivation_and_context}
To develop theories on approximation \cite{Mhaskar_Deep_vs_Shallow_16}, optimization \cite{Sun2019optimization,Bartlett2018_Representing}, and generalization \cite{Kawaguchi_Generalization_DL_19,Jakubovitz_generalization'19} for the state-of-the-art deep networks solving either classification \cite{Giryes_TSP_16,Giryes_TSP_Correction_20} or regression AI/ML problems \cite{Poggio_Theo_Issues_Dnets_2020,Fan2019_DL_Sel_Overview}, considerable ToDL advancements have been made via numerous frameworks. These frameworks include mean field theory \cite{Mei_PNAS_NN_landscape_18,NIPS_Expo_Expre_2016,Schoenholz_Deep_Info_Propag_19}, random matrix theory \cite{PMLR-v70-pennington17a,Becker_Geo_Ener_LandS_2020}, tensor factorization \cite{Haeffele2015_Global_Opt,Cohen_Tens_Anal_16}, optimization theory \cite{Convex_NNs_NIPS2006,Fbach_Convex_NNs_14,Feizi2017_Porcupine_17,GADT_Huang_15}, kernel learning \cite{jacot2018neural,arora2019exact,Belkin2018_KL}, linear algebra \cite{Georgiev2018_Lin_Algeb,Tsapanos_Paraboloid_NN_19}, spline theory \cite{PMLR_Spline_Theory_18,Mad_Max_DL_18}, theoretical neuroscience \cite{Richards_DL_FramW_19,Marblestone_DL_NS_16}, high-dimensional probability \cite{Zou2018_SGD}, high-dimensional statistics \cite{allenzhu2018convergence},  manifold theory \cite{NIPS_Expo_Expre_2016,Zhu_LDMNet_18}, Fourier analysis \cite{Xu2019_Frequency_Principle}, and scattering networks \cite{Wiatowski_EP_18,Wiatowski_Mat_Theo_DCNN_18}. 
     
Compared with shallow FNNs, several ToDL works have established that deep FNNs have the power of exponential expressivity \cite{Rolnick_ICLR_18,NIPS_Expo_Expre_2016,Lin_2017,PMLR-v49-Eldan_16,Mhaskar_Deep_vs_Shallow_16,Poggio_IJAC_17}. The exponential expressivity of deep networks has stimulated lines of ToDL research regarding the universal approximation capability of ResNet \cite{Lin2018_ResNet_2Approx} and slim (and sparse) networks \cite{Fan2018_Slim_20}, the quasi-equivalence of depth and width \cite{Fan_Dep_Wid_duality_20}, the width-efficiency of rectified linear unit (ReLU) FNNs \cite{Zhu_view_from_width_2017}, the improving effect of depth and width \cite{Kenji_Effect_Dep_Width_19}, and \textit{un-rectifying} signal representation \cite{Hwang_unrectifying_net_20}. Representation research -- in the context of ToDL -- overlaps with approximation research using deep neural networks and, in particular, deep ReLU FNNs. For deep ReLU FNNs approximating various functions, the authors of \cite{Hanin_2019_Univ_Approx,grohs2019deep,IGP_error_bounds_19,PP_optimal_approx_18,HB_DNN_19} developed approximation theories. Although these approximation theories are interesting in their own rights and have the potential to inspire much more research, they focus on a scalar input and a scalar output. In the age of big data, nevertheless, ToDL\footnote{ToDL is also being developed under the names \textit{learning theory} and \textit{the mathematics of data science}. For all these research fields, no distinction in scope is delineated by the AI/ML and mathematics research communities.} research needs to account for deep learning in high dimension. Deep learning in high dimension through the lenses of high-dimensional probability \cite{vershynin_2018}, high-dimensional statistics \cite{wainwright_2019}, and empirical process theory \cite{Ben_Process_Theory_2018} which are crucial tools of ToDL, high-dimensional bounds afford useful insights on approximation with deep ReLU FNNs fed with input data in high dimension. High dimension considered fittingly with its timeliness, the works in \cite{Kutyniok2019_theoretical_Anals_PDEs} and \cite{Schwab_19} presented FNN-based approximations applicable for parametric partial differential equations.

In consideration of high dimension, high-dimensional signal processing based wireless communications technologies such as massive multiple-input multiple-output (MIMO) and massively parallel IoT (Internet of things) are on the verge of being pervasive in 5G (fifth generation) and beyond \cite{M-MIMO_is_reality'19}. Toward 5G and beyond, the IEEE communications and signal processing research communities have swiftly adopted deep learning for solving numerous classification and regression AI/ML problems in wireless communications, wireless networking, and signal processing \cite{QFQH_18,Chen_Saaf_ANN_ML_19,Huang_DL_5G'20}. These research fields' future strides in deep learning will benefit from fundamental ToDL interpretations. ToDL interpretations, specifically, inspire and inform intelligently adaptive AI/ML-based wireless communications algorithms that employ efficient architectures predicted by ToDL. From ToDL's fundamental vantage point, an important gap is the performance quantification of the proposed deep learning algorithms fed with input signals modeled by matrix-vector products. For matrix-vector products' approximation with deep ReLU FNNs, consequently, we develop a deep approximation theory useful for guiding the training of teacher deep ReLU FNNs in view of the emerging teacher-student AI/ML paradigms \cite{Hinton_KD_14,Romero_FitNets'15,Saputra2019_KD_Regressor,Vapnik_LUPI_JMLR'15,Phuong_Under_KD_19} that are essential for solving numerous AI/ML problems.

\subsection{Contributions}
\label{subsec: DSFC_contib}
The contributions of this paper are threefold: 
\begin{enumerate}
\item For a matrix-vector product approximation with a deep ReLU FNN, we develop a deep approximation theory by deriving error bounds in Lebesgue and Sobolev norms. 
\item We present computer experiments whose empirical results corroborate our deep approximation theory. 
\item Implying broader relevance, we discuss the applications of our developed theory in the context of teacher-student AI/ML paradigms solving various AI/ML problems in wireless communications and signal processing; network science and graph signal processing; and network neuroscience and brain physics \cite{Bassett_Net_NS_17,Brin_Physics_Lynn_2019}.	
\end{enumerate}
These contributions bear significance in the sense that they inspire numerous works on ToDL geared toward deep representation and the performance quantification of AI/ML algorithms applied to the aforementioned research fields.

The remainder of this paper is organized as follows. Sec. \ref{sec: prelims} highlights the preliminaries. Sec. \ref{sec: matrix_vector_product_approximation} details our problem motivation and problem formulation. Sec. \ref{sec: dev_approx_theory} documents the developed deep approximation theory. Sec. \ref{sec: comp_experiments} reports our computer experiments. Sec. \ref{sec: applications} presents applications. Finally, Sec. \ref{sec: Conc_rem_and_res_outlook} draws concluding remarks and research outlook. 
 
\textit{Notation:} italic and bold lowercase(uppercase) letters denote scalars and vectors(matrices), respectively. Bold calligraphic letters such as $\bm{\mathcal{NN}}$ and $\bm{\mathcal{W}}$ indicate FNNs and a Sobolev space, respectively. $\mathbb{C}^{n}(\mathbb{R}^{n})$ and $\mathbb{C}^{m\times n}(\mathbb{R}^{m\times n})$ represent the sets of $n$-dimensional vectors of complex(real) numbers and $m\times n$ complex(real) matrices, respectively. $\mathbb{R}_{+}$, $\mathbb{N}$, and $\mathbb{N}_0^n(\mathbb{N}_0)$ denote the sets of positive real numbers, natural numbers, and ($n$-dimensional)natural numbers including zero, respectively. $\mathbb{N}_{\geq k}\eqdef \{k, k+1, \ldots \}$ denotes the sets of natural numbers greater than or equal to $k\in \mathbb{N}_0$. For $n\geq2$, $[n]\eqdef\{1, \ldots, n\}$. $\sim$, $\eqdef$, $|\cdot|$, $(\cdot)^{T}$,  $\bm{I}_n$, $\bm{0}_{m\times n}$, and $\bm{0}$ stand for distributed as, equal by definition, cardinality, transpose, an $n\times n$ identity matrix, an $m\times n$ zero matrix, and a zero vector (matrix) whose dimension will be clear in context, respectively. $\textnormal{max}$, $\textnormal{vec}(\cdot)$, $\textnormal{diag}(\cdot)$, $\textnormal{Re}\{\cdot\}$, $\textnormal{Im}\{\cdot\}$, $\mathbb{P}(\cdot)$, and $\mathcal{N}(\bm{0}, \sigma^2\bm{I}_n)$$\big(\mathcal{CN}(\bm{0}, \sigma^2\bm{I}_n)\big)$ express maximum, vectorization, a (block) diagonal matrix, a real part, an imaginary part, probability, and a normal (complex normal) distribution with a zero mean and covariance $\sigma^2\bm{I}_n$, respectively. For $\bm{W}\in\mathbb{R}^{m\times n}$ and $\bm{x}\in\mathbb{R}^{n}$, $w_{i,j}$ and $x_i$ represent the $(i,j)$-th element of $\bm{W}$ and the $i$-th entry of $\bm{x}$, respectively. Per the MATLAB$^{\textregistered}$ syntax, $\bm{W}(i,:)\big(\bm{W}(:,j)\big)$ denotes the $i$-th row($j$-th column) of $\bm{W}$. $\|\bm{W}\|_{\ell_0}\eqdef |{(i, j)}: w_{i,j}\neq 0 |$; $\|\bm{x}\|_\infty\eqdef \max_{i=1, \ldots, n} |x_i|$; and $\|\bm{W}\|_\infty\eqdef \max_{i, j} |w_{i,j}|$. The horizontal concatenation of $n$ vectors(matrices) is denoted as $[\bm{w}_1(\bm{W}_1), \ldots, \bm{w}_n(\bm{W}_n)]$. A sequence of $K$ matrix-vector tuples is expressed as $\big[ [\bm{W}_1, \bm{b}_1], \ldots, [\bm{W}_K, \bm{b}_K] \big]$. For $A\subset\mathbb{R}^d$ (the standard topology), $\overline{A}$ denotes the closure of $A$. If $A, B\subset\mathbb{R}^d$, we write $A\subset\subset B$ provided that $\overline{A}$ is compact in $B$. For $\bm{\alpha}\in\mathbb{N}_0^d$, we let $|\bm{\alpha}|\eqdef \sum_{i=1}^d \alpha_i$. With respect to (w.r.t.) $L^p$ (Lebesgue) spaces and $f(x):  \mathbb{R}^d\to\mathbb{R}$, $\|f\|_{L^\infty(\Omega)}\eqdef \textnormal{inf}\big\{ C\geq 0: |f(x)|\leq C, \hspace{0.5mm} \forall x\in\Omega \big| \Omega\subset\mathbb{R}^d \big\}$. For a function $f: X\to \mathbb{R}$, supp $f$ and ran $f$ stand for the support of $f$ and the range of $f$, respectively. For functions $f: X\to Y$ and $g: Y\to Z$, their composition is denoted as $g\circ f: X\to Z$. A function $f: \Omega\to \mathbb{R}^m$ is an $L$-Lipschitz continuous whenever $|f(x)-f(y)|\leq L|x-y|$ for a constant $L>0$ and $\Omega\subset\mathbb{R}^d$. 

\section{Preliminaries}
\label{sec: prelims}
\subsection{The Mathematics of FNNs}
\label{subsec: FNNs_math}
Based on \cite[Definition II.1]{grohs2019deep} and \cite[Definition 1.1]{HB_DNN_19}, we state the following definition on the mathematics of FNNs.\footnote{To stick to our notation, we assume a vector of FNN inputs and outputs.} 
\begin{definition}
	\label{def: NN_definition}
	Suppose $K, N_0, N_1, N_2,\ldots, N_K\in \mathbb{N}$, $K\geq2$, and $\bm{x}\in \mathbb{R}^{N_0}$. A map $\bm{\Phi}: \mathbb{R}^{N_0}\to\mathbb{R}^{N_K}$ characterized as 
	\begin{equation}
	\label{NN_def}
	\bm{\Phi(x)}\eqdef \bm{A}_K(\rho(\bm{A}_{K-1}(\rho(\ldots \rho(\bm{A}_1(\bm{x})))))),  
	\end{equation}
	with affine linear maps $\bm{A}_k: \mathbb{R}^{N_{k-1}}\to\mathbb{R}^{N_k}$ and an element-wise ReLU activation function $\rho(\bm{x})\eqdef[\rho(x_1), \ldots, \rho(x_{N_0})]^T$ -- $\rho(a)\eqdef\textnormal{max}(a, 0)$ -- is called a ReLU FNN. The ReLU FNN's depth is equal to $K$ and denoted as $\mathcal{L}(\bm{\Phi}) \eqdef K$, $N_0$ is the dimension of the $0$-th layer (input layer), $N_k$ is the dimension of the $k$-th layer, and $N_K$ is the dimension of the output layer. The affine linear map corresponding to the $k$-th layer is defined via $\bm{A}_k(\bm{x})=\bm{W}_k\bm{x}+\bm{b}_k$ with $\bm{W}_k\in \mathbb{R}^{N_k\times N_{k-1}}$ and $\bm{b}_k\in\mathbb{R}^{N_k}$. W.r.t. $\bm{W}_k$ and $\bm{b}_k$, $w_{k,i,j}$ denotes the weight associated with the edge between the $j$-th node of the $(k-1)$-th layer and the $i$-th node of the $k$-th layer and $b_{k,i}$ is the weight associated with the $i$-th node of the $k$-th layer. For the ReLU FNN described by (\ref{NN_def}), the total number of neurons, the network connectivity, the maximum width, and the maximum absolute value of the weights are denoted and defined as $\mathcal{N}(\bm{\Phi})\eqdef \sum_{k=0}^K N_k$, $\mathcal{M}(\bm{\Phi})\eqdef\sum_{k=1}^{K} (\|\bm{W}_k\|_{\ell_0}+\|\bm{b}_k\|_{\ell_0})$, $\mathcal{W}(\bm{\Phi})\eqdef \max_{k=0, \ldots, K} N_k$, and $\mathcal{B}(\bm{\Phi})\eqdef \max_{k\in [K] } \big\{ \|\bm{W}_k\|_{\infty}, \|\bm{b}_k\|_{\infty}\big\}$, respectively. The class of FNNs $\bm{\Phi}: \mathbb{R}^{N_0}\to \mathbb{R}^{N_K}$ with no more than $K$ layers, connectivity no more than $M$, input dimension $N_0$, output dimension $N_K$, and activation function $\rho$ is denoted as $\bm{\mathcal{NN}}_{K, M, \rho}^{N_0, N_K}$. To this end, $\bm{\mathcal{NN}}_{\infty, M, \rho}^{N_0, N_K}\eqdef   \bigcup_{K\in\mathbb{N}}   \bm{\mathcal{NN}}_{K, M, \rho}^{N_0, N_K}$, $\bm{\mathcal{NN}}_{K, \infty, \rho}^{ N_0, N_K}\eqdef \bigcup_{M\in\mathbb{N}} \bm{\mathcal{NN}}_{K, M, \rho}^{N_0, N_K}$, and $\bm{\mathcal{NN}}_{\infty, \infty,\rho}^{ N_0, N_K}\eqdef  \bigcup_{K\in\mathbb{N}} \bm{\mathcal{NN}}_{K, \infty, \rho}^{N_0, N_K}$.             
\end{definition}

Stating Definition \ref{def: NN_definition} differently, the underneath definition follows from \cite[Definition 2.2]{IGP_error_bounds_19} and \cite[Definition 2.1]{PP_optimal_approx_18}.     
\begin{definition}
	\label{def: NN_eq_definition}
	Suppose $K, N_0, N_1, \ldots, N_K\in\mathbb{N}$ and $K\geq 2$. A FNN denoted by $\bm{\Phi}$ is a sequence of matrix-vector tuples
	\begin{equation}
	\label{NN_eq_def}
	\bm{\Phi}\eqdef\big[ [\bm{W}_1, \bm{b}_1], \ldots, [\bm{W}_K, \bm{b}_K] \big], 
	\end{equation}
	where $\bm{W}_k\in\mathbb{R}^{N_k\times N_{k-1}}$ and $\bm{b}_k\in\mathbb{R}^{N_k}$, $k\in[K]$, $\mathcal{N}(\bm{\Phi})=\sum_{k=0}^K N_k$, and $\mathcal{M}(\bm{\Phi})=\sum_{k=1}^{K} (\|\bm{W}_k\|_{\ell_0}+\|\bm{b}_k\|_{\ell_0})$. For $\bm{x}\in\mathbb{R}^{N_0}$ and $\rho(a)\eqdef\textnormal{max}(a, 0)$, a map $\bm{\Phi}: \mathbb{R}^{N_0}\to\mathbb{R}^{N_K}$ is defined as $\bm{\Phi}(\bm{x})=\bm{x}_K$, where $\bm{x}_K$ is obtained recursively as  
	\begin{subequations}
		\begin{align}
		\label{NN_recur_1}
		\bm{x}_0&\eqdef \bm{x}; \hspace{2mm} \bm{x}_k\eqdef\rho (\bm{W}_k\bm{x}_{k-1}+\bm{b}_k), \hspace{2mm} k\in [K-1]    \\
		\label{NN_recur_2}
		\bm{x}_K&\eqdef \bm{W}_K\bm{x}_{K-1}+\bm{b}_K,	 
		\end{align}
	\end{subequations}     
	where $\rho(\bm{y})=[\rho(y_1), \ldots, \rho(y_m)]^T$ for $\bm{y}=[y_1, \ldots, y_m]^T$.    
\end{definition}

Expressed via (\ref{NN_eq_def}), the concatenation of two FNNs follows.  
\begin{definition}[\textbf{Concatenation of FNNs} {\textbf{\cite[Definition 2.2]{PP_optimal_approx_18}}}]
	\label{def: NN_concatenation}
	Let $K_1, N_0^{1}, \ldots, N_{K_1}^{1}\in\mathbb{N}$ and $K_2, N_0^{2}, \ldots,  N_{K_2}^{2}, \in\mathbb{N}$. Define two FNNs as $\bm{\Phi}^1\eqdef\big[ [\bm{W}_1^{1}, \bm{b}_1^{1}], \ldots, [\bm{W}_{K_1}^{1}, \bm{b}_{K_1}^{1}] \big]$ and $\bm{\Phi}^{2}\eqdef\big[ [\bm{W}_1^{2}, \bm{b}_1^{2}], \ldots, [\bm{W}_{K_2}^{2}, \bm{b}_{K_2}^{2}] \big]$. For $N_0^{1}=N_{K_2}^{2}$, the concatenation of $\bm{\Phi}^{1}$ and $\bm{\Phi}^{2}$ is given by   
	\begin{multline}
	\label{eq_NN_concatenation}
	\bm{\Phi}^{1}\bullet\bm{\Phi}^{2}\eqdef \big[[\bm{W}_1^{2}, \bm{b}_1^{2}],\ldots, [\bm{W}_{K_2-1}^{2}, \bm{b}_{K_2-1}^{2}], [\bm{W}_1^{1}\bm{W}_{K_2}^{2}, \\  \bm{W}_1^{1}\bm{b}_{K_2}^{2}+\bm{b}_1^{1}], [\bm{W}_2^{1}, \bm{b}_2^{1}], \ldots, [\bm{W}_{K_1}^{1}, \bm{b}_{K_1}^{1}] \big],  
	\end{multline}  
	where $(\bm{\Phi}^1\bullet\bm{\Phi}^2)(\bm{x})=\bm{\Phi}^1\big(\bm{\Phi}^2(\bm{x})\big)$ and $\mathcal{L}(\bm{\Phi}^{1}\bullet\bm{\Phi}^{2})=\sum_{i=1}^2 K_i-1$.   	    
\end{definition}
For a ReLU activation $\rho(\cdot)$ and $\bm{x}\in\mathbb{R}$, $\bm{x}=\rho(\bm{x})-\rho(-\bm{x})$ \cite{grohs2019deep}. This identity is exploited to prove the following lemma.    
\begin{lemma}[\textbf{ReLU identity FNN (I-FNN)} {\textbf{\cite[Remark 2.4]{PP_optimal_approx_18}}}]
	\label{lem: NN_identity_matrix}
	For $N_0\geq 2$, $K\in\mathbb{N}_{\geq 2}$, and $\bm{x}\in\mathbb{R}^{N_0}$, one can construct an I-FNN $\bm{\Phi}_K^{\bm{I}_{N_0}}$ with $K$ layers and at most $2N_0K$ $\{-1,1\}$-valued weights such that $\bm{\Phi}_K^{\bm{I}_{N_0}}(\bm{x})=\bm{x}$. Toward this end,  
	\begin{multline}
	\label{NN_Id_1}
	\bm{\Phi}_K^{\bm{I}_{N_0}}\eqdef\big[[[\bm{I}_{N_0}, -\bm{I}_{N_0}]^T, \bm{0}], \overbrace{[\bm{I}_{2N_0}, \bm{0}],\ldots, [\bm{I}_{2N_0}, \bm{0}]}^{(K-2)\hspace{1mm}\textnormal{times}}, \\ 
	[[\bm{I}_{N_0}, -\bm{I}_{N_0}], \bm{0}]\big].
	\end{multline}
	For $K=1$, an I-FNN is constructed as $\bm{\Phi}_1^{\bm{I}_{N_0}}\eqdef \big[[\bm{I}_{N_0}, \bm{0}]\big]$.   
	
	\proof The proof is provided as supplementary material.   
\end{lemma}

Echoing the fact FNNs are universal approximators, the following proposition follows through (\ref{eq_NN_concatenation}) and (\ref{NN_Id_1}).  
\begin{proposition}[\textbf{Representation by ReLU FNNs}]
	\label{prop: proposition_1}
	Let $\bm{W}\in\mathbb{R}^{m\times n}$, $\bm{x}\in \mathbb{R}^{n}$, and $\rho(x)\eqdef\textnormal{max}(x, 0), x\in\mathbb{R}$. For an affine transformation $\bm{A}: \mathbb{R}^{n}\to \mathbb{R}^{m}$ defined as $\bm{A}=\bm{W}\bm{x}$, the following holds true.  
	\begin{enumerate}
		\item $\bm{A}$ can be represented by $\bm{\mathcal{NN}}_{2, M, \rho}^{n, m}$ with $M=2\|\bm{W}\|_{\ell_0}+2m$. 
		\item For $K>>2$, $\bm{A}$ can be represented by $\bm{\mathcal{NN}}_{K, M, \rho}^{n, m}$ with $M=2m+2(K-2)n+4\|\bm{W}\|_{\ell_0}$. 
		\item For $K>>2$, $\bm{A}$ can also be represented by $\bm{\mathcal{NN}}_{K, M, \rho}^{n, m}$ with $M=2Km+2\|\bm{W}\|_{\ell_0}$. 
	\end{enumerate}     
	\proof The proof is deferred to Appendix \ref{proof_prop: proposition_1}.
\end{proposition}
Meanwhile, the parallelization of FNNs is formalized below. 
\begin{lemma}[\textbf{Parallelization of FNNs}]
\label{lem: NN_parallelization}
Consider $n$ FNNs $\bm{\Phi}^{i}\eqdef\big[ [\bm{W}_1^i, \bm{b}_1^i], \ldots, [\bm{W}_{K}^i, \bm{b}_{K}^i] \big]$, $i\in [n]$. Let $K, n\in\mathbb{N}$ and $K, n\geq 2$. Let the parallelization of $\bm{\Phi}^{1},  \ldots, \bm{\Phi}^{n}$ be denoted and defined as $\mathcal{P}(\bm{\Phi}^{1},  \ldots, \bm{\Phi}^{n})\eqdef\big[ [\tilde{\bm{W}}_1, \tilde{\bm{b}}_1], \ldots, [\tilde{\bm{W}}_{K}, \tilde{\bm{b}}_{K}]\big]$. If $\mathcal{P}(\bm{\Phi}^{1},  \ldots, \bm{\Phi}^{n})(\bm{x})=[(\bm{\Phi}^{1}(\bm{x}))^T,  \ldots, (\bm{\Phi}^{n}(\bm{x}))^T]^T$ for $\bm{x}\in\mathbb{R}^{N_0}$, then for $\tilde{\bm{\Phi}}\eqdef [\bm{\Phi}^{1},  \ldots, \bm{\Phi}^{n}]$ and $2\leq k\leq K$
\begin{subequations}
\begin{align}
\label{Parallel_condi_1}
\hspace{-3mm}\tilde{\bm{W}}_1=&[(\bm{W}_1^1)^T, \ldots, (\bm{W}_1^n)^T]^T; \tilde{\bm{b}}_1=[(\bm{b}_1^1)^T, \ldots, (\bm{b}_1^n)^T]^T       \\
\label{Parallel_condi_2}
\hspace{-3mm}\tilde{\bm{W}}_k=&\textnormal{diag}(\bm{W}_k^1, \ldots, \bm{W}_k^n);  \tilde{\bm{b}}_k=[(\bm{b}_k^1)^T, \ldots, (\bm{b}_k^n)^T]^T,
\end{align}
\end{subequations}
where $\mathcal{L}(\mathcal{P}(\tilde{\bm{\Phi}}))=K$, $\mathcal{M}\big(\mathcal{P}(\tilde{\bm{\Phi}})\big)=\sum_{i=1}^n \mathcal{M}(\bm{\Phi}^{i})$, and $\mathcal{N}\big(\mathcal{P}(\tilde{\bm{\Phi}})\big)=\sum_{i=1}^n \mathcal{N}(\bm{\Phi}^{i})-(n-1)N_0$.       
\proof The proof is provided as supplementary material.  
\end{lemma}

Moreover, we adopt the following lemma. 
\begin{lemma}[{\textbf{Superposition and parallelization of FNNs \cite[Lemma II.7]{grohs2019deep}}}]
\label{lem: NN_superposition_parallelization}
Suppose $\rho(x)\eqdef \textnormal{max}(x, 0)$, $a_i\in\mathbb{R}$; $n, d_i, K_i, M_i\in \mathbb{N}$; $\bm{\Phi}_i\in \bm{\mathcal{NN}}^{d_i, 1}_{K_i, M_i, \rho}$, $i\in [n]$; and $d=\sum_{i=1}^{n} d_i$. Let $\bm{x}=[\bm{x}_1^T, \ldots,  \bm{x}_n^T]^T\in\mathbb{R}^d$ with $\bm{x}_i\in\mathbb{R}^{d_i}$, $i\in [n]$. There exist FNNs $\bm{\Phi}^1\in\bm{\mathcal{NN}}^{d, n}_{K, M, \rho}$ and $\bm{\Phi}^2\in\bm{\mathcal{NN}}^{d, 1}_{K, M+n, \rho}$ with $K=\max_{i} K_i$, $\mathcal{W}(\bm{\Phi}^1)=\mathcal{W}(\bm{\Phi}^2)\leq \sum_{i=1}^n \max\{2, \mathcal{W}(\bm{\Phi}_i)\}$, and $M=\sum_{i=1}^n \big(M_i+\mathcal{W}(\bm{\Phi}_i)+2(K-K_i)+1\big)$ fulfilling $\bm{\Phi}^1(\bm{x})=\big[a_1\bm{\Phi}_1(\bm{x}_1), \ldots, a_n\bm{\Phi}_n(\bm{x}_n)  \big]^T$ and $\bm{\Phi}^2(\bm{x})=\sum_{i=1}^n a_i\bm{\Phi}_i(\bm{x}_i)$. Furthermore, the weights of $\bm{\Phi}^1$ and $\bm{\Phi}^2$ comprise the weights of $\bm{\Phi}_i$, $i\in[n]$, $\{a_1, \ldots, a_n\}$, and $\pm1$'s.   
\proof The proof is deferred to Appendix \ref{proof_lem: NN_superposition_parallelization}.      	
\end{lemma}

\subsection{Function Norms and Spaces}
\label{subsec: norms_and_functional_spaces}    
We acquire the following lemma proved in \cite[p. 325-326]{PP_optimal_approx_18}. 
\begin{lemma}[\textbf{$L^p$-norm of a composition function {\cite[Lemma 5.2]{PP_optimal_approx_18}}}]
\label{composition_norm}	
Suppose $m, n\in\mathbb{N}$ with $n\leq m$. Let $U\subset \mathbb{R}^m$ be open, $t: U\to \mathbb{R}^n$ be continuously differentiable, and $\varnothing\neq K \subset U$ be compact. Assume that $Dt(x)\in\mathbb{R}^{n\times m}$ has a full rank $\forall x\in K$. Then, there is a constant $C$ fulfilling $\int_{K} (f\hspace{0.5mm}\circ\hspace{0.5mm}t )(x)dx \leq C\cdot\int_{t(K)} f(y)dy$ for all Borel measurable functions such that $f: t(K)\to \mathbb{R}_{+}$. For all $0<p<\infty$ and $f: t(K)\to \mathbb{R}$ being measurable  
\begin{equation}
\label{Compositon_L_p_norm_2}
\| f \circ t\|_{L^{p}(K)}\leq C^{1/p} \cdot \|f\|_{L^p(t(K))}.
\end{equation}	    
	   
\end{lemma}

For an open set $\Omega\in\mathbb{R}^d$ and $n\in\mathbb{N}_0$, let $C^n(\Omega)$ denote the set of $n$ times continuously differentiable functions on $\Omega$. The space of test functions is denoted as $C_c^{\infty}(\Omega)\eqdef \big\{f\in C^\infty(\Omega) | \textnormal{supp} \hspace{1mm} f\subset\subset \Omega\big\}$. For $f\in C^n(\Omega)$ and $\bm{\alpha}\in\mathbb{N}_0^d$ with $|\bm{\alpha}|\leq n$, $D^{\bm{\alpha}}f\eqdef\partial^{|\bm{\alpha}|}f\big/ \partial x_1^{\alpha_1}\ldots \partial x_d^{\alpha_d}$. Meanwhile, we hereby define a Sobolev space \cite{IGP_error_bounds_19,Pinkus99approximationtheory}.    
\begin{definition}[\textbf{Sobolev space {\cite[Definition 3.1]{IGP_error_bounds_19}}}]
\label{Sobolev_space_def}
Let $n,d\in\mathbb{N}$ and $\Omega\subset \mathbb{R}^d$ denote an open subset of $\mathbb{R}^d$. Let $L^p(\Omega)$ denote the Lebesgue spaces on $\Omega$ for $1\leq p\leq \infty$. The Sobolev space $\bm{\mathcal{W}}^{n,p}(\Omega)$ is defined $\forall\bm{\alpha}\in \mathbb{N}_0^d$ as $\bm{\mathcal{W}}^{n,p}(\Omega)\eqdef\big\{ f\in L^p(\Omega): D^{\bm{\alpha}}f\in L^p(\Omega) \hspace{1mm} \textnormal{with} \hspace{1mm} |\bm{\alpha}|\leq n   \big\}$ equipped with the norm defined for $f\in \bm{\mathcal{W}}^{n,p}(\Omega)$ as $\|f\|_{\bm{\mathcal{W}}^{n,p}(\Omega)}\eqdef \big( \sum_{0\leq|\bm{\alpha}|\leq n} \|D^{\bm{\alpha}}f \|_{L^p(\Omega)}^p\big)^{1/p}$, $1\leq p <\infty$, and 
\begin{equation}
\label{Sob_norm_2}
\|f\|_{\bm{\mathcal{W}}^{n,\infty}(\Omega)}\eqdef \max_{0\leq|\bm{\alpha}|\leq n} \|D^{\bm{\alpha}}f\|_{L^\infty(\Omega)}. 
\end{equation}         	
\end{definition} 
Hereinafter, $\|f\|_{\bm{\mathcal{W}}^{n,p}(\Omega)}=\|f\|_{\bm{\mathcal{W}}^{n,p}}=\|f(x)\|_{\bm{\mathcal{W}}^{n,p}(\Omega; dx)}$. In the meantime, Definition \ref{Sobolev_space_def} is generalized below.       
\begin{definition}[\textbf{Sobolev spaces of $m$ vector-valued functions {\cite[Definition B.1]{IGP_error_bounds_19}}}]
\label{Sobolev_space_vector_valued_def}	
Let $n\in\mathbb{N}_0$ and $m\in\mathbb{N}_{\geq 2}$. The Sobolev spaces of $m$ vector-valued functions are defined as $\bm{\mathcal{W}}^{n,p}(\Omega; \mathbb{R}^m)\eqdef \big\{(f_1, \ldots, f_m): f_i\in \bm{\mathcal{W}}^{n,p}(\Omega) \big\}$ with $\|f\|_{\bm{\mathcal{W}}^{n,p}(\Omega;\mathbb{R}^m)}\eqdef \big(\sum_{i=1}^m \|f_i\|_{\bm{\mathcal{W}}^{n,p}(\Omega)}^p\big)^{1/p}, \hspace{1mm} 1\leq p<\infty$;  
\begin{equation}
\label{Sob_norm_4}
\|f\|_{\bm{\mathcal{W}}^{n,\infty}(\Omega; \mathbb{R}^m)}\eqdef \max_{i=1, \ldots, m} \|f_i\|_{\bm{\mathcal{W}}^{n, \infty}(\Omega)}.
\end{equation}     	 
\end{definition}    
We also acquire the following definition.   
\begin{definition}[{\textbf{Sobolev semi-norm \cite[Definition B.2]{IGP_error_bounds_19}}}]
\label{Sobolev_semi_norm_def}
For $f\in\bm{\mathcal{W}}^{n,p}(\Omega; \mathbb{R}^m)$ with $n,k\in\mathbb{N}_0$, $k\leq n$, and $m\in\mathbb{N}$, the Sobolev semi-norm is defined as $|f|_{\bm{\mathcal{W}}^{k,p}(\Omega; \mathbb{R}^m)}\eqdef \big(\sum_{i=1, |\bm{\alpha}|=k}^m \|D^{\bm{\alpha}} f_i \|_{L^p(\Omega)}^p\big)^{1/p}, \hspace{1mm} 1\leq p<\infty$;
\begin{equation}
\label{Sob_semi_norm_2}
|f|_{\bm{\mathcal{W}}^{k,\infty}(\Omega; \mathbb{R}^m)}\eqdef \max_{i=1, \ldots, m: |\bm{\alpha}|=k} \|D^{\bm{\alpha}} f_i \|_{L^{\infty}(\Omega)}.
\end{equation}  
\end{definition}
For $m=1$ and $k=0$, Definition \ref{Sobolev_semi_norm_def} corroborates that $|f|_{\bm{\mathcal{W}}^{k,p}(\Omega)}=\|f\|_{L^p(\Omega)}$ \cite{IGP_error_bounds_19}. Accordingly, we herein define the Lebesgue spaces of $m$ vector-valued functions. 
\begin{definition}[{\textbf{Lebesgue spaces of vector-valued functions}}]
\label{Lebesgue_space_vector_valued_def}	
Let $m\in\mathbb{N}_{\geq 2}$ and $1\leq p \leq \infty$. Then, the Lebesgue spaces of $m$ vector-valued functions are defined as $L^p(\Omega; \mathbb{R}^m)\eqdef \{(f_1, \ldots, f_m): f_i\in L^p(\Omega) \}$ with $\|f\|_{L^p(\Omega;\mathbb{R}^m)}\eqdef \big(\sum_{i=1}^m \|f_i\|_{L^p(\Omega)}^p\big)^{1/p}, \hspace{1mm} 1\leq p<\infty$; 
	\begin{equation}
	\label{Lebesgue_norm_2}
	\|f\|_{L^{\infty}(\Omega; \mathbb{R}^m)}\eqdef \max_{i=1, \ldots, m} \|f_i\|_{L^{\infty}(\Omega)}.
	\end{equation}    	 
\end{definition}      	
We also adopt the corollary below proved in \cite[p. 23-24]{IGP_error_bounds_19}.  
\begin{corollary}[\textbf{Sobolev (semi-)norm of a composition function {\cite[Corollary B.5]{IGP_error_bounds_19}}}]
\label{coro: Sobolev_function_composition}
For $d,m\in \mathbb{N}$, suppose $\Omega_1\subset \mathbb{R}^d$ and $\Omega_2\subset \mathbb{R}^m$ are both open, bounded, and convex. Then, there is a constant $C=C(d,m)>0$ with the following property: if $f\in\bm{\mathcal{W}}^{1,\infty}(\Omega_1; \mathbb{R}^m)$ and $g\in\bm{\mathcal{W}}^{1,\infty}(\Omega_2)$ are Lipschitz continuous functions such that ran $f\subset \Omega_2$, then $g\circ f\in \bm{\mathcal{W}}^{1,\infty}(\Omega_1)$ and it holds that
\begin{subequations}
\begin{align}
\label{gof_Sob_semi_norm}
|g \circ f|_{\bm{\mathcal{W}}^{1,\infty}(\Omega_1)} &\leq C  |g|_{\bm{\mathcal{W}}^{1,\infty}(\Omega_2)} |f|_{\bm{\mathcal{W}}^{1,\infty}(\Omega_1; \mathbb{R}^m)}   \\
\label{gof_Sob_norm}
\|g \circ f \|_{\bm{\mathcal{W}}^{1,\infty}(\Omega_1)}& \leq C \max \{ \|g\|_{L^{\infty}(\Omega_2)}, \tilde{g}\tilde{f} \},
\end{align}
\end{subequations} 
where $\tilde{g}=|g|_{\bm{\mathcal{W}}^{1,\infty}(\Omega_2)}$ and $\tilde{f}=|f|_{\bm{\mathcal{W}}^{1,\infty}(\Omega_1; \mathbb{R}^m)}$. 	
\end{corollary}	 

\section{Matrix-Vector Product Approximation}
\label{sec: matrix_vector_product_approximation}
\subsection{Problem Motivation}
\label{subsec: prob_motivation}
It has been corroborated theoretically and empirically that deep networks -- in comparison with shallow networks -- have the power of exponential expressivity \cite{Rolnick_ICLR_18,NIPS_Expo_Expre_2016,Lin_2017,PMLR-v49-Eldan_16,Mhaskar_Deep_vs_Shallow_16,Poggio_IJAC_17}. Exponential expressivity, however, comes at a price of the considerable training difficulty of deep architectures.\footnote{Regarding AI/ML algorithms proposed for wireless communications, signal processing, and networking \cite{QFQH_18,Chen_Saaf_ANN_ML_19,Huang_DL_5G'20,Wang_DL_China_Commun'17}, we have witnessed that most reported empirical results are based on shallow networks.} Deep architectures' training often associates -- due to successive non-linearities \cite{Romero_FitNets'15} -- mathematically intractable non-convex optimization problems. Such problems' gradient-based solutions exhibit an increment in the probability of finding poor local minima with the depth of an architecture \cite{Erhan_DN_training_difficulty'09}. Along with this global optimization issue, plateaus and saddle points are common in deep architecture learning \cite{Sun2019optimization}. Deep architecture learning is also hampered by local optimization issues such as gradient vanishing/explosion and the possibility of many \textit{dead ReLUs} (in case of ReLU networks) \cite{Sun2019optimization,Bengio_Prac_Recomm_2012,Glorot_Understanding_10}.   

By easing the training difficulty of deep networks for AI/ML classification problems via a model compression, knowledge distillation \cite{Hinton_KD_14} has recently emerged as a promising deep learning paradigm based on a student-teacher AI/ML paradigm. In this paradigm, a student network is trained to predict the hard classification labels as well as the softened output of a teacher network \cite{Hinton_KD_14}. The teacher network's extra information helps the student network to generalize better and such knowledge distillation often provides an improvement in inference speed \cite{Polino_model_compression'18}. To explain these benefits of knowledge distillation, the authors of \cite{Phuong_Under_KD_19} pursued a theoretical work on deep linear classifiers. For these classifiers, it was found that three factors explain the success of knowledge distillation: data geometry, optimization bias, and strong monotonicity \cite{Phuong_Under_KD_19}.     

Following its initial success in easing the training difficulty of deep architectures, knowledge distillation has been deployed to solve many AI/ML classification problems. Regarding the classification problem of learning from noisy labels, the authors of \cite{Li_learning_from_noisy_labels'17} proposed a unified distillation framework that employs label relations in semantic knowledge graph and a small clean dataset. From a small clean dataset, the authors distilled the learned knowledge to facilitate model learning from the entire noisy dataset \cite{Li_learning_from_noisy_labels'17}. Besides learning from noisy dataset, knowledge distillation was also exploited for model compression that would enable the efficient execution of deep learning models in resource-constrained environments \cite{Polino_model_compression'18}. For resource-constrained environments such as mobile or embedded devices, the authors of \cite{Polino_model_compression'18} proposed two model compression schemes that jointly leverage weight quantization and knowledge distillation of larger teacher networks into compressed student networks. Moreover, knowledge distillation was also exploited to blend an LSTM teacher with a CNN student such that more accurate CNNs -- more computationally efficient than LSTMs -- can be trained under the guidance of LSTMs \cite{Geras_blending'16}.

By extending knowledge distillation, the authors of \cite{Romero_FitNets'15} introduced \textit{hint training} which is a training scheme based on a \textit{hint} defined as the output of a teacher’s hidden layer deployed to guide a \textit{guided layer} of a student \cite{Romero_FitNets'15}. Student models -- for AI/ML classification problems -- that are thinner and deeper were successfully trained using hint training \cite{Romero_FitNets'15}. By exploiting hint training and knowledge distillation, the authors of \cite{Saputra2019_KD_Regressor} introduced two teacher-student training schemes for an AI/ML regression problem called deep pose regressor. For such a regressor, the authors proposed attentive imitation loss and attentive hint training \cite{Saputra2019_KD_Regressor} which are, respectively, modifications of knowledge distillation and hint training. Moreover, the authors of \cite{Vapnik_LUPI_JMLR'15} considered the scenario of an intelligent teacher providing a student -- only during training -- with privileged information and introduced another emerging teacher-student AI/ML paradigm dubbed learning using privileged information \cite{Vapnik_LUPI_JMLR'15}. Unifying learning using privileged information \cite{Vapnik_LUPI_JMLR'15} and knowledge distillation \cite{Hinton_KD_14}, the authors of \cite{Lopez-Paz_generalized_distillation'16} proposed generalized distillation which is a teacher-student AI/ML paradigm broadly applicable in the context of \textit{machines-teaching-machines} \cite{Lopez-Paz_generalized_distillation'16}.

On the other hand, apart from the contaminating Gaussian noise or error vector,  matrix-vector products model numerous problems in network science and graph signal processing; network neuroscience and brain physics; and wireless communications and signal processing. Beginning with wireless communications and signal processing, matrix-vector products model -- amongst others -- the following problems: a MIMO received signal \cite{EBRAAAH07}; a received signal over a doubly selective orthogonal frequency division multiplexing (OFDM) channel \cite{P_Sch_LCE_OFDM_04} and a doubly selective MIMO-OFDM channel \cite{Yu_MIMO_OFDM_05}; a massively concurrent non-orthogonal multiple-access (MC-NOMA) \cite{Razvan_6G_19} received signal; and a multi-carrier code division multiple access (MC-CDMA) \cite{Book_OFDM_MC-CDMA_06,Helard:hal-00283645} -- also known as OFDM/CDMA -- received signal. Continuing to network science and graph signal processing, matrix-vector products model -- among others -- the following problems: non-parametric regression for graph-aware signal reconstruction; joint inference of signals and graphs; identification of directed graph topologies and tracking of dynamic networks \cite{Giannakis_Topo_ident_18,Shen_Ten_Decomp_17}; and graph Fourier transforms \cite{Ortega_GSP_18}.  \vspace{-0.28mm}     

In light of the highlighted teacher-student AI/ML paradigms and the prevalence of modeling using matrix-vector products, teacher deep ReLU FNNs $\bm{\Phi}_{D, \varepsilon}^{T}(\bm{W},\bm{x})$ -- fed with $[(\textnormal{vec}(\bm{W}))^T, \bm{x}^T]^T$ for $\bm{W}\in\mathbb{R}^{m\times n}$ and $\bm{x}\in\mathbb{R}^{n}$ -- that can effectively approximate $\bm{Wx}\in\mathbb{R}^{m}$ can be deployed in the training of lightweight (compressed) student ReLU FNNs that can perform as good as teacher ReLU FNNs. In this vein, three major AI/ML problems arise: modeling $\bm{W}$ (e.g., MIMO channel modeling), estimating $\bm{W}$ (e.g., blind estimation of an OFDM channel), and inferring $\bm{x}$ (e.g., MIMO and OFDM signal detection) using ReLU FNNs. Beginning with ReLU FNN based inference of $\bm{x}$, once a teacher deep ReLU FNN $\bm{\Phi}_{D, \varepsilon}^{T}(\tilde{\bm{W}},\bm{x})$ -- for $\tilde{\bm{W}}=\textnormal{diag}(1, \ldots, 1)\in\{0,1\}^{m\times n}$ -- is successfully trained, a lightweight student ReLU FNN $\bm{\Phi}_{D, \varepsilon}^{S}(\bm{x})$ can be trained using knowledge distillation \cite{Hinton_KD_14} or hint training \cite{Romero_FitNets'15} by exploiting, respectively, the softened output of the teacher output (as well as the hard training labels of the student regarding $\bm{x}$) and the hint layer of the teacher. Continuing to the estimation of $\bm{W}$ using teacher-student ReLU FNNs, if a teacher deep ReLU FNN $\bm{\Phi}_{D, \varepsilon}^{T}(\bm{Wx},\bm{x}^T(\bm{x}\bm{x}^T)^{-1})$ is successfully trained, a lightweight student ReLU FNN $\bm{\Phi}_{D, \varepsilon}^{S}(\bm{Wx})$ can be trained using attentive imitation loss or attentive hint training \cite{Saputra2019_KD_Regressor}. On modeling $\bm{W}$ using teacher-student ReLU FNNs, furthermore, in case a teacher deep ReLU FNN $\bm{\Phi}_{D, \varepsilon}^{T}(\bm{W},\bm{x})$ is successfully trained, a lightweight student ReLU FNN $\bm{\Phi}_{D, \varepsilon}^{S}(\bm{x})$ can be trained using attentive imitation loss or attentive hint training \cite{Saputra2019_KD_Regressor}.   

The success of the aforementioned three solutions -- based on teacher-student AI/ML paradigms -- depends on the successful training of matrix-vector product approximating teacher deep ReLU FNNs. Such a deep ReLU FNN's training can be guided by an optimal deep ReLU FNN architecture whose depth, connectivity, maximum width, the maximum absolute value of the weights, and the number of neurons are all predicted by a theory. Such a theory will help ease the considerable training difficulty of teacher deep ReLU FNNs. This is in line with the overarching goal of ToDL research which aims to identify the critical ingredients that contribute to successful training of deep neural networks \cite{Sun_SPM_global_landscape'20}. Accordingly, a deep approximation theory on error bounds of a matrix-vector product approximation using deep ReLU FNNs is needed. By the same token, facilitating an understanding on a deep representation which is one of the three fundamental problems on ToDL \cite{Sun_SPM_global_landscape'20,Poggio_Theo_Issues_Dnets_2020}, quantification of error bounds on a ReLU FNN based matrix-vector product approximation inspires much more research toward an interpretable/explainable AI/ML.

\subsection{Problem Formulation}
\label{subsec: Problem_formulation}
Let $\bm{W}\in\mathbb{R}^{m\times n}$ and $\bm{x}\in\mathbb{R}^{n}$ be a matrix and a vector whose product is going to be approximated by a ReLU FNN $\bm{\Phi}_{D, \varepsilon}^{\Cross}$ fed with $[(\textnormal{vec}(\bm{W}))^T, \bm{x}^T]^T$ -- denoted as $\bm{\Phi}_{D, \varepsilon}^{\Cross}\big(\bm{W},\bm{x}\big)$. Suppose $\bm{w}_i^T\eqdef (\bm{W}(i,:))^T\in\mathbb{R}^n$. Then, $\bm{W}= [\bm{w}_1^T, \ldots, \bm{w}_m^T]^T$ and    
\begin{equation}
\label{W_x_product}
\bm{W}\bm{x}=[\bm{w}_1\bm{x}, \dots, \bm{w}_m\bm{x}]^T\in\mathbb{R}^{m}. 
\end{equation} 
Regarding the product in (\ref{W_x_product}) and its approximation $\bm{\Phi}_{D, \varepsilon}^{\Cross}\big(\bm{W},\bm{x}\big)$, we are going to derive the error bounds for $\| \bm{\Phi}_{D, \varepsilon}^{\Cross}(\bm{W},\bm{x})-\bm{W}\bm{x} \|_{L^{\infty}([-D, D]^{2n}; \mathbb{R}^{m})}$ and $\| \bm{\Phi}_{D, \varepsilon}^{\Cross}(\bm{W},\bm{x})-\bm{W}\bm{x}\|_{\bm{\mathcal{W}}^{1,\infty}((-D, D)^{2n}; \mathbb{R}^{m})}$. 

For a complex matrix $\bm{W}\in\mathbb{C}^{m\times n}$ and a complex vector $\bm{x}\in\mathbb{C}^{n}$, we are also going to derive an error bound for $\| \bm{\Phi}_{D, \varepsilon}^{\Cross_c}(\bm{W}_{1,2},\bm{x}_{1,2}) - \bm{p}_{1,2} \|_{L^{\infty}([-D, D]^{2n}; \mathbb{R}^{2m})}$, where $\bm{W}_{1,2}\eqdef [\bm{W}_1, \bm{W}_2 ]\in\mathbb{R}^{m\times 2n}$ for $\bm{W}_1\eqdef \textnormal{Re}\{\bm{W}\}\in\mathbb{R}^{m\times n}$ and $\bm{W}_2\eqdef \textnormal{Im}\{\bm{W}\}\in\mathbb{R}^{m\times n}$; $\bm{x}_{1,2}\eqdef [ \bm{x}_1^T, \bm{x}_2^T ]^T\in\mathbb{R}^{2n}$ for $\bm{x}_1\eqdef \textnormal{Re}\{\bm{x}\}\in\mathbb{R}^{n}$ and $\bm{x}_2\eqdef \textnormal{Im}\{\bm{x}\}\in\mathbb{R}^{n}$; $\bm{p}_{1,2}\eqdef[\bm{p}_1^T, \bm{p}_2^T]^T\in\mathbb{R}^{2m}$ for $\bm{p}_1\eqdef \textnormal{Re}\{\bm{W}\bm{x}\}\in\mathbb{R}^{m}$ and $\bm{p}_2\eqdef \textnormal{Im}\{\bm{W}\bm{x}\}\in\mathbb{R}^{m}$; and $\bm{\Phi}_{D, \varepsilon}^{\Cross_c}(\bm{W}_{1,2},\bm{x}_{1,2})$ is a ReLU FNN fed with $[(\textnormal{vec}(\bm{W}_{1,2}))^T, \bm{x}_{1,2}^T]^T$.     

\section{The Developed Deep Approximation Theory}
\label{sec: dev_approx_theory}
\subsection{Error Bounds in Lebesgue Norms}
\label{subsec: error_bounds_Lebesgue_norms}
First, we state the following proposition proved using the \textquotedblleft sawtooth'' construction of \cite{Yarotsky_error_bounds_17} and \cite{Telgarsky2015_DFNN}.
\begin{proposition}[{\textbf{\cite[Proposition III.1]{grohs2019deep}}}]
	\label{prop_approx_x^2}
	For all $\varepsilon\in(0, 1/2)$ and $\rho(x)\eqdef \textnormal{max}(x,0)$, there exist a constant $C>0$ and a FNN $\bm{\Phi}_{\varepsilon}^{\textnormal{sq}}\in\bm{\mathcal{NN}}^{1,1}_{\infty, \infty, \rho}$ satisfying $\mathcal{L}(\bm{\Phi}_{\varepsilon}^{\textnormal{sq}})\leq C\log_2(\varepsilon^{-1})$, $\mathcal{W}(\bm{\Phi}_{\varepsilon}^{\textnormal{sq}})= 4$, $\mathcal{B}(\bm{\Phi}_{\varepsilon}^{\textnormal{sq}})\leq 4$, $\bm{\Phi}_{\varepsilon}^{\textnormal{sq}}(0)=0$, and $\| \bm{\Phi}_{\varepsilon}^{\textnormal{sq}}(x)-x^2 \|_{L^{\infty}([0,1])}\leq \varepsilon$. 
	\proof The proof is provided as supplementary material.     
\end{proposition}  
An economic implementation of this approximating ReLU FNN is also possible as stated in \cite[Proposition III.2]{DL_Approx_Theory'21}. Building from Proposition \ref{prop_approx_x^2}, the following theorem follows. 
\begin{theorem}
\label{Theorem_upper_approximation_bound}
Let $D\in \mathbb{R}_{+}$, $\bm{W}\in\mathbb{R}^{m\times n}$, and $\bm{x}\in\mathbb{R}^{n}$. For all $\varepsilon \in (0, 1/2)$ and $\rho(x)\eqdef \textnormal{max}(x,0)$, there exists a constant $C>0$ such that there is a FNN $\bm{\Phi}_{D, \varepsilon}^{\Cross}\in\bm{\mathcal{NN}}_{\infty, \infty, \rho}^{n(m+1),m}$ satisfying $\mathcal{L}( \bm{\Phi}_{D, \varepsilon}^{\Cross} )\leq C\log_2 (nD^2 \varepsilon^{-1})$, $\mathcal{W}( \bm{\Phi}_{D, \varepsilon}^{\Cross} )\leq 12mn$, $\mathcal{B}( \bm{\Phi}_{D, \varepsilon}^{\Cross})\leq \max\{4, 2D^2 \}$, $\bm{\Phi}_{D, \varepsilon}^{\Cross}(\bm{0},\bm{x})=\bm{\Phi}_{D, \varepsilon}^{\Cross}(\bm{W},\bm{0})=\bm{0}$ and     
\begin{equation}
\label{matrix_vector_product_approx_NN_bound}
\| \bm{\Phi}_{D, \varepsilon}^{\Cross}(\bm{W},\bm{x})-\bm{W}\bm{x} \|_{L^{\infty}([-D, D]^{2n}; \mathbb{R}^{m})}\leq \varepsilon.   
\end{equation}	
	
\proof The proof is provided in Appendix \ref{proof_Theorem_upper_approximation_bound}.    
\end{theorem}  
    
Based on Theorem \ref{Theorem_upper_approximation_bound}, the following corollary follows. 
\begin{corollary}
\label{Coro_complex_upper_approximation_bound}
Let $D\in \mathbb{R}_{+}$, $\bm{W}\in\mathbb{C}^{m\times n}$, $\bm{x}\in\mathbb{C}^{n}$, $\bm{W}_1\eqdef \textnormal{Re}\{\bm{W}\}\in\mathbb{R}^{m\times n}$, $\bm{W}_2\eqdef \textnormal{Im}\{\bm{W}\}\in\mathbb{R}^{m\times n}$, $\bm{x}_1\eqdef \textnormal{Re}\{\bm{x}\}\in\mathbb{R}^{n}$, and $\bm{x}_2\eqdef \textnormal{Im}\{\bm{x}\}\in\mathbb{R}^{n}$. Let $\bm{p}_1\eqdef \textnormal{Re}\{\bm{W}\bm{x}\}\in\mathbb{R}^{m}$, $\bm{p}_2\eqdef \textnormal{Im}\{\bm{W}\bm{x}\}\in\mathbb{R}^{m}$, $\bm{p}_{1,2}\eqdef[\bm{p}_1^T, \bm{p}_2^T]^T\in\mathbb{R}^{2m}$, $\bm{W}_{1,2}\eqdef [\bm{W}_1, \bm{W}_2 ]\in\mathbb{R}^{m\times 2n}$, and $\bm{x}_{1,2}\eqdef [ \bm{x}_1^T, \bm{x}_2^T ]^T\in\mathbb{R}^{2n}$. For all $\varepsilon \in (0, 1/2)$ and $\rho(x)\eqdef \textnormal{max}(x,0)$, there exists a constant $C>0$ such that there is a FNN $\bm{\Phi}_{D, \varepsilon}^{\Cross_c}\in\bm{\mathcal{NN}}_{\infty, \infty, \rho}^{2n(m+1),2m}$ satisfying $\mathcal{L}( \bm{\Phi}_{D, \varepsilon}^{\Cross_c} )\leq C\log_2 (4nD^2 \varepsilon^{-1})$, $\mathcal{W}( \bm{\Phi}_{D, \varepsilon}^{\Cross_c})\leq 48mn$, $\mathcal{B}( \bm{\Phi}_{D, \varepsilon}^{\Cross_c} )\leq \max\{4, 2D^2 \}$, $\bm{\Phi}_{D, \varepsilon}^{\Cross_c}(\bm{0},\bm{x}_{1,2})=\bm{\Phi}_{D, \varepsilon}^{\Cross_c}(\bm{W}_{1,2},\bm{0} )=\bm{0}$ and     
\begin{equation}
\label{complex_matrix_vector_product_approx_NN_bound}
\| \bm{\Phi}_{D, \varepsilon}^{\Cross_c}\big(\bm{W}_{1,2},\bm{x}_{1,2} \big)- \bm{p}_{1,2} \|_{L^{\infty}([-D, D]^{2n}; \mathbb{R}^{2m})}\leq \varepsilon.
\end{equation}  
\proof The proof is relegated to Appendix \ref{proof_Coro_complex_upper_approximation_bound}.    
\end{corollary} 
\begin{remark}
	\label{rem_upper_lebsequ_bound}
	Theorem \ref{Theorem_upper_approximation_bound} and Corollary \ref{Coro_complex_upper_approximation_bound} affirm that $\varepsilon\to 0$ at an exponential rate  provided that $\mathcal{L}( \bm{\Phi}_{D, \varepsilon}^{\Cross} ), \mathcal{L}( \bm{\Phi}_{D, \varepsilon}^{\Cross_c} ) \to \infty$. Therefore, as ReLU FNNs get much deeper, they would result in the lowest approximation error.    
\end{remark}

\subsection{Error Bounds in Sobolev Norms}
\label{subsec: error_bounds_Sob_norms}
First, we state the proposition proved in \cite[p. 29-30]{IGP_error_bounds_19}.   
\begin{proposition}[{\textbf{\cite[Proposition C.1]{IGP_error_bounds_19}}}]
\label{Sob_prop_approx_x^2}
For all $\varepsilon\in(0, 1/2)$ and $\rho(x)\eqdef \textnormal{max}(x,0)$, there exist constants $C_1, C_2, C_3, C_4>0$ such that there is a FNN $\bm{\Phi}_{\varepsilon}^{\textnormal{sq}}\in\bm{\mathcal{NN}}^{1,1}_{\infty, \infty, \rho}$ with $\mathcal{M}(\bm{\Phi}_{\varepsilon}^{\textnormal{sq}})\leq C_1\log_2(\varepsilon^{-1})$, $\mathcal{L}(\bm{\Phi}_{\varepsilon}^{\textnormal{sq}})\leq C_2\log_2(\varepsilon^{-1})$, $\mathcal{N}(\bm{\Phi}_{\varepsilon}^{\textnormal{sq}})\leq C_3\log_2(\varepsilon^{-1})$, and $\bm{\Phi}_{\varepsilon}^{\textnormal{sq}}(0)=0$ such that   
\begin{subequations}
\begin{align}
\label{Sob_approx_bound_x^2}
\| \bm{\Phi}_{\varepsilon}^{\textnormal{sq}}(x)-x^2 \|_{\bm{\mathcal{W}}^{1,\infty}((0,1); dx)}& \leq \varepsilon  \\
\label{Sob_semi_norm_bound_x^2}
|\bm{\Phi}_{\varepsilon}^{\textnormal{sq}} |_{\bm{\mathcal{W}}^{1,\infty}((0,1))}\leq & C_4.
\end{align}
\end{subequations}       
\end{proposition} 
Based on Proposition \ref{Sob_prop_approx_x^2}, the following result follows.  
\begin{proposition}[{\textbf{\cite[Proposition C.2]{IGP_error_bounds_19}}}]
	\label{Sob_Proposition_product_NN}
	For all $\varepsilon \in (0, 1/2)$, $D\in \mathbb{R}_{+}$, and $\rho(x)\eqdef \textnormal{max}(x,0)$, there exist constants $\bar{C}, C_1>0$ and $C_2=C_2(D)>0$ and $\bm{\Phi}_{D, \varepsilon}^{\tilde{\times}(w,x)}\in\bm{\mathcal{NN}}_{\infty, \infty, \rho}^{2,1}$: 
	\begin{subequations}
	\begin{align}
	\label{w_x_product_prop_1}
	\| \bm{\Phi}_{D, \varepsilon}^{\tilde{\times}(w,x)}(w,x)-wx \|_{\bm{\mathcal{W}}^{1,\infty}((-D, D)^2; dwdx)}&\leq \varepsilon      \\
	\label{w_x_product_prop_2}
	\bm{\Phi}_{D, \varepsilon}^{\tilde{\times}(w,x)}(0,x)=\bm{\Phi}_{D, \varepsilon}^{\tilde{\times}(w,x)}(x,0)=0, \hspace{1mm} &\forall x\in\mathbb{R}  \\
	\label{w_x_product_prop_3}
	| \bm{\Phi}_{D, \varepsilon}^{\tilde{\times}(w,x)}|_{\bm{\mathcal{W}}^{1,\infty}((-D, D)^2)}&\leq \bar{C}D    
	\end{align}
	\end{subequations}
	\begin{equation}
	\label{w_x_product_prop_4}
	\mathcal{L}( \bm{\Phi}_{D, \varepsilon}^{\tilde{\times}(w,x)}),\mathcal{M}( \bm{\Phi}_{D, \varepsilon}^{\tilde{\times}(w,x)}), \mathcal{N}\big( \bm{\Phi}_{D, \varepsilon}^{\tilde{\times}(w,x)}\big) \leq C_1\log_2(\varepsilon^{-1})+C_2.
	\end{equation}
	\proof The proof is provided in Appendix \ref{proof_Sob_Proposition_product_NN}. 
\end{proposition}

Building from Proposition \ref{Sob_Proposition_product_NN}, the following theorem follows.    
\begin{theorem}
\label{Theorem_Sobolev_upper_approximation_bound}
Let $D\in \mathbb{R}_{+}$, $\bm{W}\in\mathbb{R}^{m\times n}$, and $\bm{x}\in\mathbb{R}^{n}$. For all $\varepsilon \in (0, 1/2)$ and $\rho(x)\eqdef \textnormal{max}(x,0)$, there exist constants $\bar{\bar{C}}_1, \bar{\bar{C}}_2>0$ such that there is a FNN $\bm{\Phi}_{D, \varepsilon}^{\Cross}\in\bm{\mathcal{NN}}_{\infty, \infty, \rho}^{n(m+1),m}$ satisfying $\mathcal{L}( \bm{\Phi}_{D, \varepsilon}^{\Cross})\leq \bar{\bar{C}}_1 \log_2(\varepsilon^{-1})+\bar{\bar{C}}_2$, $\mathcal{M}( \bm{\Phi}_{D, \varepsilon}^{\Cross})\leq \bar{\bar{C}}_1 \log_2(\varepsilon^{-1})+\bar{\bar{C}}_2$, $\mathcal{N}( \bm{\Phi}_{D, \varepsilon}^{\Cross})\leq \bar{\bar{C}}_1 \log_2(\varepsilon^{-1})+\bar{\bar{C}}_2$, $\bm{\Phi}_{D, \varepsilon}^{\Cross}(\bm{0},\bm{x})=\bm{\Phi}_{D, \varepsilon}^{\Cross}(\bm{W},\bm{0})=\bm{0}$ and     
\begin{equation}
\label{matrix_vector_product_Sob_approx_NN_bound}
\| \bm{\Phi}_{D, \varepsilon}^{\Cross}(\bm{W},\bm{x})-\bm{W}\bm{x}\|_{\bm{\mathcal{W}}^{1,\infty}((-D, D)^{2n}; \mathbb{R}^{m})}\leq \varepsilon.   
\end{equation}		
\proof The proof is deferred to Appendix \ref{proof_Theorem_Sobolev_upper_approximation_bound}.    
\end{theorem}

\begin{figure*}
	\vspace{-0.9cm}
	\subfloat{
		\begin{minipage}[c][1\width]{0.30\textwidth}
			\centering
			\includegraphics[width=1.1\textwidth]{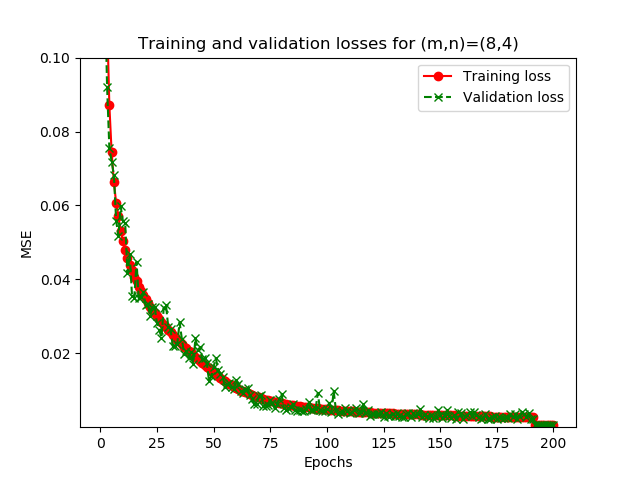}
	\end{minipage}}
	\hfill	
	\subfloat{
		\begin{minipage}[c][1\width]{0.25\textwidth}
			\centering
			\includegraphics[width=1.1\textwidth]{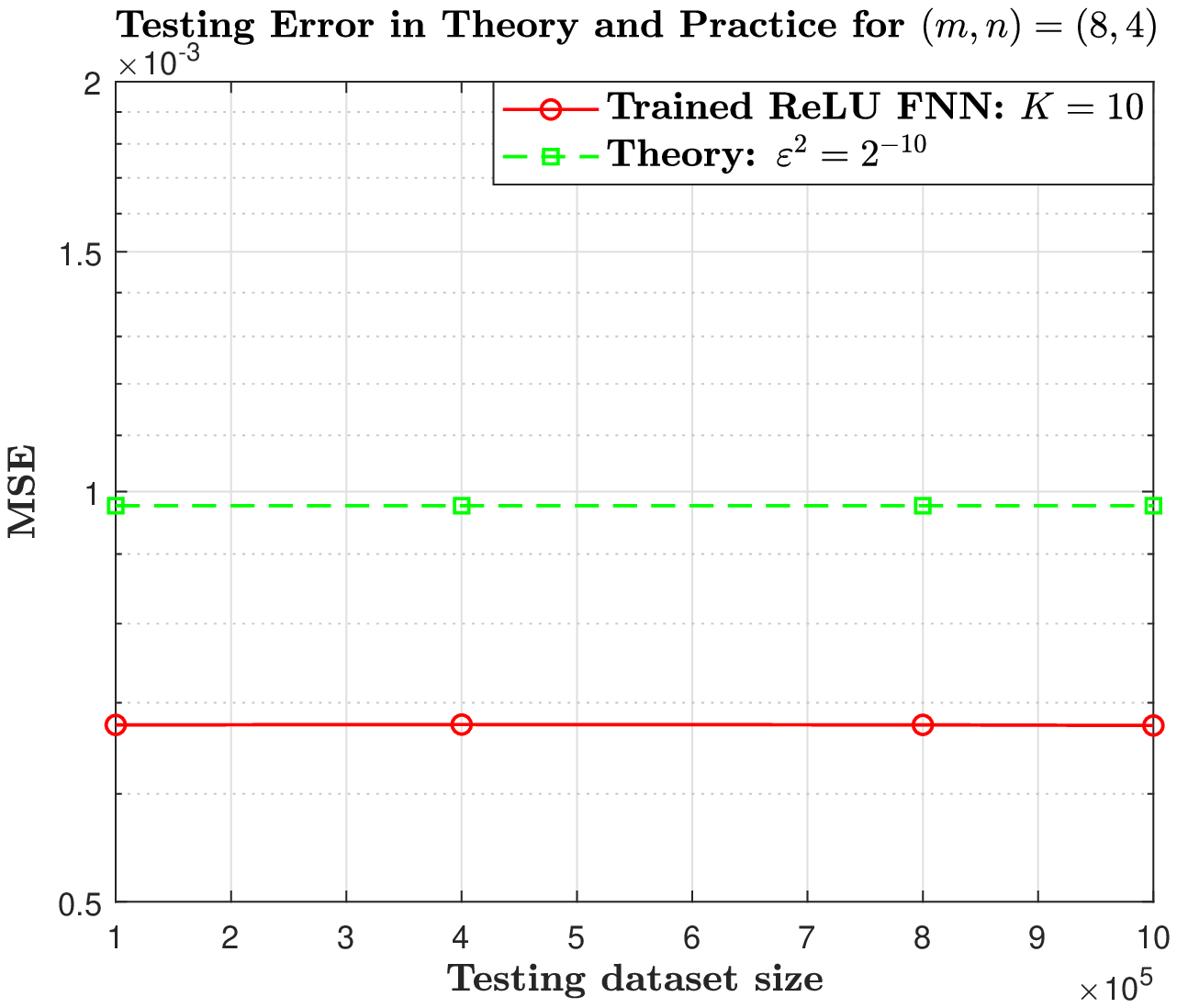}
	\end{minipage}} \vspace{-0.70cm}
	\caption{Training loss and testing error of the deep ReLU product FNN characterized by Theorem \ref{Theorem_upper_approximation_bound}: $K=10$ and $\varepsilon^2=2^{-10}$.}
	\label{fig: ReLU_FNN_training_loss_and_testing_error_equispaced_product}
\end{figure*}

\begin{figure*}
	\vspace{-0.9cm}
	\subfloat{
		\begin{minipage}[c][1\width]{0.30\textwidth}
			\centering
			\includegraphics[width=1.1\textwidth]{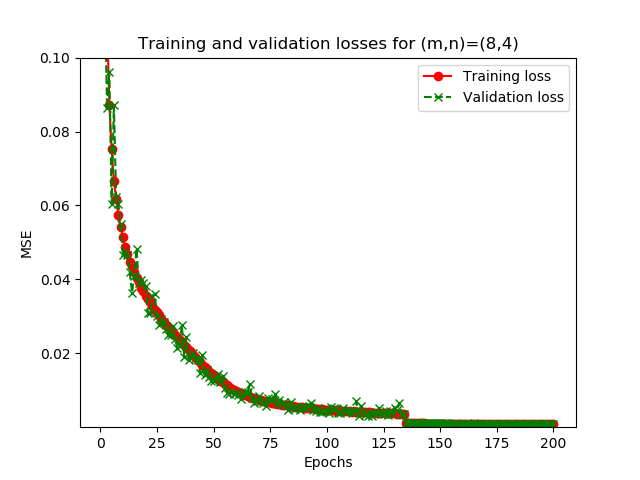}
	\end{minipage}}
	\hfill 		
	\subfloat{
		\begin{minipage}[c][1\width]{0.25\textwidth}
			\centering
			\includegraphics[width=1.1\textwidth]{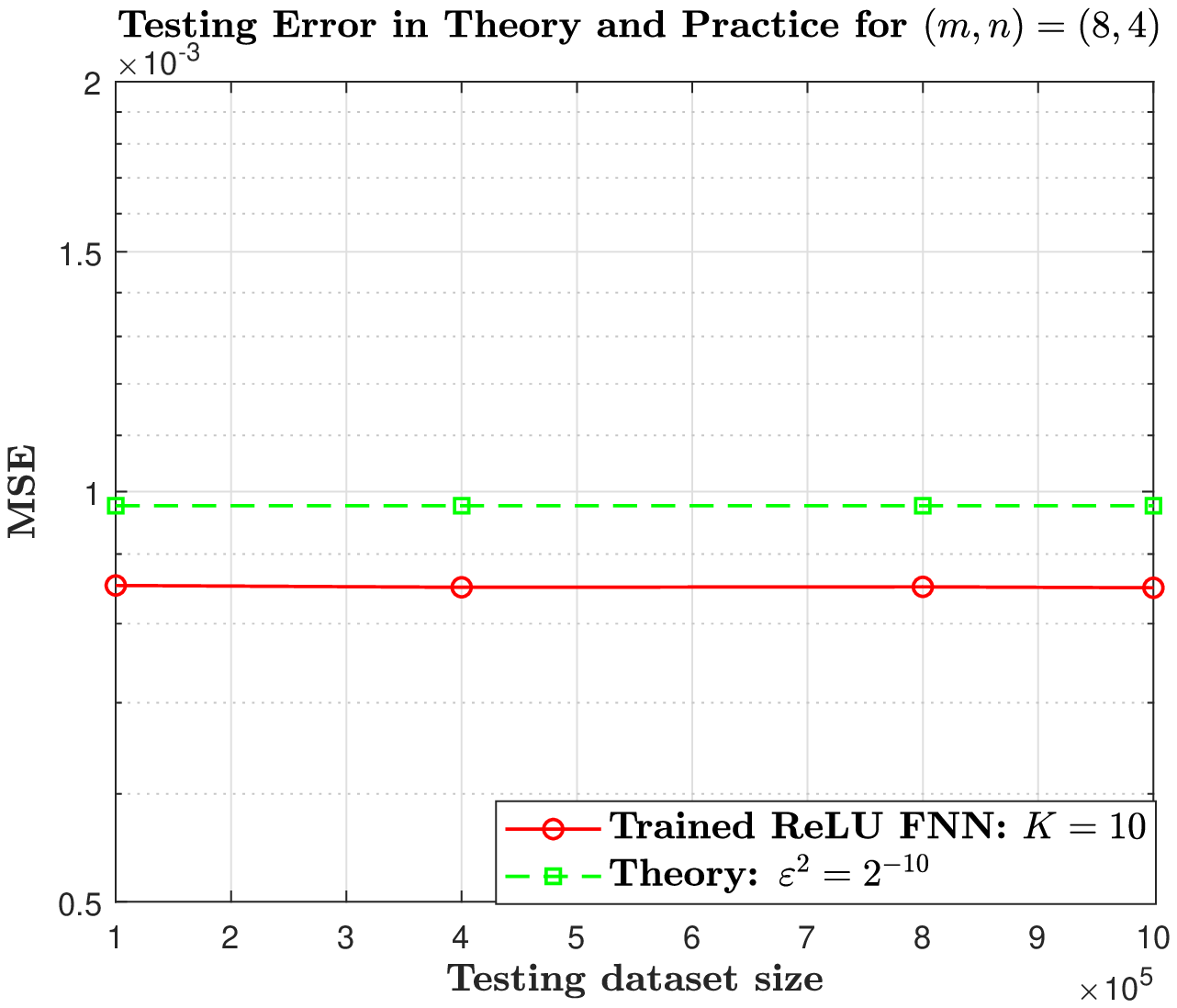}
	\end{minipage}}  \vspace{-0.70cm}
	\caption{Training loss and testing error of the deep ReLU product FNN characterized by Theorem \ref{Theorem_upper_approximation_bound}: $K=10$ and $\varepsilon^2=2^{-10}$.}
	\label{fig: ReLU_FNN_training_loss_and_testing_error_equispaced_product_2}
\end{figure*}

\section{Computer Experiments}
\label{sec: comp_experiments}
This section presents the procedures and empirical results of our computer experiments conducted to assess the accuracy of a matrix-vector product approximation using deep ReLU FNNs. The deep ReLU FNNs' approximations that have been experimented are a real and a complex matrix-vector product approximations as quantified by Theorem \ref{Theorem_upper_approximation_bound} and Corollary \ref{Coro_complex_upper_approximation_bound}, respectively.      

Carried out on the NIST GPU (graphics processing unit) cluster dubbed \textsc{ENKI}, \textsc{Keras} \cite{Keras_web} with \textsc{TensorFlow} \cite{TensorFlow_web} as a backend was used for the offline training and online testing of deep ReLU FNNs approximating both real and complex matrix-vector products. Our matrix-vector product approximating network training employed the (hyper)parameters listed in Table \ref{table: Hyperparameters}. Besides Table \ref{table: Hyperparameters}'s (hyper)parameters, we deployed four \textsc{Keras} \textit{callbacks}: \textit{model checkpointing}, \textit{TensorBoard}, \textit{early stopping}, and \textit{learning rate reduction} \cite[Ch. 7]{Chollet_DL_with_Python'18}. Regarding the latter two, early stopping and learning rate reduction callbacks were set to have patience over twenty epochs. Per twenty epochs that render performance stagnation, learning rate reduction callback was set to reduce the learning rate by 0.1. Moreover, training and testing datasets were generated in MATLAB$^{\textregistered}$ and then uploaded into \textsc{ENKI}.
\begin{table}	
	\caption{(Hyper)parameters unless otherwise mentioned.}		
	\label{table: Hyperparameters}
	\centering
	\begin{tabular}{ | l | l | l | } 
		\hline      
		\textbf{(Hyper)parameters} & \textbf{Type/Value} & \textbf{Remark(s)}     \\  \hline 
		$(m,n)$  & $(8,4)$   & Chosen dimensions  \\  \hline 
		Learning rate  & 0.001    & Initial value  \\  \hline      
		Epoch size & 200      & Maximum epoch size.      \\  \hline            
		Batch size & 200       & Vary large and small batch         \\    
		&    &  sizes have also been tried.   \\      \hline 
		Optimizer & \textit{Adam} \cite{Adam_ICLR_15} & Adam produced the least          \\   
		&    &  MSE and converged fast.      \\   \hline
		Activation  & ReLU     & The focus of this work has                         \\   
		function &     & been on ReLU FNNs.            \\   \hline 
		Depth of FNNs  & $10$  & Chosen in line with the    \\      
		($K$) &     &  developed theory      \\   \hline
		Training-validation  & 75$\%$ to 25$\%$    & 75$\%$ of the training dataset               \\   
		split &     & is used for training; 25$\%$ of      \\   
		&     & it is used for validation.          \\ \hline
		Initialization for  & \textit{He normal}  & $\bm{W}_k(i,:)\sim \mathcal{N}(\bm{0}, \sigma^2\bm{I}_{N_{k-1}})$,                \\   
		all layers & \cite{He_delving_deep'15}    & $\sigma^2=2/N_{k-1}$ and $i\in[N_k]$.     \\ \hline  
		Kernel   & \texttt{max}$\_$\texttt{norm(8)}  & Deduced from our developed              \\   
		constraint &     &  theory (valid for all layers).           \\  
		\hline
	\end{tabular}
\end{table}  

\subsection{Real Matrix-Vector Product Approximation}
\label{sec: prod_function}
To verify and interpret the theory predicted by Theorem \ref{Theorem_upper_approximation_bound}, we generated real testing and training datasets. The training dataset $\mathcal{D}\eqdef \{(\bm{a}_i, \bm{b}_i)\}_{i=1}^{\check{n}}$ -- $\check{n}=10^6$, $\bm{a}_i=[(\textnormal{vec}(\bm{W}_i))^T, \bm{x}_i^T]^T \in\mathbb{R}^{n(m+1)}$, and $\bm{b}_i=\bm{W}_i\bm{x}_i\in\mathbb{R}^m$ -- was formed from $2^{10}+1$ equispaced points. From these equispaced points generated over $[-2,2]$, specifically, all elements of $\bm{W}_i$ and $\bm{x}_i$ were randomly drawn. Besides, we formed four testing datasets $\mathcal{T}_j\eqdef \{(\bm{a}_{i,j}, \bm{b}_{i,j})\}_{i=1}^{\breve{n}_j}$ -- $\breve{n}_j$ is the size of the $j$-th testing dataset, $\bm{a}_{i,j}=[(\textnormal{vec}(\bm{W}_{i,j}))^T, \bm{x}_{i,j}^T]^T \in\mathbb{R}^{n(m+1)}$, $\bm{b}_{i,j}=\bm{W}_{i,j}\bm{x}_{i,j}\in\mathbb{R}^{m}$, and $j\in[4]$ -- from $2^9+1$ equispaced points. From these equispaced points generated within $[-2,2]$, all elements of $\bm{W}_{i,j}$ and $\bm{x}_{i,j}$ were randomly chosen.  

Setting $\varepsilon=2^{-5}$, Theorem \ref{Theorem_upper_approximation_bound} predicts that the mean squared error (MSE) exhibited by an approximating FNN ($\bm{\Phi}_{D, \varepsilon}^{\Cross}$) is upper bounded by $\varepsilon^2=2^{-10}$ such that $\mathcal{L}( \bm{\Phi}_{D, \varepsilon}^{\Cross} )\leq 9C$, $\mathcal{W}( \bm{\Phi}_{D, \varepsilon}^{\Cross} )\leq 384$, and $\mathcal{B}( \bm{\Phi}_{D, \varepsilon}^{\Cross})\leq 8$ -- w.r.t. $(m, n, D)=(8, 4, 2)$. Thus, choosing $C=2$ leads to $\mathcal{L}( \bm{\Phi}_{D, \varepsilon}^{\Cross} )\leq 18$. Toward this end, we trained a deep ReLU FNN $\bm{\Phi}_{D, \varepsilon}^{\Cross}$ with MSE as a loss function, per-layer neurons $N_0=36$, $N_1=\ldots=N_9=384$, and $N_{10}=8$, and $\mathcal{L}( \bm{\Phi}_{D, \varepsilon}^{\Cross} )=10$ -- a depth-10 deep network. Despite deep networks' power of exponential expressivity, training them is very difficult; cf. Sec. \ref{subsec: prob_motivation}. Guided by Theorem \ref{Theorem_upper_approximation_bound}, however, we succeeded at training a deep ReLU FNN, as shown in Figs. \ref{fig: ReLU_FNN_training_loss_and_testing_error_equispaced_product} and \ref{fig: ReLU_FNN_training_loss_and_testing_error_equispaced_product_2}. Figs. \ref{fig: ReLU_FNN_training_loss_and_testing_error_equispaced_product} and \ref{fig: ReLU_FNN_training_loss_and_testing_error_equispaced_product_2} -- on their left sides -- show the lowest MSE training losses observed at the 200-th epoch and, respectively, equal to $6.3456\times 10^{-4}$ and $8.2080\times 10^{-4}$. Manifesting these training losses, the trained deep ReLU FNNs were tested with four different size testing datasets as shown, respectively, on the right sides of Figs. \ref{fig: ReLU_FNN_training_loss_and_testing_error_equispaced_product} and \ref{fig: ReLU_FNN_training_loss_and_testing_error_equispaced_product_2}. These plots corroborate that the approximation MSE exhibited by the trained deep ReLU FNNs is less than $\varepsilon^2=2^{-10}$. This validates Theorem \ref{Theorem_upper_approximation_bound}.
\begin{figure*}[ht]
	\vspace{-0.9cm}
	\subfloat{
		\begin{minipage}[c][1\width]{0.30\textwidth}
			\centering
			\includegraphics[width=1.1\textwidth]{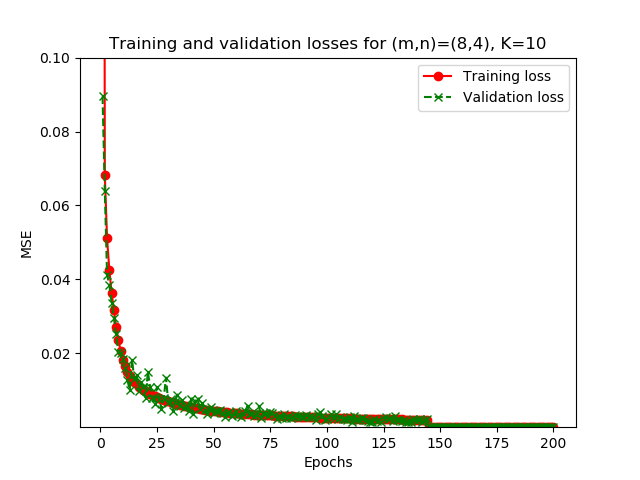}
	\end{minipage}}
	\hfill 		
	\subfloat{
		\begin{minipage}[c][1\width]{0.25\textwidth}
			\centering
			\includegraphics[width=1.1\textwidth]{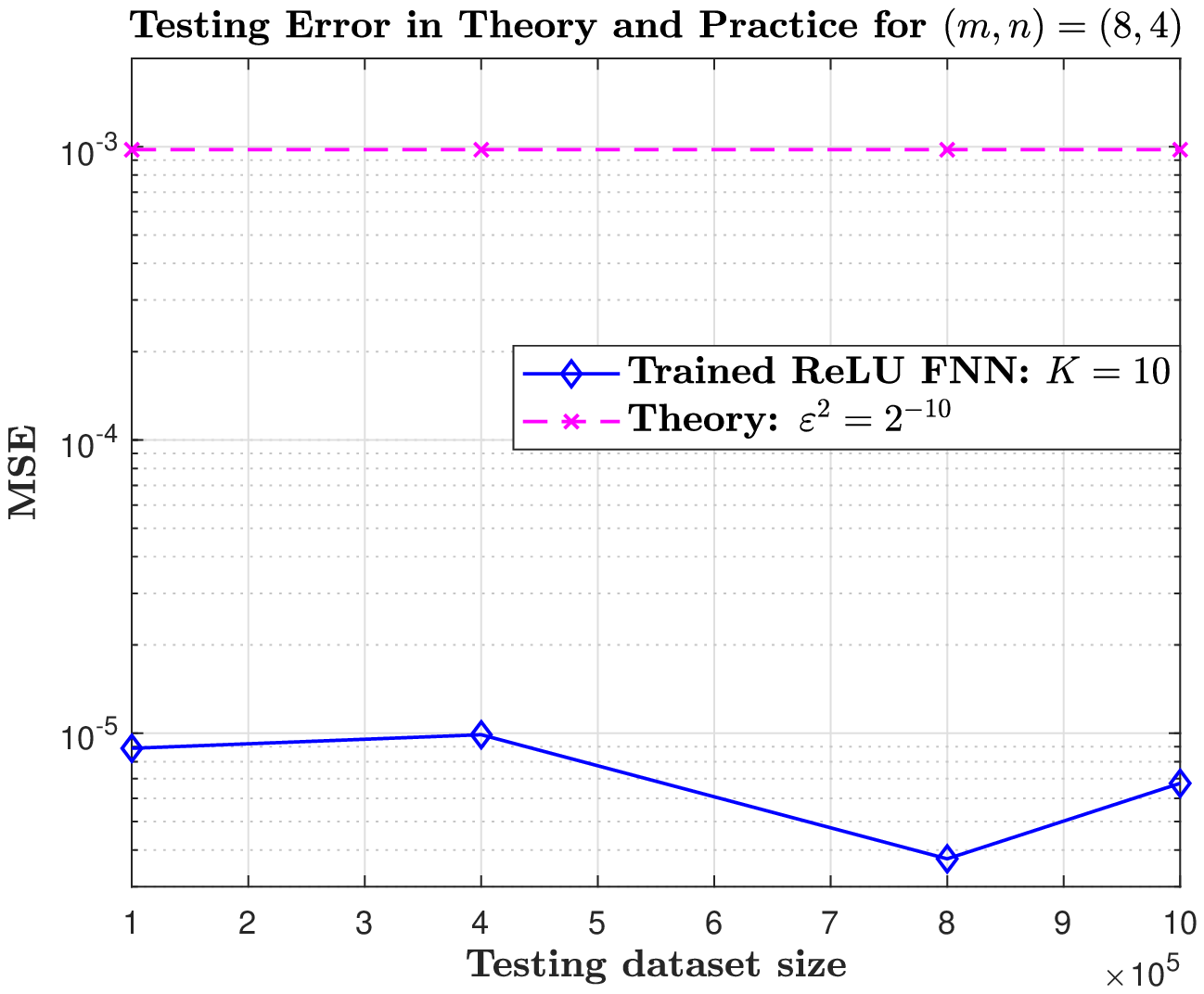}
	\end{minipage}}  \vspace{-0.70cm}
	\caption{Training loss and testing error of the complex product deep ReLU FNN per Corollary \ref{Coro_complex_upper_approximation_bound}: $K=10$ and $\varepsilon^2=2^{-10}$.}
	\label{fig: ReLU_FNN_training_loss_and_testing_error_random_product}
\end{figure*}
\begin{figure*}[ht]
	\vspace{-0.9cm}
	\subfloat{
		\begin{minipage}[c][1\width]{0.30\textwidth}
			\centering
			\includegraphics[width=1.1\textwidth]{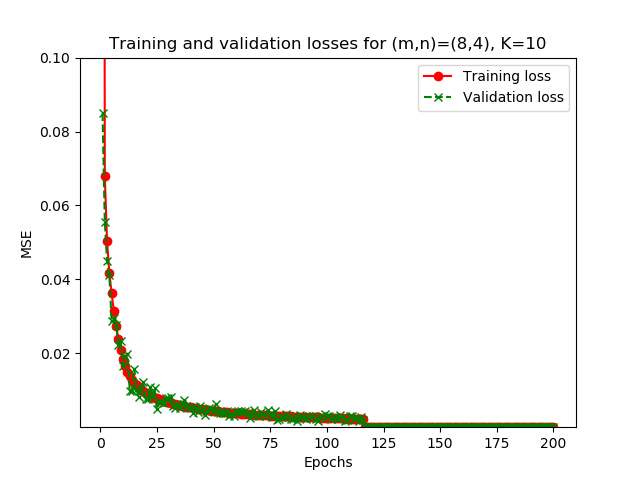}
	\end{minipage}}
	\hfill 		
	\subfloat{
		\begin{minipage}[c][1\width]{0.25\textwidth}
			\centering
			\includegraphics[width=1.1\textwidth]{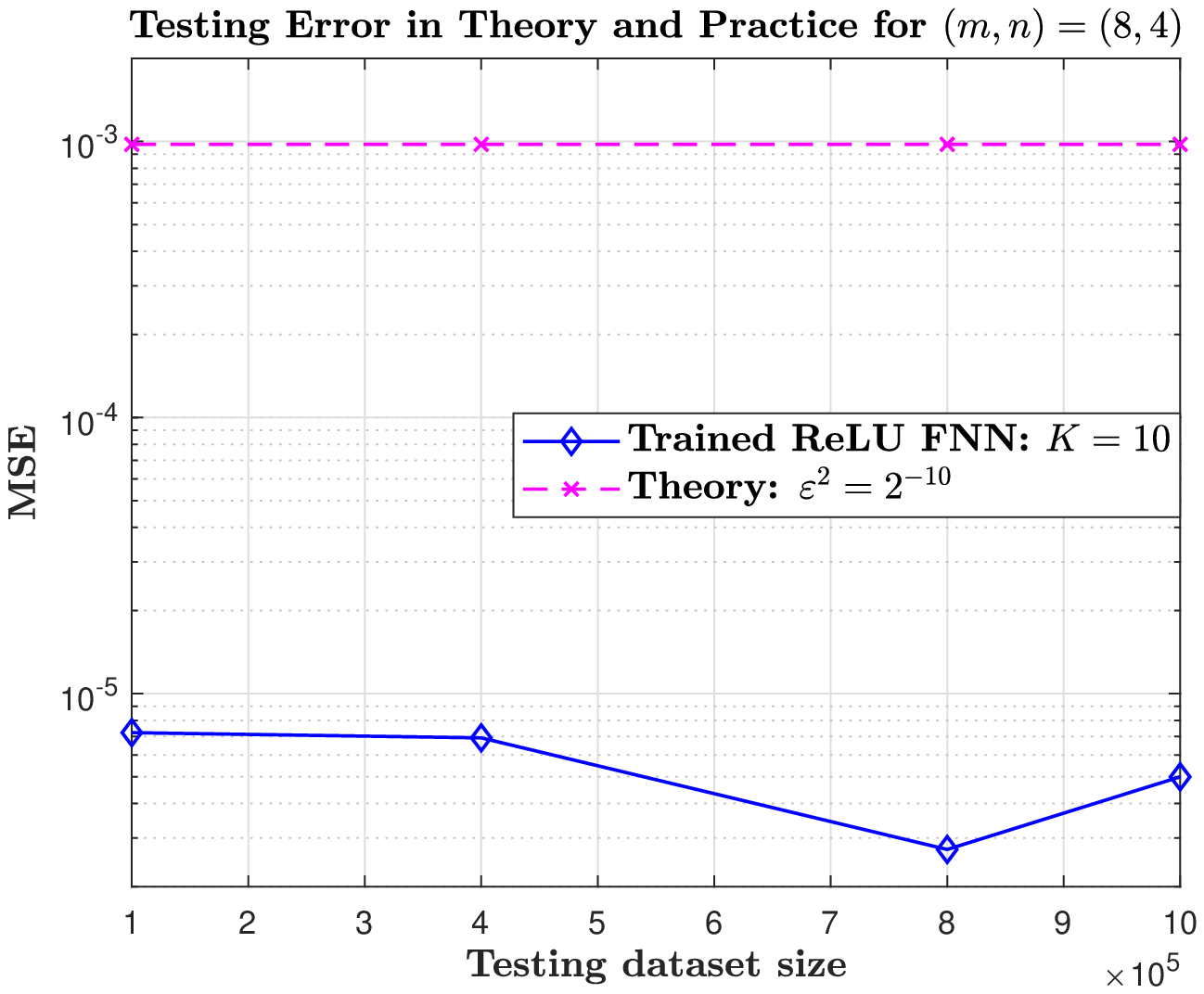}
	\end{minipage}}  \vspace{-0.70cm}
	\caption{Training loss and testing error of the complex product deep ReLU FNN per Corollary \ref{Coro_complex_upper_approximation_bound}: $K=10$ and $\varepsilon^2=2^{-10}$.}
	\label{fig: ReLU_FNN_training_loss_and_testing_error_random_product2}
\end{figure*}
\vspace{-0.30cm}
\subsection{Complex Matrix-Vector Product Approximation}
\label{sec: Wi_Commun}
To validate and interpret the theory predicted by Corollary \ref{Coro_complex_upper_approximation_bound}, we considered the reception of QPSK (quadrature phase shift keying) symbols $\bm{x}_i\eqdef [x_{i,1}, \ldots, x_{i,n}]^T\in\mathbb{C}^n$ -- $\textnormal{Re}\{x_{i,j}\}, \textnormal{Im}\{x_{i,j}\}\in\{-1/\sqrt{2}, 1/\sqrt{2}\}$ -- via a MIMO channel $\bm{H}_i\in\mathbb{C}^{m\times n}$ \cite{EBRAAAH07}. Such a channel is modeled by a Rayleigh fading, i.e., $\bm{H}_i\sim\mathcal{CN}(\bm{0}, \bm{I}_m)$. Having implemented $\bm{H}_i$ and $\bm{x}_i$, we generated testing and training datasets. The training dataset was formed as $\mathcal{D}^c\eqdef \{(\bm{c}_i, \bm{d}_i)\}_{i=1}^{\check{n}}$, where $\check{n}=10^6$, $\bm{c}_i=[\textnormal{Re}\{(\textnormal{vec}(\bm{H}_i))^T\}, \textnormal{Im}\{(\textnormal{vec}(\bm{H}_i))^T\}, \textnormal{Re}\{\bm{x}_i^T\},\textnormal{Im}\{\bm{x}_i^T\}]^T\in\mathbb{R}^{2n(m+1)}$, and $\bm{d}_i=[\textnormal{Re}\{(\bm{H}_i\bm{x}_i)^T\}, \textnormal{Im}\{(\bm{H}_i\bm{x}_i)^T\}]^T \in\mathbb{R}^{2m}$. We also generated four testing datasets $\mathcal{T}_j^c\eqdef \{(\bm{c}_{i,j}, \bm{d}_{i,j})\}_{i=1}^{\breve{n}_j}$, where $\breve{n}_j$ is the size of the $j$-th testing dataset, $\bm{c}_{i,j}=[\textnormal{Re}\{(\textnormal{vec}(\bm{H}_{i,j}))^T\}, \textnormal{Im}\{(\textnormal{vec}(\bm{H}_{i,j}))^T\}, \textnormal{Re}\{\bm{x}_{i,j}^T\},\textnormal{Im}\{\bm{x}_{i,j}^T\}]^T\\\in\mathbb{R}^{2n(m+1)}$ for $\bm{H}_{i,j}\sim\mathcal{CN}(\bm{0}, \bm{I}_m)$ and $\bm{x}_{i,j}$ comprising $n$ QPSK symbols, $\bm{d}_{i,j}=[\textnormal{Re}\{(\bm{H}_{i,j}\bm{x}_{i,j})^T\}, \textnormal{Im}\{(\bm{H}_{i,j}\bm{x}_{i,j})^T\}]^T \in\mathbb{R}^{2m}$, and $j\in[4]$.

Since the realizations of $X\sim\mathcal{N}(0,1)$ fall within $[-3,3]$ with a probability equals to 0.9973\footnote{If $x_i$ is the realization of $X$, $\mathbb{P}(x_i\in[-3,3])=p=F(3)-F(-3)$ for $F(x)$ being the CDF (cumulative distribution function) of the standard normal distribution. In MATLAB$^{\textregistered}$, $p=\texttt{normcdf}(3)-\texttt{normcdf}(-3)=0.9973$.} and a QPSK symbol is drawn from $\{-1/\sqrt{2}, 1/\sqrt{2}\}$, all elements of $\bm{c}_i$ and $\bm{c}_{i,j}$ fall -- with high probability -- within $ [-3/\sqrt{2}, 3/\sqrt{2}]$. Consequently, Corollary \ref{Coro_complex_upper_approximation_bound} can be exploited to characterize the approximating FNN ($\bm{\Phi}_{D, \varepsilon}^{\Cross_c}$) -- with an MSE bound equals to $\varepsilon^2$ -- for the aforementioned complex product $\bm{Hx}$ over $[-D,D]=[-3,3]$. Thus, Corollary \ref{Coro_complex_upper_approximation_bound} predicts w.r.t. $(m, n, C, \varepsilon)=(8, 4, 1, 2^{-5})$ that $\mathcal{L}( \bm{\Phi}_{D, \varepsilon}^{\Cross_c} )\leq 13$, $\mathcal{W}( \bm{\Phi}_{D, \varepsilon}^{\Cross_c})\leq 1536$, $\mathcal{B}( \bm{\Phi}_{D, \varepsilon}^{\Cross_c} )\leq 18$, and an approximation MSE bound equals to $2^{-10}$. Guided by these parameters, we successfully trained a depth-10 ReLU FNN $\bm{\Phi}_{D, \varepsilon}^{\Cross_c}$ with $N_0=72$, $N_1=1536$, $N_2=\ldots=N_9=384$, and $N_{10}=16$, as demonstrated on the left sides of Figs. \ref{fig: ReLU_FNN_training_loss_and_testing_error_random_product} and \ref{fig: ReLU_FNN_training_loss_and_testing_error_random_product2}. Figs. \ref{fig: ReLU_FNN_training_loss_and_testing_error_random_product} and \ref{fig: ReLU_FNN_training_loss_and_testing_error_random_product2} -- on their left sides -- show the lowest MSE training losses obtained at the 200-th epoch and, respectively, equal to $8.88643217\times 10^{-6}$ and $7.22278731\times 10^{-6}$. Exhibiting these MSE training losses, the trained deep ReLU FNNs were tested with four different size testing datasets as shown, respectively, on the right sides of Figs. \ref{fig: ReLU_FNN_training_loss_and_testing_error_random_product} and \ref{fig: ReLU_FNN_training_loss_and_testing_error_random_product2}. These plots demonstrate that the approximation MSE exhibited by the trained deep ReLU FNNs is less than $\varepsilon^2=2^{-10}$. This substantiates Corollary \ref{Coro_complex_upper_approximation_bound}.

\section{Applications}
\label{sec: applications}
As validated in Sec. \ref{sec: comp_experiments}, our developed theory on approximating FNNs for real and complex matrix-vector products serves to guide and ease the training of teacher deep ReLU FNNs which can, in turn, be employed to train lightweight student FNNs in the context of teacher-student AI/ML paradigms. Toward this end, the underneath applications follow.
\subsubsection{Wireless Communications and Signal Processing Applications} 
Corollary \ref{Coro_complex_upper_approximation_bound} can guide the training of teacher deep ReLU FNNs $\bm{\Phi}_{D, \varepsilon}^{\Cross_c}$ which will be deployed to train lightweight student ReLU FNNs using hint training \cite{Romero_FitNets'15} or attentive hint training \cite{Saputra2019_KD_Regressor} for blind detection of a MIMO received signal; blind estimation of a doubly selective (MIMO-)OFDM channel; blind detection of a (MIMO-)OFDM signal received through a doubly selective (MIMO-)OFDM channel; blind detection of a MC-NOMA received signal and a MC-CDMA received signal; blind estimation of a doubly selective MC-CDMA channel; and blind estimation of a triply-selective MIMO channel \cite{C_Xiao_TS_MIMO_Ch_04}.     
\subsubsection{Network Science and Graph Signal Processing Applications}
Theorem \ref{Theorem_upper_approximation_bound} can guide the training of teacher deep ReLU FNNs $\bm{\Phi}_{D, \varepsilon}^{\Cross}$ which will be used to train lightweight student ReLU FNNs using hint training \cite{Romero_FitNets'15} or attentive hint training \cite{Saputra2019_KD_Regressor} for joint inference of signals and graphs; non-parametric regression for graph-aware signal reconstruction; identification of directed graph topologies and tracking of dynamic networks \cite{Giannakis_Topo_ident_18,Shen_Ten_Decomp_17}; and graph Fourier transforms \cite{Ortega_GSP_18}. 
\subsubsection{Network Neuroscience and Brain Physics Applications}
being one of the various interdisciplinary fields that constitute brain physics \cite{Lynn_Bra_Phy_Nat_2019}, network neuroscience aims to build a fundamental mechanistic understanding of how neuron level processes contribute to the structure and function of large-scale circuits, brain systems, and whole-brain structure and function \cite{Bassett_Net_NS_17}. It also inquires about perception-action coupling, brain-behavior interactions, and social networks \cite{Bassett_Net_NS_17,Brin_Physics_Lynn_2019}. 

To model brain network structure, an adjacency matrix is constructed from experimental data specifying the physical interconnections between neurons or brain regions \cite[Fig. 1, p. 8]{Brin_Physics_Lynn_2019}\cite{Fornito_Brain_Nets_Analysis_16}. To model brain network function, a similarity matrix is made from experimental data on the activity of neurons or brain regions; e.g., fMRI (functional magnetic resonance imaging) data of blood oxygen level in different parts of the brain \cite[Fig. 2, p. 17]{Brin_Physics_Lynn_2019}. If a similarity or adjacency matrix is denoted by $\bm{W}\in\mathbb{R}^{m\times m}$, $\textnormal{diag}(\bm{W}, \ldots, \bm{W})\textnormal{vec}(\bm{I}_m)$ -- for $\textnormal{diag}(\bm{W}, \ldots, \bm{W})\in\mathbb{R}^{m^2\times m^2}$ -- can be approximated using a deep ReLU FNN $\bm{\Phi}_{D, \varepsilon}^{\Cross}$, as asserted by Theorem \ref{Theorem_upper_approximation_bound}. Theorem \ref{Theorem_upper_approximation_bound} can, thus, guide the training of teacher deep ReLU FNNs $\bm{\Phi}_{D, \varepsilon}^{\Cross}$ which will be used to train -- using attentive hint training \cite{Saputra2019_KD_Regressor} -- lightweight student ReLU FNNs. The trained student ReLU FNNs can then be concatenated and trained with AEs for an efficient dimensionality that may offer further insights into the brain's structural or functional networks.

\section{Concluding Remarks and Research Outlook}
\label{sec: Conc_rem_and_res_outlook}
Regarding a matrix-vector product approximation with deep ReLU FNNs, we derived error bounds -- in Lebesgue and Sobolev norms -- that constitute our developed deep approximation theory. Guided by this theory, we successfully trained deep ReLU FNNs whose test outcomes justify the validity of our deep approximation theory. This deep approximation theory is also important for guiding and easing the training of teacher deep ReLU FNNs in consideration of the emerging teacher-student AI/ML paradigms that are essential for solving several AI/ML problems in wireless communications and signal processing; network science and graph signal processing; and network neuroscience and brain physics.

In the context of ToDL, this paper inspires numerous lines of research including high-dimensional signal processing that might be pursued through the lenses of high-dimensional probability \cite{vershynin_2018}, high-dimensional statistics \cite{wainwright_2019}, and empirical process theory \cite{Ben_Process_Theory_2018}; deep ReLU FNNs based tighter error bounds for a matrix-vector product approximation that might be derived using \textit{sparse grids} (following the lead of \cite{Hadrien_New_Error_Bounds_19}); and error bounds for a matrix-matrix, a tensor-matrix, and a tensor-tensor product approximation -- applicable in interference excision and channel estimation \cite{Getu_dissertation_19}, and brain imaging \cite{HBLPRFI15} -- with deep ReLU FNNs. Furthermore, this paper inspires much more applied research on the novel training of lightweight student models -- critical for resource-constrained environments -- given the emerging teacher-student AI/ML paradigms.

\appendices

\section{Proofs of Known Results Used}
For the sake of clarity, completeness, and rigor, we hereby provide our proofs of the known results that informed our theoretical developments in Appendices \ref{proof_Theorem_upper_approximation_bound}, \ref{proof_Coro_complex_upper_approximation_bound}, and \ref{proof_Theorem_Sobolev_upper_approximation_bound}. 

\subsection{Proof of Lemma \ref{lem: NN_superposition_parallelization}}
\label{proof_lem: NN_superposition_parallelization}
The proof of Lemma \ref{lem: NN_superposition_parallelization} was highlighted in \cite[p. 6]{grohs2019deep} through the proof of \cite[Lemma II.6]{grohs2019deep}. We herein provide our proof.

Considering $K=\max_i K_i$ is the depth of the deepest FNN, we first make all FNNs $\bm{\Phi}_i\in \bm{\mathcal{NN}}^{d_i, 1}_{K_i, M_i, \rho}$ to be of depth $K$. This can be achieved via the concatenation of I-FNN $\bm{\Phi}_{K-K_i+1}^{\bm{I}_1}$ and $\bm{\Phi}_i$ when $K_i\neq K$, where it follows through Definition \ref{def: NN_eq_definition} and Lemma \ref{lem: NN_identity_matrix} that $\bm{\Phi}_i=\big[ [\bm{W}_1^i, \bm{b}_1^i], \ldots, [\bm{W}_{K_i}^i, \bm{b}_{K_i}^i] \big]$ and $\bm{\Phi}_{K-K_i+1}^{\bm{I}_1}=\big[[[1, -1]^T, \bm{0}], \overbrace{[\bm{I}_{2}, \bm{0}],\ldots, [\bm{I}_{2}, \bm{0}]}^{(K-K_i-1)\hspace{1mm}\textnormal{times}}, [[1, -1], \bm{0}]\big]$. Employing these expressions in (\ref{eq_NN_concatenation}) for $K-K_i-1>0$ gives
\begin{multline}
\label{Two_NN_concatenation_1}
\bm{\Phi}'_i\equiv\bm{\Phi}_{K-K_i+1}^{\bm{I}_1}\bullet\bm{\Phi}_i=\big[[\bm{W}_1^{i}, \bm{b}_1^{i}],\ldots, [\bm{W}_{K_i-1}^{i}, \bm{b}_{K_i-1}^{i} ], \\ [\widetilde{\bm{W}}_{K_i}^i,       \widetilde{\bm{b}}_{K_i}^i ],  \overbrace{[\bm{I}_{2}, \bm{0}],\ldots, [\bm{I}_{2}, \bm{0}]}^{(K-K_i-1)\hspace{1mm}\textnormal{times}}, [[1, -1], \bm{0}] \big],  
\end{multline}
where $\bm{\Phi}'_i(\bm{x})=(\bm{\Phi}_{K-K_i+1}^{\bm{I}_1}\bullet\bm{\Phi}_i)(\bm{x})=\bm{\Phi}_i(\bm{x})$, $\widetilde{\bm{W}}_{K_i}^i=[(\bm{W}_{K_i}^i)^T, -(\bm{W}_{K_i}^i)^T]^T$, and $\widetilde{\bm{b}}_{K_i}^i=[(\bm{b}_{K_i}^i)^T, -(\bm{b}_{K_i}^i)^T]^T$. For $K-K_i-1=0$, (\ref{Two_NN_concatenation_1}) leads to $\bm{\Phi}'_i\equiv\bm{\Phi}_{K-K_i+1}^{\bm{I}_1}\bullet\bm{\Phi}_i$:   
\begin{equation}
\label{Two_NN_concatenation_2}
\bm{\Phi}'_i=\big[[\bm{W}_1^{i}, \bm{b}_1^{i}],\ldots, [\widetilde{\bm{W}}_{K_i}^i, \widetilde{\bm{b}}_{K_i}^i ], [[1, -1], \bm{0}] \big].   
\end{equation}
Meanwhile, it is deduced via (\ref{Two_NN_concatenation_1}) and (\ref{Two_NN_concatenation_2}) that $\bm{\Phi}_i(\bm{x}_i)=\bm{\Phi}'_i(\bm{x}_i)\in \bm{\mathcal{NN}}^{d_i, 1}_{K, M_i+\mathcal{W}(\bm{\Phi}_i)+2(K-K_i)+1, \rho}$, $i\in[n]$. Consequently,  $\bm{\Phi}^1(\bm{x})=[a_1\bm{\Phi}'_1(\bm{x}_1), \ldots, a_n\bm{\Phi}'_n(\bm{x}_n) ]^T$ and $\bm{\Phi}^2(\bm{x})=\sum_{i=1}^n a_i\bm{\Phi}'_i(\bm{x}_i)=[1, \ldots, 1]\bm{\Phi}^1(\bm{x})$ that are the parallelization and superposition of the outputs of $n$ $\bm{\Phi}'_i$s, respectively. Therefore, $\mathcal{L}(\bm{\Phi}^1)=\mathcal{L}(\bm{\Phi}^2)=K$; $\mathcal{M}(\bm{\Phi}^1)=M=\sum_{i=1}^n (M_i+\mathcal{W}(\bm{\Phi}_i)+2(K-K_i)+1)$; $\mathcal{M}(\bm{\Phi}^2)=M+n$; $\mathcal{W}(\bm{\Phi}^1)=\mathcal{W}(\bm{\Phi}^2)\leq \sum_{i=1}^n \max\{2,\mathcal{W}(\bm{\Phi}_i)\}$; $\bm{\Phi}^1\in\bm{\mathcal{NN}}^{d, n}_{K, M, \rho}$; and $\bm{\Phi}^2\in\bm{\mathcal{NN}}^{d, 1}_{K, M+n, \rho}$. Since $\bm{\Phi}^1$ and $\bm{\Phi}^2$ are constructed from $\bm{\Phi}'_i$s defined via (\ref{Two_NN_concatenation_1}) or (\ref{Two_NN_concatenation_2}), they are made up of the weights of $\bm{\Phi}_i$, $i\in[n]$, $\pm1$'s, and $\{a_1, \ldots, a_n\}$.   \QEDclosed 

\subsection{Proof of Proposition \ref{Sob_Proposition_product_NN}}
\label{proof_Sob_Proposition_product_NN}
A similar proof is provided in \cite[p. 31-33]{IGP_error_bounds_19}. We present our proof which our developments of Appendix \ref{proof_Theorem_Sobolev_upper_approximation_bound} are based upon.  

W.r.t. the identity $wx=\frac{1}{2}\big( (w+x)^2-w^2-x^2\big)$ for $x,y\in\mathbb{R}$, $wx=4D^2\big(w/2D\big) \big(x/2D\big)$ is also equated as 
	\begin{equation}
	\label{Sob_polarization_identity_3}
	wx=2D^2\Big( \Big( \frac{|w+x|}{2D}\Big)^2-\Big( \frac{|w|}{2D}\Big)^2-\Big( \frac{|x|}{2D}\Big)^2\Big).	
	\end{equation}
	In line with (\ref{Sob_polarization_identity_3}) and the identity $|x|=\rho(x)+\rho(-x)$, \cite[Proposition 3]{Yarotsky_error_bounds_17} asserts that $\bm{\Phi}_{D, \varepsilon}^{\tilde{\times}(w,x)}$ can be expressed in terms of a square ReLU FNN $\bm{\Phi}_{\delta}^{\textnormal{sq}}$ which is characterized in Proposition \ref{Sob_prop_approx_x^2}. Consequently, 
	\begin{equation}
	\label{Sob_product_NN_expression}
	\bm{\Phi}_{D, \varepsilon}^{\tilde{\times}(w,x)}(w,x)=2D^2\big( \bm{\Phi}_{\delta}^{\textnormal{sq}}(\gamma|w+x| )-\bm{\Phi}_{\delta}^{\textnormal{sq}}(\gamma|w|)-\bm{\Phi}_{\delta}^{\textnormal{sq}}(\gamma|x|)\big), 
	\end{equation} 
	where $\gamma=\frac{1}{2D}$ and it follows through (\ref{Sob_approx_bound_x^2}) that $\| \bm{\Phi}_{\delta}^{\textnormal{sq}}(x)-x^2 \|_{\bm{\mathcal{W}}^{1,\infty}((0,1); dx)}\leq \delta$.  

Meanwhile, implementing $|w+x|$, $|w|$, and $|x|$ requires 3 additional layers. Hence, employing this fact and \cite[Proposition 3]{Yarotsky_error_bounds_17}, $\mathcal{L}\big( \bm{\Phi}_{D, \varepsilon}^{\tilde{\times}(w,x)}  \big)\leq C_0\log_2(\varepsilon^{-1})+C_0\log_2(6D^2)+3\leq C_0\log_2(\varepsilon^{-1})+C_1$, where $C_0$ is a constant that emanates from the depth constraint of Proposition \ref{Sob_prop_approx_x^2}. As there would be 9 additional neurons in the implementations of the absolute value expressions of (\ref{Sob_product_NN_expression}), \cite[Proposition 3]{Yarotsky_error_bounds_17} also asserts that $\mathcal{N}\big( \bm{\Phi}_{D, \varepsilon}^{\tilde{\times}(w,x)}  \big)\leq 3C_0\log_2(\varepsilon^{-1})+3C_0\log_2(6D^2)+9\leq C'\log_2(\varepsilon^{-1})+C_2$. Moreover, since there would be 17 additional non-zero weights to implement $|w+x|$, $|w|$, and $|x|$ using the aforementioned additional three layers, $\mathcal{M}\big( \bm{\Phi}_{D, \varepsilon}^{\tilde{\times}(w,x)}  \big) \leq  3C_0\log_2(\varepsilon^{-1})+3C_0\log_2(6D^2)+17 \leq C''\log_2(\varepsilon^{-1})+C_3$. As $C', C''>0$, choosing $C_1=C_1(D), C_2=C_2(D), C_3=C_3(D)>0$ satisfy (\ref{w_x_product_prop_4}). Employing (\ref{Sob_product_NN_expression}), $\bm{\Phi}_{D, \varepsilon}^{\tilde{\times}(w,x)}(0,x)=\bm{\Phi}_{D, \varepsilon}^{\tilde{\times}(w,x)}(x,0)=0$, $\forall x\in\mathbb{R}$. Thus, (\ref{w_x_product_prop_2}) is satisfied.

To continue, we define the following transformations: $t_w:(-D, D)^2\to (0,1), \hspace{1mm} (w,x)\mapsto |w|/2D$; $t_x:(-D, D)^2\to (0,1), \hspace{1mm} (w,x)\mapsto |x|/2D$; $t_{wx}:(-D, D)^2\to (0,1), \hspace{1mm} (w,x)\mapsto |w+x|/2D$; and $f: (0,1)\to\mathbb{R}, \hspace{1mm} x\mapsto x^2$. Using these transformations, (\ref{Sob_polarization_identity_3}) and (\ref{Sob_product_NN_expression}) can be, respectively, expressed as $wx=2D^2\big( f\circ t_{wx}-f\circ t_w-f\circ t_x\big)$ and $\bm{\Phi}_{D, \varepsilon}^{\tilde{\times}(w,x)}(w,x)=2D^2\big( \bm{\Phi}_{\delta}^{\textnormal{sq}}\circ t_{wx}-\bm{\Phi}_{\delta}^{\textnormal{sq}}\circ t_w-\bm{\Phi}_{\delta}^{\textnormal{sq}}\circ t_x \big)$. Using these transformations, collecting similar terms, and applying the properties -- in line with Definition \ref{Sobolev_space_def} and (\ref{Sob_norm_2}) -- of $L^{\infty}$-norm:   
	\begin{multline}
	\label{Sob_product_approx_bound_analysis_3}
	\hspace{-3mm}\| \bm{\Phi}_{D, \varepsilon}^{\tilde{\times}(w,x)}(w,x)-wx\|_{\bm{\mathcal{W}}^{1,\infty}((-D, D)^2; dwdx)}\leq  2D^2\sum_{u\in \{w,x,wx\}} \\
	\|(\bm{\Phi}_{\delta}^{\textnormal{sq}}-f)\circ t_{u}\|_{\bm{\mathcal{W}}^{1,\infty}((-D, D)^2)} \stackrel{(a)}{\leq} 2D^2C\sum_{u\in \{w,x,wx\}} \max\big\{ \\ 
	 \|\bm{\Phi}_{\delta}^{\textnormal{sq}}-f \|_{L^{\infty}((0, 1)^2)},  |\bm{\Phi}_{\delta}^{\textnormal{sq}}-f |_{\bm{\mathcal{W}}^{1,\infty}((0, 1)^2)}   |  t_{u}|_{\bm{\mathcal{W}}^{1,\infty}((-D, D)^2)}  \big\} \\ 
	 \stackrel{(b)}{\leq} 2D^2C \sum_{u\in\{w,x,wx\}} \max\big\{ \big\|\bm{\Phi}_{\delta}^{\textnormal{sq}}-f \big\|_{L^{\infty}((0, 1)^2)},    (2D)^{-1} \times \\ |\bm{\Phi}_{\delta}^{\textnormal{sq}}-f |_{\bm{\mathcal{W}}^{1,\infty}((0,1)^2)} \big\}\leq 6D^2C \|\bm{\Phi}_{\delta}^{\textnormal{sq}}-x^2 \|_{L^{\infty}((0, 1)^2)}.  
	\end{multline}
where $(a)$ follows from Corollary \ref{coro: Sobolev_function_composition} via (\ref{gof_Sob_norm}) and $(b)$ is due to Definition \ref{Sobolev_semi_norm_def} -- via (\ref{Sob_semi_norm_2}) -- that gives $|  t_{u} |_{\bm{\mathcal{W}}^{1,\infty}((-D, D)^2)}=1/2D$. Deploying Proposition \ref{prop_approx_x^2}, $\big\|\bm{\Phi}_{\delta}^{\textnormal{sq}}(x)-x^2 \big\|_{L^{\infty}((0, 1)^2)}\leq \delta$. Thus, setting $\varepsilon=6D^2C\delta$, (\ref{Sob_product_approx_bound_analysis_3}) leads to (\ref{w_x_product_prop_1}).

Using Definition \ref{Sobolev_semi_norm_def} -- via (\ref{Sob_semi_norm_2}) -- in the above-mentioned expression of $\bm{\Phi}_{D, \varepsilon}^{\tilde{\times}(w,x)}(w,x)$ and applying (\ref{gof_Sob_semi_norm}),   
	\begin{multline}
	\label{Sob_semi_norm_two_variables_1}
	|\bm{\Phi}_{D, \varepsilon}^{\tilde{\times}(w,x)}|_{\bm{\mathcal{W}}^{1,\infty}((-D, D)^2)} \leq   
	2D^2C \sum_{u\in \{w, x, wx\} }    \\   |  \bm{\Phi}_{\delta}^{\textnormal{sq}}(x) |_{\bm{\mathcal{W}}^{1,\infty}((0, 1))}   | t_{u}  |_{\bm{\mathcal{W}}^{1,\infty}((-D, D)^2)}   \stackrel{(a)}{\leq} 3CC_4D, 
	\end{multline}
	where $(a)$ follows from (\ref{Sob_semi_norm_bound_x^2}) and $| t_{u} |_{\bm{\mathcal{W}}^{1,\infty}((-D, D)^2)}=1/2D$. Choosing $\bar{C}=3CC_4>0$, (\ref{Sob_semi_norm_two_variables_1}) asserts that $|\bm{\Phi}_{D, \varepsilon}^{\tilde{\times}(w,x)}|_{\bm{\mathcal{W}}^{1,\infty}((-D, D)^2)}\leq\bar{C}D$. This proves (\ref{w_x_product_prop_3}).   \QEDclosed

\subsection{Proof of Proposition \ref{Proposition_product_NN}}
\label{proof_Proposition_product_NN}
A similar proof is available in \cite[p. 8-9]{grohs2019deep}. We henceforth provide our proof which is going to exploit Lemma \ref{composition_norm}. 

W.r.t. (\ref{Sob_polarization_identity_3}) and the identity $|x|=\rho(x)+\rho(-x)$, \cite[Proposition 3]{Yarotsky_error_bounds_17} guarantees that $\bm{\Phi}_{D, \varepsilon}^{\tilde{\times}(w,x)}$ can be expressed via the square FNN $\bm{\Phi}_{\delta}^{\textnormal{sq}}$ characterized in Proposition \ref{prop_approx_x^2}. Consequently, 
\begin{equation}
\label{product_NN_expression}
\bm{\Phi}_{D, \varepsilon}^{\tilde{\times}(w,x)}(w,x)=2D^2\big( \bm{\Phi}_{\delta}^{\textnormal{sq}}( \gamma|w+x|)-\sum_{u\in \{|w|, |x| \}} \bm{\Phi}_{\delta}^{\textnormal{sq}}(\gamma u )\big),  
\end{equation}
where $\gamma=1/2D$. Using (\ref{product_NN_expression}) and Lemma \ref{lem: NN_superposition_parallelization}, the parameters of $\bm{\Phi}_{D, \varepsilon}^{\tilde{\times}(w,x)}$ can be deduced from the parameters of $\bm{\Phi}_{\delta}^{\textnormal{sq}}$ stated in Proposition \ref{prop_approx_x^2}. Thus, $\mathcal{W}\big(\bm{\Phi}_{D, \varepsilon}^{\tilde{\times}(w,x)}\big)\leq\sum_{i=1}^3 \max\big\{2,4 \big\}=12$; $\mathcal{B}\big(\bm{\Phi}_{D, \varepsilon}^{\tilde{\times}(w,x)}\big)\leq \max\{4, 2D^2 \}$, as (\ref{product_NN_expression}) can be implemented by multiplying the square FNNs' outputs by $2D^2$ and adding; and $\mathcal{L}\big(\bm{\Phi}_{D, \varepsilon}^{\tilde{\times}(w,x)}\big)\leq C'\log_2 (\delta^{-1})$, as asserted by Proposition \ref{prop_approx_x^2}. It also follows directly from (\ref{product_NN_expression}) that $\bm{\Phi}_{D, \varepsilon}^{\tilde{\times}(w,x)}(0,x)=\bm{\Phi}_{D, \varepsilon}^{\tilde{\times}(w,x)}(x,0)=0$, $\forall x\in\mathbb{R}$. 

Meanwhile, we define the following transformations: $t_w:[-D, D]^2\to [0,1], \hspace{1mm} (w,x)\mapsto |w|/2D$; $t_x:[-D, D]^2\to [0,1], \hspace{1mm} (w,x)\mapsto |x|/2D$; $t_{wx}:[-D, D]^2\to [0,1], \hspace{1mm} (w,x)\mapsto |w+x|/2D$; and $f: [0,1]\to\mathbb{R}, \hspace{1mm} x\mapsto x^2$. Deploying these transformations, (\ref{Sob_polarization_identity_3}) and (\ref{product_NN_expression}) can be, respectively, equated as $wx=2D^2\big( f\circ t_{wx}-f\circ t_w-f\circ t_x\big)$ and $\bm{\Phi}_{D, \varepsilon}^{\tilde{\times}(w,x)}(w,x)=2D^2\big( \bm{\Phi}_{\delta}^{\textnormal{sq}}\circ t_{wx}-\bm{\Phi}_{\delta}^{\textnormal{sq}}\circ t_w-\bm{\Phi}_{\delta}^{\textnormal{sq}}\circ t_x  \big)$. Employing these relations in the left-hand side of (\ref{product_approx_bound}), collecting similar terms, and applying the properties of $L^{\infty}$-norm produce the expression   
\begin{multline}
\label{product_approx_bound_analysis_3}
\| \bm{\Phi}_{D, \varepsilon}^{\tilde{\times}(w,x)}(w,x)-wx\|_{L^{\infty}([-D, D]^2)}\leq  \\
2D^2\sum_{u\in\{w, x, wx \}}\|\big(\bm{\Phi}_{\delta}^{\textnormal{sq}}-f\big)\circ t_{u}\|_{L^{\infty}([-D, D]^2)} \\
\stackrel{(a)}{\leq} 2D^2\sum_{u\in\{w, x, wx \}} \|\bm{\Phi}_{\delta}^{\textnormal{sq}}(x)-x^2 \|_{L^{\infty}( t_{u}([-D, D]^2))}, 
\end{multline}
where $(a)$ follows from Lemma \ref{composition_norm} through (\ref{Compositon_L_p_norm_2}). From the above-defined transformations, $t_{u}([-D, D]^2)=[0,1]$. Consequently, 
\begin{equation}
\label{product_approx_bound_analysis_5}
\frac{\| \bm{\Phi}_{D, \varepsilon}^{\tilde{\times}(w,x)}(w,x)-wx\|_{L^{\infty}([-D, D]^2)}}{6D^2}\leq \|\bm{\Phi}_{\delta}^{\textnormal{sq}}(x)-x^2 \|_{L^{\infty}([0, 1])}. 
\end{equation}  
Using Proposition \ref{prop_approx_x^2}, $\| \bm{\Phi}_{\delta}^{\textnormal{sq}}(x)-x^2   \|_{L^{\infty}([0,1])}\leq \delta$. Accordingly, 
\begin{equation}
\label{product_approx_bound_analysis_6}
\| \bm{\Phi}_{D, \varepsilon}^{\tilde{\times}(w,x)}(w,x)-wx\|_{L^{\infty}([-D, D]^2)}\leq \varepsilon, 
\end{equation}
where we chose $\varepsilon=6D^2\delta$. This leads to (\ref{product_approx_bound}) and the depth constraint $\mathcal{L}\big(\bm{\Phi}_{D, \varepsilon}^{\tilde{\times}(w,x)}\big)\leq C'\log_2(\delta^{-1})\leq C\log_2 (D^2 \varepsilon^{-1})$, where $C$ is chosen as $C>>C'\log_2 6$. \QEDclosed

\section{Proof of Proposition \ref{prop: proposition_1}}
\label{proof_prop: proposition_1}
Expressed through a ReLU activation function applied component-wise, $\bm{x}=\rho(\bm{x})-\rho(-\bm{x})$ \cite{grohs2019deep}. Exploiting this identity for $\bm{A}=\bm{W}\bm{x}$, $\bm{A}=\bm{I}_{m}\rho(\bm{A})-\bm{I}_{m}\rho(-\bm{A})$. Accordingly, $\bm{A}=\bm{I}_{m}\rho(\bm{W}\bm{x})-\bm{I}_{m}\rho(-\bm{W}\bm{x})$. 

Realizing that $\rho(\bm{W}\bm{x})$ is a transformation with $\bm{x}$ as the input of a ReLU FNN, $\bm{W}$ as the weight matrix of a ReLU FNN, and $\rho(\cdot)$ as an element-wise ReLU activation, $\rho(\bm{W}\bm{x})$ is represented by $\bm{\mathcal{NN}}_{1, M', \rho}^{n, m}$ with $M'=\|\bm{W}\|_{\ell_0}$. W.r.t. this 1-layer ReLU FNN, $\bm{I}_{m}\rho(\bm{W}\bm{x})$ can be realized as adding a linear operation on every corresponding output, i.e., multiplying every respective output by 1 w.r.t. an additional layer having no activation function. Hence, $\bm{I}_{m}\rho(\bm{W}\bm{x})$ can be represented by $\bm{\mathcal{NN}}_{2, M_1, \rho}^{n, m}$ with $M_1=\|\bm{W}\|_{\ell_0}+m$ and no activation at the output layer. By the same token, $-\bm{I}_{m}\rho(-\bm{W}\bm{x})$ can also be represented by $\bm{\mathcal{NN}}_{2, M_2, \rho}^{n, m}$ with $M_2=\|-\bm{W}\|_{\ell_0}+m=\|\bm{W}\|_{\ell_0}+m$ and also without any activation function at the output layer. Similar to the concepts of \cite[Lemma D.2]{Arora_ICLR_18}, an addition operation using two FNNs can be conceived as putting two ReLU FNNs parallelly and adding the outputs of the corresponding output neurons. Doing so results in a 2-layer ReLU FNN with a hidden layer double the size of the ReLU FNNs representing the summands. As a result, $\bm{A}=\bm{I}_{m}\rho(\bm{W}\bm{x})-\bm{I}_{m}\rho(-\bm{W}\bm{x})$ can be represented by $\bm{\mathcal{NN}}_{2, M, \rho}^{n, m}$ with $M=2\|\bm{W}\|_{\ell_0}+2m$ and without any activation at the output layer. This completes the proof of $1)$.   

Utilizing the result of $1)$, let $\bm{\Phi}^1\eqdef \big[[[\bm{W}^T, -\bm{W}^T]^T, \bm{0}], [[\bm{I}_{m}, -\bm{I}_{m}], \bm{0}] \big]\in\bm{\mathcal{NN}}_{2, M, \rho}^{n, m}$ and $\bm{\Phi}^2\eqdef\bm{\Phi}_{K-1}^{\bm{I}_{n}}$, an I-FNN defined via (\ref{NN_Id_1}). As $\bm{\Phi}_{K-1}^{\bm{I}_{n}}(\bm{x})=\bm{x}$, $\bm{A}=\bm{A}(\bm{\Phi}_{K-1}^{\bm{I}_{n}})=\bm{\Phi}^1(\bm{\Phi}_{K-1}^{\bm{I}_{n}})=\bm{\Phi}^1\bullet\bm{\Phi}_{K-1}^{\bm{I}_{n}}$, the concatenation of $\bm{\Phi}^1$ and $\bm{\Phi}_{K-1}^{\bm{I}_{n}}$. To this end, deploying (\ref{eq_NN_concatenation}) and (\ref{NN_Id_1}) leads to             
\begin{multline}
\label{concatenated_NN_1}
\bm{\Phi}^1\bullet\bm{\Phi}_{K-1}^{\bm{I}_{n}}=\big[[[\bm{I}_{n}, -\bm{I}_{n}]^T, \bm{0}], \overbrace{[\bm{I}_{2n}, \bm{0}],\ldots, [\bm{I}_{2n}, \bm{0}]}^{(K-1)-2\hspace{1mm}\textnormal{times}}, \\
[ [\bm{B}_1, \bm{B}_2], \bm{0}], [[\bm{I}_{m}, -\bm{I}_{m}], \bm{0}] \big], 
\end{multline}
where $\bm{B}_1=[\bm{W}^T, -\bm{W}^T]^T$ and $\bm{B}_2=[-\bm{W}^T, \bm{W}^T]^T$. 
Hence, $\mathcal{L}(\bm{\Phi}^1\bullet\bm{\Phi}_{K-1}^{\bm{I}_{n}})=K$ and $\mathcal{M}(\bm{\Phi}^1\bullet\bm{\Phi}_{K-1}^{\bm{I}_{n}})=2m+2(K-2)n+4\|\bm{W}\|_{\ell_0}$. Accordingly, $\bm{A}$ can also be represented by $\bm{\mathcal{NN}}_{K, M, \rho}^{n, m}$ with $M=2m+2(K-2)n+4\|\bm{W}\|_{\ell_0}$ and no activation at the output layer. This completes the proof of $2)$.   

Similarly, if we let $\bm{\Phi}^1\eqdef\bm{\Phi}_{K-1}^{\bm{I}_{m}}$ -- an I-FNN defined via (\ref{NN_Id_1}) -- and $\bm{\Phi}^2\eqdef \big[[[\bm{W}^T, -\bm{W}^T]^T, \bm{0}], [[\bm{I}_{m}, -\bm{I}_{m}], \bm{0}] \big]\in\bm{\mathcal{NN}}_{2, M, \rho}^{n, m}$, then $\bm{A}=\bm{\Phi}_{K-1}^{\bm{I}_{m}}(\bm{A})=\bm{\Phi}_{K-1}^{\bm{I}_{m}}(\bm{\Phi}^2)=\bm{\Phi}_{K-1}^{\bm{I}_{m}}\bullet\bm{\Phi}^2$, the concatenation of $\bm{\Phi}_{K-1}^{\bm{I}_{m}}$ and $\bm{\Phi}^2$. Consequently, employing (\ref{eq_NN_concatenation}) and (\ref{NN_Id_1}) gives          
\begin{multline}
\label{concatenated_NN_2}
\bm{\Phi}_{K-1}^{\bm{I}_m}\bullet\bm{\Phi}^2=\big[ [ [\bm{W}^T, -\bm{W}^T ]^T, \bm{0}],  [ [\bm{I}(\bm{A}_1), \bm{I}(\bm{A}_2)], \bm{0} ],  \\
\overbrace{[\bm{I}_{2m}, \bm{0}],\ldots, [\bm{I}_{2m}, \bm{0}]}^{(K-1)-2\hspace{1mm}\textnormal{times}}, [ [\bm{I}_{m}, -\bm{I}_{m}], \bm{0} ] \big], 
\end{multline}
where $\bm{I}(\bm{A}_1)=[\bm{I}_{m}, -\bm{I}_{m}]^T$, $\bm{I}(\bm{A}_2)=[-\bm{I}_{m}, \bm{I}_{m}]^T$. To this end, $\mathcal{L}(\bm{\Phi}_{K-1}^{\bm{I}_{m}}\bullet\bm{\Phi}^2)=K$ and $\mathcal{M}(\bm{\Phi}_{K-1}^{\bm{I}_{m}}\bullet\bm{\Phi}^2)=2Km+2\|\bm{W}\|_{\ell_0}$. Therefore, $\bm{A}$ can also be represented by $\bm{\mathcal{NN}}_{K, M, \rho}^{n, m}$ with $M=2Km+2\|\bm{W}\|_{\ell_0}$ and no output activation. This completes the proof of $3)$.   \QEDclosed

\section{Proof of Theorem \ref{Theorem_upper_approximation_bound}}
\label{proof_Theorem_upper_approximation_bound}
We start by stating the following proposition. 
\begin{proposition}
	\label{Proposition_product_NN}
	For all $\varepsilon \in (0, 1/2)$, $D\in \mathbb{R}_{+}$, and $\rho(x)\eqdef \textnormal{max}(x,0)$, there exists a constant $C>0$ such that there is a FNN $\bm{\Phi}_{D, \varepsilon}^{\tilde{\times}(w,x)}\in\bm{\mathcal{NN}}_{\infty, \infty, \rho}^{2,1}$ satisfying $\mathcal{L}(\bm{\Phi}_{D, \varepsilon}^{\tilde{\times}(w,x)})\leq C\log_2 (D^2 \varepsilon^{-1})$, $\mathcal{W}(\bm{\Phi}_{D, \varepsilon}^{\tilde{\times}(w,x)})\leq 12$, $\mathcal{B}(\bm{\Phi}_{D, \varepsilon}^{\tilde{\times}(w,x)})\leq \max\{4, 2D^2 \}$, $\bm{\Phi}_{D, \varepsilon}^{\tilde{\times}(w,x)}(0,x)=\bm{\Phi}_{D, \varepsilon}^{\tilde{\times}(w,x)}(x,0)=0$, $\forall x\in\mathbb{R}$, and     
	\begin{equation}
	\label{product_approx_bound}
	\| \bm{\Phi}_{D, \varepsilon}^{\tilde{\times}(w,x)}(w,x)-wx\|_{L^{\infty}([-D, D]^2)}\leq \varepsilon.   
	\end{equation}
	\proof The proof is relegated to Appendix \ref{proof_Proposition_product_NN}.  
\end{proposition}
 
To proceed further, we prove the following proposition.    
\begin{proposition}
	\label{Proposition_N_product_and_sum_NN}
	Let $\bm{w}^T=[w_1, \ldots, w_{n}]^T\in\mathbb{R}^{n}$ and $\bm{x}=[x_1, \ldots, x_{n}]^T\in\mathbb{R}^{n}$. For all $\varepsilon \in (0, 1/2)$, $D\in \mathbb{R}_{+}$, and $\rho(x)\eqdef \textnormal{max}(x,0)$, there exists a constant $C>0$ such that there is a FNN $\bm{\Phi}_{D, \varepsilon}^{\tilde{\times}(\bm{w},\bm{x})}\in\bm{\mathcal{NN}}_{\infty, \infty, \rho}^{2n,1}$ satisfying $\mathcal{L}(\bm{\Phi}_{D, \varepsilon}^{\tilde{\times}(\bm{w},\bm{x})})\leq C\log_2 (nD^2 \varepsilon^{-1})$, $\mathcal{W}(\bm{\Phi}_{D, \varepsilon}^{\tilde{\times}(\bm{w},\bm{x})})\leq 12n$, $\mathcal{B}(\bm{\Phi}_{D, \varepsilon}^{\tilde{\times}(\bm{w},\bm{x})})\leq \max\{4, 2D^2 \}$, $\bm{\Phi}_{D, \varepsilon}^{\tilde{\times}(\bm{w},\bm{x})}(\bm{0},\bm{x})=\bm{\Phi}_{D, \varepsilon}^{\tilde{\times}(\bm{w},\bm{x})}(\bm{x},\bm{0})=0$, $\forall \bm{x}\in\mathbb{R}^n$, and     
	\begin{equation}
	\label{N_product_and_sum_approx_NNbound}
	\| \bm{\Phi}_{D, \varepsilon}^{\tilde{\times}(\bm{w},\bm{x})}(\bm{w},\bm{x})-\bm{w}\bm{x} \|_{L^{\infty}([-D, D]^{2n})}\leq \varepsilon.   
	\end{equation}
\end{proposition}
\proof Because $\bm{w}\bm{x}=\sum_{i=1}^n w_ix_i$, this sum of products can be realized via a ReLU FNN $\bm{\Phi}_{D, \varepsilon}^{\tilde{\times}(w,x)}$ characterized in Proposition \ref{Proposition_product_NN}. Specifically, $\bm{w}\bm{x}$ can be implemented via the superposition of $n$ $\bm{\Phi}_{D, \varepsilon}^{\tilde{\times}(w,x)}$s that are, respectively, fed with input tuples $(w_i, x_i)$, $i\in[n]$. Consequently,   
\begin{equation}
\label{N_product_and_N_sum_NN}
\bm{\Phi}_{D, \varepsilon}^{\tilde{\times}(\bm{w},\bm{x})}(\bm{w},\bm{x})=\sum_{i=1}^n \bm{\Phi}_{D, \tilde{\varepsilon}}^{\tilde{\times}(w,x)}(w_i,x_i),  
\end{equation}
where $\| \bm{\Phi}_{D, \tilde{\varepsilon}}^{\tilde{\times}(w,x)}(w_i,x_i)-w_ix_i\|_{L^{\infty}([-D, D]^2)}\leq \tilde{\varepsilon}, \hspace{1mm}\forall i$, $\mathcal{L}( \bm{\Phi}_{D, \tilde{\varepsilon}}^{\tilde{\times}(w_i,x_i)})\leq C\log_2 (D^2 \tilde{\varepsilon}^{-1})$, $\mathcal{W}(\bm{\Phi}_{D, \tilde{\varepsilon}}^{\tilde{\times}(w_i,x_i)})\leq 12$, and $\mathcal{B}(\bm{\Phi}_{D, \tilde{\varepsilon}}^{\tilde{\times}(w_i,x_i)})\leq \max\{4, 2D^2 \}$. Moreover, $\bm{\Phi}_{D, \tilde{\varepsilon}}^{\tilde{\times}(w_i,x_i)}(0,x_i)=\bm{\Phi}_{D, \tilde{\varepsilon}}^{\tilde{\times}(w_i,x_i)}(x_i,0)=0$, $\forall x_i\in\mathbb{R}$, and $i\in[n]$. Deploying these FNN constraints, we can infer from (\ref{N_product_and_N_sum_NN}) that $\bm{\Phi}_{D, \varepsilon}^{\tilde{\times}(\bm{w},\bm{x})}(\bm{0},\bm{x})=\bm{\Phi}_{D, \varepsilon}^{\tilde{\times}(\bm{w},\bm{x})}(\bm{x},\bm{0})=0$, $\forall \bm{x}\in\mathbb{R}^n$. Besides, using Lemma \ref{lem: NN_superposition_parallelization} and using the aforementioned constraints, $\mathcal{W}(\bm{\Phi}_{D, \varepsilon}^{\tilde{\times}(\bm{w},\bm{x})})\leq 12n=\sum_{i=1}^n \max\{2,12\}$ and $\mathcal{B}(\bm{\Phi}_{D, \tilde{\varepsilon}}^{\tilde{\times}(w_i,x_i)})\leq \max\{4, 2D^2 \}$, as (\ref{N_product_and_N_sum_NN}) can be implemented as a superposition of $n$ parallelized FNNs with an output weight matrix equated to $[1, 1, \ldots, 1]\in\mathbb{R}^{1\times n}$.              

W.r.t. our subsequent analyses, we define these parameters for $\mathcal{D}\eqdef \big\{(w_1,x_1), \ldots, (w_i,x_i), \ldots, (w_n,x_n) \big\}$, $i\in[n]$: 
\begin{subequations}
\begin{align}
\label{transformation_7}
t_{w_ix_i}:[-D, D]^{2n}& \to [-D,D]^2, \hspace{2mm} \mathcal{D} \mapsto (w_i,x_i)  \\ 
\label{transformation_8}
g: [-D,D]^2& \to\mathbb{R}, \hspace{1mm} (w,x)\mapsto wx.
\end{align}
\end{subequations}
Deploying (\ref{transformation_7})-(\ref{transformation_8}) in (\ref{N_product_and_N_sum_NN}) and $\bm{w}\bm{x}=\sum_{i=1}^n w_ix_i$,  
\begin{equation}
	\label{N_sum_product_NN_transformations}	
	\big(\bm{\Phi}_{D,\varepsilon}^{\tilde{\times}(\bm{w},\bm{x})}(\bm{w},\bm{x}), \bm{w}\bm{x}  \big) =\sum_{i=1}^n \big(\bm{\Phi}_{D,\tilde{\varepsilon}}^{\tilde{\times}(w,x)}, g\big)\circ t_{w_ix_i}. 
\end{equation}
Using (\ref{N_sum_product_NN_transformations}) in the left-hand side of (\ref{N_product_and_sum_approx_NNbound}), collecting similar terms, and applying the properties of $L^{\infty}$-norm:
\begin{multline}
\label{N_product_and_sum_approx_NNbound_4}
\|\bm{\Phi}_{D,\varepsilon}^{\tilde{\times}(\bm{w},\bm{x})}(\bm{w},\bm{x})-\bm{w}\bm{x}\|_{L^{\infty}([-D, D]^{2n})} \leq\sum_{i=1}^n \| ( \bm{\Phi}_{D,\tilde{\varepsilon}}^{\tilde{\times}(w,x)}-g) \\ \circ t_{w_ix_i}    \|_{L^{\infty}([-D, D]^{2n})}  
\stackrel{(a)}{\leq}   \sum_{i=1}^n \| \bm{\Phi}_{D,\tilde{\varepsilon}}^{\tilde{\times}(w,x)}-g    \|_{L^{\infty}(t_{w_ix_i}([-D, D]^{2n}))} \\ 
 \stackrel{(b)}{=} \sum_{i=1}^n \| \bm{\Phi}_{D,\tilde{\varepsilon}}^{\tilde{\times}(w,x)}(w,x)-wx    \|_{L^{\infty}([-D, D]^{2})}\stackrel{(c)}{=} \sum_{i=1}^n \tilde{\varepsilon},    
\end{multline}
where $(a)$, $(b)$, and $(c)$ follow, respectively, from Lemma \ref{composition_norm} via (\ref{Compositon_L_p_norm_2}), (\ref{transformation_7})-(\ref{transformation_8}), and (\ref{product_approx_bound}) -- w.r.t. the $ \tilde{\varepsilon}$ constraint of $\bm{\Phi}_{D,\tilde{\varepsilon}}^{\tilde{\times}(w,x)}$. Realizing the superposition of $n$ parallelized FNNs per Lemma \ref{lem: NN_superposition_parallelization} by choosing $\varepsilon=n\tilde{\varepsilon}$, $\mathcal{L}( \bm{\Phi}_{D,\varepsilon}^{\tilde{\times}(\bm{w},\bm{x})})\leq C\log_2 (D^2 \tilde{\varepsilon}^{-1})=C\log_2(nD^2 \varepsilon^{-1})$. This finishes the proof of Proposition \ref{Proposition_N_product_and_sum_NN}.     \QEDclosed

Employing Proposition \ref{Proposition_N_product_and_sum_NN} and Lemma \ref{lem: NN_superposition_parallelization} on the parallelization of FNNs, the proof of Theorem \ref{Theorem_upper_approximation_bound} resumes in the sequel. 

From (\ref{W_x_product}), $\bm{W}\bm{x}=[\bm{w}_1\bm{x}, \dots, \bm{w}_m\bm{x}]^T\in\mathbb{R}^{m}$. Per Proposition \ref{Proposition_N_product_and_sum_NN}, $m$ FNNs are approximating the $m$ affine transformations w.r.t. $\bm{x}$ and the weights $\bm{w}_1,  \ldots, \bm{w}_m$. Hence, it can be inferred from (\ref{N_product_and_sum_approx_NNbound}) that       
\begin{equation}
\label{N_product_and_sum_approx_NNbound_j}
\| \bm{\Phi}_{D, \varepsilon}^{\tilde{\times}(\bm{w},\bm{x})}(\bm{w}_i,\bm{x})-\bm{w}_i\bm{x} \|_{L^{\infty}([-D, D]^{2n})}\leq \varepsilon,    
\end{equation}
where $\mathcal{L}( \bm{\Phi}_{D, \varepsilon}^{\tilde{\times}(\bm{w},\bm{x})} )\leq C\log_2 (nD^2 \varepsilon^{-1})$, $\mathcal{W}(\bm{\Phi}_{D, \varepsilon}^{\tilde{\times}(\bm{w},\bm{x})})\leq 12n$, $\mathcal{B}(\bm{\Phi}_{D, \varepsilon}^{\tilde{\times}(\bm{w},\bm{x})})\leq \max\{4, 2D^2 \}$, and $\bm{\Phi}_{D, \varepsilon}^{\tilde{\times}(\bm{w},\bm{x})}(\bm{0},\bm{x})=\bm{\Phi}_{D, \varepsilon}^{\tilde{\times}(\bm{w},\bm{x})}(\bm{x},\bm{0})=0$, $\forall \bm{x}\in\mathbb{R}^{n}$. Accordingly,  
\begin{equation}
\label{paralleized_FNN_outputs_Leb}
\bm{W}\bm{x}\approx[\bm{\Phi}_{D, \varepsilon}^{\tilde{\times}(\bm{w},\bm{x})}(\bm{w}_1,\bm{x}), \dots, \bm{\Phi}_{D, \varepsilon}^{\tilde{\times}(\bm{w},\bm{x})}(\bm{w}_m,\bm{x})]^T, 
\end{equation}
where $(\bm{\Phi}_{D, \varepsilon}^{\tilde{\times}(\bm{w},\bm{x})}(\bm{w}_i,\bm{x}))^T=\bm{\Phi}_{D, \varepsilon}^{\tilde{\times}(\bm{w},\bm{x})}(\bm{w}_i,\bm{x})$, $i\in [m]$.

Per Lemma \ref{lem: NN_superposition_parallelization} on the parallelization of FNNs, there exist a FNN which exactly implements the parallelization of the FNNs -- fed with $\bm{x}_i=[\bm{w}_i,\bm{x}^T]^T$, $i\in [m]$ -- that produce the right-hand side of (\ref{paralleized_FNN_outputs_Leb}). Thus, w.r.t. the constraints of Lemma \ref{lem: NN_superposition_parallelization}, (\ref{paralleized_FNN_outputs_Leb}) can also be expressed via the parallelization of $\tilde{\bm{\Phi}}\eqdef [\bm{\Phi}_{D, \varepsilon}^{\tilde{\times}(\bm{w},\bm{x})}, \ldots, \bm{\Phi}_{D, \varepsilon}^{\tilde{\times}(\bm{w},\bm{x})}]$ for $(\bm{v}_i)\eqdef (\bm{w}_i,\bm{x})$, $i\in[m]$, as 
\begin{equation}
\label{parallelized_output_leb_1}
\mathcal{P}(\tilde{\bm{\Phi}})(\bm{W}, \bm{x})=[\bm{\Phi}_{D, \varepsilon}^{\tilde{\times}(\bm{w},\bm{x})}(\bm{v}_1), \dots, \bm{\Phi}_{D, \varepsilon}^{\tilde{\times}(\bm{w},\bm{x})}(\bm{v}_m)]^T, 
\end{equation}  
where -- by Lemma \ref{lem: NN_superposition_parallelization} -- the parallelized FNN denoted by $\mathcal{P}(\tilde{\bm{\Phi}})$ is fed with $[(\textnormal{vec}(\bm{W}))^T,\bm{x}^T]^T$, $\mathcal{L}(  \mathcal{P}(\tilde{\bm{\Phi}}))\leq C\log_2 (nD^2 \varepsilon^{-1})$, $\mathcal{W}\big(\mathcal{P}(\tilde{\bm{\Phi}})\big)\leq \sum_{i=1}^{m} 12n=12mn$, and $\mathcal{B}\big( \mathcal{P}(\tilde{\bm{\Phi}}) \big)\leq \max\{4, 2D^2 \}$. 

Because $\bm{\Phi}_{D, \varepsilon}^{\Cross}$ is also fed with $[(\textnormal{vec}(\bm{W}))^T,\bm{x}^T]^T$ to generate an approximation to $\bm{W}\bm{x}$ which is also realizable via the parallelized FNN of (\ref{parallelized_output_leb_1}), $\bm{\Phi}_{D, \varepsilon}^{\Cross}(\bm{W},\bm{x})=\mathcal{P}(\tilde{\bm{\Phi}})(\bm{W},\bm{x})$. Hence, (\ref{parallelized_output_leb_1}) corroborates w.r.t. every scalar output that 
\begin{equation}
\label{parallelized_output_leb_2}
\bm{\Phi}_{D, \varepsilon}^{\Cross}(\bm{W},\bm{x})=[\bm{\Phi}_{D, \varepsilon}^{\tilde{\times}(\bm{w},\bm{x})}(\bm{v}_1), \dots,  \bm{\Phi}_{D, \varepsilon}^{\tilde{\times}(\bm{w},\bm{x})}(\bm{v}_m)]^T,  
\end{equation}
where $\bm{\Phi}_{D, \varepsilon}^{\Cross}\big(\bm{0},\bm{x}\big)=\bm{\Phi}_{D, \varepsilon}^{\Cross}\big(\bm{W},\bm{0}\big)=\bm{0}$. Meanwhile, employing Definition \ref{Lebesgue_space_vector_valued_def} and (\ref{Lebesgue_norm_2}) in the left-hand side of (\ref{matrix_vector_product_approx_NN_bound})           
\begin{multline}
\label{Lebesgue_norm_3}
\|\bm{\Phi}_{D, \varepsilon}^{\Cross}(\bm{W},\bm{x})-\bm{W}\bm{x}\|_{L^{\infty}([-D, D]^{2n}; \mathbb{R}^{m})}= \max_{i=1, \ldots, m}  \\ 
 \| \bm{\Phi}_{D, \varepsilon}^{\tilde{\times}(\bm{w},\bm{x})}(\bm{w}_i,\bm{x})-\bm{w}_i\bm{x} \|_{L^{\infty}([-D, D]^{2n})}  
\stackrel{(a)}{\leq} \varepsilon, 
\end{multline}
where $(a)$ follows from (\ref{N_product_and_sum_approx_NNbound_j}) and the bound of (\ref{matrix_vector_product_approx_NN_bound}) is obtained. This completes the proof of (\ref{matrix_vector_product_approx_NN_bound}). Since $\bm{\Phi}_{D, \varepsilon}^{\Cross}(\bm{W},\bm{x})=\mathcal{P}(\tilde{\bm{\Phi}})(\bm{W},\bm{x})$, the aforementioned constraints of $\mathcal{P}(\tilde{\bm{\Phi}})$ are also the constraints of $\bm{\Phi}_{D, \varepsilon}^{\Cross}$ mentioned in Theorem \ref{Theorem_upper_approximation_bound}.  \QEDclosed

\section{Proof of Corollary \ref{Coro_complex_upper_approximation_bound}}
\label{proof_Coro_complex_upper_approximation_bound}
 We are going to deploy Theorem \ref{Theorem_upper_approximation_bound}, Lemma \ref{lem: NN_parallelization}, Lemma \ref{lem: NN_superposition_parallelization}, and the following identities:      %
  \begin{subequations}
  	\begin{align}
  	\label{Re_bm_W_times_bm_x}
  	\bm{p}_1\eqdef \textnormal{Re}\{\bm{W}\bm{x}\} &= \textnormal{Re}\{\bm{W}\}\textnormal{Re}\{\bm{x}\}- \textnormal{Im}\{\bm{W}\}\textnormal{Im}\{\bm{x}\}   \\
  	\label{Im_bm_W_times_bm_x}
  	\bm{p}_2\eqdef\textnormal{Im}\{\bm{W}\bm{x}\} &= \textnormal{Re}\{\bm{W}\}\textnormal{Im}\{\bm{x}\}+ \textnormal{Im}\{\bm{W}\}\textnormal{Re}\{\bm{x}\}.
  	\end{align}
  \end{subequations}
  Since $\bm{W}_1=\textnormal{Re}\{\bm{W}\}\in\mathbb{R}^{m\times n}$, $\bm{W}_2=\textnormal{Im}\{\bm{W}\}\in\mathbb{R}^{m\times n}$, $\bm{x}_1=\textnormal{Re}\{\bm{x}\}\in\mathbb{R}^{n}$, and $\bm{x}_2=\textnormal{Im}\{\bm{x}\}\in\mathbb{R}^{n}$, (\ref{Re_bm_W_times_bm_x}) and (\ref{Im_bm_W_times_bm_x}) can be approximated via the superposition of two FNNs characterized via Theorem \ref{Theorem_upper_approximation_bound}. Thus, 
  \begin{subequations}
  	\begin{align}
  	\label{Re_bm_A_times_bm_x_approx}
  	\bm{p}_1\approx &\bm{\Phi}_{D, \tilde{\varepsilon}}^{\Cross}(\bm{W}_1,\bm{x}_1)-\bm{\Phi}_{D, \tilde{\varepsilon}}^{\Cross}(\bm{W}_2,\bm{x}_2)         \\
  	\label{Im_bm_A_times_bm_x_approx}
  	\bm{p}_2\approx & \bm{\Phi}_{D, \tilde{\varepsilon}}^{\Cross}(\bm{W}_1,\bm{x}_2)+\bm{\Phi}_{D, \tilde{\varepsilon}}^{\Cross}(\bm{W}_2,\bm{x}_1).    
  	\end{align}
  \end{subequations}
  Per the superposition of FNNs stated in Lemma \ref{lem: NN_superposition_parallelization}, (\ref{Re_bm_A_times_bm_x_approx}) and (\ref{Im_bm_A_times_bm_x_approx}) can be represented via FNNs $\bm{\Phi}_{D, \tilde{\varepsilon}}^{\Cross_{s_1}}$ and $\bm{\Phi}_{D, \tilde{\varepsilon}}^{\Cross_{s_2}}$ such that 
  \begin{equation}
  \label{superposition_FNNs}
  [\bm{p}_1, \bm{p}_2] \approx [\bm{\Phi}_{D, \tilde{\varepsilon}}^{\Cross_{s_1}} (\bm{W}_{1,2}, \bm{x}_{1,2}),  \bm{\Phi}_{D, \tilde{\varepsilon}}^{\Cross_{s_2}}(\bm{W}_{1,2}, \bm{x}_{1,2})], 
  \end{equation}   
  where $\bm{W}_{1,2}=[\bm{W}_1, \bm{W}_2]$, $\bm{x}_{1,2}=[ \bm{x}_1^T, \bm{x}_2^T ]^T$, $\mathcal{L}( \bm{\Phi}_{D, \tilde{\varepsilon}}^{\Cross_{s_i}})\leq C\log_2 (nD^2 \tilde{\varepsilon}^{-1})$, $\mathcal{W}( \bm{\Phi}_{D, \tilde{\varepsilon}}^{\Cross_{s_i}})\leq 24mn$, $\mathcal{B}( \bm{\Phi}_{D, \tilde{\varepsilon}}^{\Cross_{s_i}} )\leq \max\{4, 2D^2 \}$, and $\bm{\Phi}_{D, \tilde{\varepsilon}}^{\Cross_{s_i}}(\bm{0},\bm{x}_{1,2})=\bm{\Phi}_{D, \tilde{\varepsilon}}^{\Cross_{s_i}}(\bm{W}_{1,2},\bm{0})=\bm{0}$, $i=1,2$. Per Lemma \ref{lem: NN_parallelization}, the parallelization of the FNNs of (\ref{superposition_FNNs}) fed with all elements of $\bm{W}_{1,2}$ and $\bm{x}_{1,2}$ produces $\bm{p}_{1,2}=[\bm{p}_1^T, \bm{p}_2^T]^T$. This is approximately $\bm{\Phi}_{D, \varepsilon}^{\Cross_c}(\bm{W}_{1,2},\bm{x}_{1,2})$ and hence for $(\bm{V}_{1,2})\eqdef (\bm{W}_{1,2}, \bm{x}_{1,2})$ 
  \begin{equation}
  \label{FNNs_equivs}
  \hspace{-2mm}\bm{\Phi}_{D, \varepsilon}^{\Cross_c}(\bm{W}_{1,2},\bm{x}_{1,2})=  
  [(\bm{\Phi}_{D, \tilde{\varepsilon}}^{\Cross_{s_1}} (\bm{V}_{1,2}) )^T, (\bm{\Phi}_{D, \tilde{\varepsilon}}^{\Cross_{s_2}}(\bm{V}_{1,2}))^T]^T, 
  \end{equation}        
  where $\bm{\Phi}_{D, \varepsilon}^{\Cross_c}\in\bm{\mathcal{NN}}_{\infty, \infty, \rho}^{2n(m+1),2m}$ satisfying $\mathcal{L}( \bm{\Phi}_{D, \varepsilon}^{\Cross_c} )\leq C\log_2 (nD^2 \tilde{\varepsilon}^{-1})$, $\mathcal{W}\big( \bm{\Phi}_{D, \varepsilon}^{\Cross_c} \big)\leq 48mn$, $\mathcal{B}\big( \bm{\Phi}_{D, \varepsilon}^{\Cross_c} \big)\leq \max\{4, 2D^2 \}$, and $\bm{\Phi}_{D, \varepsilon}^{\Cross_c}\big(\bm{0},\bm{x}_{1,2}\big)=\bm{\Phi}_{D, \varepsilon}^{\Cross_c}\big(\bm{W}_{1,2},\bm{0} \big)=\bm{0}$. 
  
  Deploying (\ref{FNNs_equivs}) and (\ref{Re_bm_W_times_bm_x})-(\ref{Im_bm_W_times_bm_x}) in the left-hand side of (\ref{complex_matrix_vector_product_approx_NN_bound}), 
  \begin{multline}
  \label{complex_matrix_vector_product_approx_NN_bound_pr_1}
  \| \bm{\Phi}_{D, \varepsilon}^{\Cross_c}(\bm{W}_{1,2},\bm{x}_{1,2})- \bm{p}_{1,2} \|_{L^{\infty}([-D, D]^{2n}; \mathbb{R}^{m})} \stackrel{(a)}{\leq} \sum_{i=1}^2  
   \big( \| \bm{\Phi}_{D, \tilde{\varepsilon}}^{\Cross} \\ (\bm{W}_i, \bm{x}_i)  - \bm{W}_i\bm{x}_i \|_{L^{\infty}([-D, D]^{2n}; \mathbb{R}^{m})} +          
  \| \bm{\Phi}_{D, \tilde{\varepsilon}}^{\Cross} (\bm{W}_i,  \bm{x}_{i+(-1)^{i+1}} ) \\ - \bm{W}_i\bm{x}_{i+(-1)^{i+1}} \|_{L^{\infty}([-D, D]^{2n}; \mathbb{R}^{m})} \big) \stackrel{(b)}{\leq} 4\tilde{\varepsilon},  
  \end{multline}      
  where $(a)$ follows from (\ref{Re_bm_W_times_bm_x})-(\ref{Im_bm_W_times_bm_x}), (\ref{Re_bm_A_times_bm_x_approx})-(\ref{superposition_FNNs}), and the properties of $L^{\infty}$-norm; $(b)$ follows from (\ref{matrix_vector_product_approx_NN_bound}). Setting $\varepsilon=4\tilde{\varepsilon}$ gives (\ref{complex_matrix_vector_product_approx_NN_bound}) and $\mathcal{L}( \bm{\Phi}_{D, \varepsilon}^{\Cross_c} )\leq C \log_2 (4nD^2 \varepsilon^{-1})$.      \QEDclosed

\section{Proof of Theorem \ref{Theorem_Sobolev_upper_approximation_bound}}
\label{proof_Theorem_Sobolev_upper_approximation_bound}
We begin by proving the following proposition. 
\begin{proposition}
\label{Proposition_N_product_and_sum_NN_Sobolev_bounds}
Let $\bm{w}^T=[w_1, \ldots, w_{n}]^T\in\mathbb{R}^{n}$ and $\bm{x}=[x_1, \ldots, x_{n}]^T\in\mathbb{R}^{n}$. For all $\varepsilon \in (0, 1/2)$, $D\in \mathbb{R}_{+}$, and $\rho(x)\eqdef \textnormal{max}(x,0)$, there exist constants $\bar{C}_1, \bar{C}_2, \bar{C}_3>0$ such that there is a FNN $\bm{\Phi}_{D, \varepsilon}^{\tilde{\times}(\bm{w},\bm{x})}\in\bm{\mathcal{NN}}_{\infty, \infty, \rho}^{2n,1}$ satisfying $\bm{\Phi}_{D, \varepsilon}^{\tilde{\times}(\bm{w},\bm{x})}(\bm{0},\bm{x})=\bm{\Phi}_{D, \varepsilon}^{\tilde{\times}(\bm{w},\bm{x})}(\bm{x},\bm{0})=0$, $\forall \bm{x}\in\mathbb{R}^n$, and 
\begin{equation}
\label{N_product_and_sum_approx_NN_Sobolev_bound}
\| \bm{\Phi}_{D, \varepsilon}^{\tilde{\times}(\bm{w},\bm{x})}(\bm{w},\bm{x})-\bm{w}\bm{x}\|_{\bm{\mathcal{W}}^{1,\infty}((-D, D)^{2n}; dwdx)}\leq \varepsilon  
\end{equation}
\begin{equation}
\label{L_M_N_epsilon_limit}
\hspace{-2.5mm}\mathcal{M}( \bm{\Phi}_{D, \varepsilon}^{\tilde{\times}(\bm{w},\bm{x})}), \mathcal{N}( \bm{\Phi}_{D, \varepsilon}^{\tilde{\times}(\bm{w},\bm{x})}), \mathcal{L}( \bm{\Phi}_{D, \varepsilon}^{\tilde{\times}(\bm{w},\bm{x})}) \leq  \bar{C}_1 \log_2(\varepsilon^{-1})+  \bar{C}_2
\end{equation}
\begin{equation}
\label{Sob_semi_norm_bm_w_bm_x}
|\bm{\Phi}_{D, \varepsilon}^{\tilde{\times}(\bm{w},\bm{x})}|_{\bm{\mathcal{W}}^{1,\infty}((-D, D)^{2n})} \leq \bar{C}_3D.
\end{equation}
\end{proposition} 
 
\proof Since $\bm{w}\bm{x}=\sum_{i=1}^n w_ix_i$, it can be implemented via the superposition of $n$ $\bm{\Phi}_{D, \tilde{\varepsilon}}^{\tilde{\times}(w,x)}$s characterized via Proposition \ref{Sob_Proposition_product_NN} and fed, respectively, with input tuples $(w_1, x_1), \ldots, (w_n, x_n)$. Thus, the output of a ReLU FNN -- $\bm{\Phi}_{D, \varepsilon}^{\tilde{\times}(\bm{w},\bm{x})}$ -- that approximates $\bm{w}\bm{x}$ can be expressed as      
\begin{equation}
\label{N_product_and_N_sum_NN_Sobolev_bounds}
\bm{\Phi}_{D, \varepsilon}^{\tilde{\times}(\bm{w},\bm{x})}(\bm{w},\bm{x})=\sum_{i=1}^n \bm{\Phi}_{D, \tilde{\varepsilon}}^{\tilde{\times}(w,x)}(w_i,x_i),  
\end{equation}
where $\bm{\Phi}_{D, \tilde{\varepsilon}}^{\tilde{\times}(w,x)}(0,x)=0=\bm{\Phi}_{D, \tilde{\varepsilon}}^{\tilde{\times}(w,x)}(x,0)$, $\forall x\in\mathbb{R}$; 
\begin{equation}
\label{approx_limit_varepsilon}
\| \bm{\Phi}_{D, \tilde{\varepsilon}}^{\tilde{\times}(w,x)}(w,x)-wx \|_{\bm{\mathcal{W}}^{1,\infty}((-D, D)^2; dwdx)}\leq \tilde{\varepsilon} 
\end{equation}
\begin{equation}
\label{L_M_N_limit}
\hspace{-1mm}\mathcal{L}( \bm{\Phi}_{D, \tilde{\varepsilon}}^{\tilde{\times}(w,x)}), \mathcal{M}( \bm{\Phi}_{D, \tilde{\varepsilon}}^{\tilde{\times}(w,x)}), \mathcal{N}( \bm{\Phi}_{D, \tilde{\varepsilon}}^{\tilde{\times}(w,x)}) \leq C_1\log_2(\tilde{\varepsilon}^{-1})+C_2.  
\end{equation}
Deploying (\ref{Sob_product_NN_expression}) in (\ref{N_product_and_N_sum_NN_Sobolev_bounds}),  
\begin{multline}
\label{Sob_sum_product_NN_expression}
\bm{\Phi}_{D, \varepsilon}^{\tilde{\times}(\bm{w},\bm{x})}(\bm{w},\bm{x})=2D^2 \times \\  \sum_{i=1}^n\Big( \bm{\Phi}_{\delta}^{\textnormal{sq}}\Big( \frac{|w_i+x_i|}{2D}\Big)-\bm{\Phi}_{\delta}^{\textnormal{sq}}\Big( \frac{|w_i|}{2D}\Big)-\bm{\Phi}_{\delta}^{\textnormal{sq}}\Big( \frac{|x_i|}{2D}\Big)\Big). 
\end{multline}
With respect to (\ref{Sob_polarization_identity_3}),    
\begin{equation}
\label{Sob_sum_polarization_identity_3}
\bm{w}\bm{x}=2D^2\sum_{i=1}^n\Big( \Big( \frac{|w_i+x_i|}{2D}\Big)^2-\Big( \frac{|w_i|}{2D}\Big)^2-\Big( \frac{|x_i|}{2D}\Big)^2\Big).
\end{equation}
For our upcoming simplifications, we define these parameters for $\mathcal{S}\eqdef \big\{(w_1,x_1), \ldots, (w_i,x_i), \ldots, (w_n,x_n) \big\}$, $i\in[n]$:
\begin{subequations}
	\begin{align}
	\label{Sob_sum_transformation_1}
	t_{w_i}:(-D, D)^{2n}&\to (0,1), \hspace{0.5mm} \mathcal{S}\mapsto |w_i|/2D    \\
	\label{Sob_sum_transformation_2}
	t_{x_i}:(-D, D)^{2n}&\to (0,1), \hspace{1mm} \mathcal{S} \mapsto |x_i|/2D   \\
	\label{Sob_sum_transformation_3}
	t_{w_ix_i}:(-D, D)^{2n}&\to (0,1), \hspace{1mm} \mathcal{S} \mapsto  |w_i+x_i|/2D  \\
	\label{Sob_sum_transformation_4}
	f: [0,1]&\to\mathbb{R}, \hspace{1mm} x\mapsto x^2.
	\end{align}
\end{subequations}
Employing (\ref{Sob_sum_transformation_1})-(\ref{Sob_sum_transformation_4}), (\ref{Sob_sum_product_NN_expression}) and (\ref{Sob_sum_polarization_identity_3}) can be expressed as
\begin{multline}
\label{Sob_sum_wx_NN_composition}
\frac{\bm{\Phi}_{D, \varepsilon}^{\tilde{\times}(\bm{w},\bm{x})}(\bm{w},\bm{x})}{2D^2}=\sum_{i=1}^n\big( \bm{\Phi}_{\delta}^{\textnormal{sq}}\circ t_{w_ix_i}-\sum_{u_i\in\{w_i,x_i\}}\bm{\Phi}_{\delta}^{\textnormal{sq}}\circ t_{u_i} \big)
\end{multline}
\begin{equation}
\label{Sob_sum_wx_composition}
\bm{w}\bm{x}=2D^2\sum_{i=1}^n\big( f\circ t_{w_ix_i}-f\circ t_{w_i}-f\circ t_{x_i}\big). 
\end{equation}  

Substituting (\ref{Sob_sum_wx_NN_composition}) and (\ref{Sob_sum_wx_composition}) in the left-hand side of (\ref{N_product_and_sum_approx_NN_Sobolev_bound}), collecting similar terms, and applying the properties of $L^{\infty}$-norm -- via Definition \ref{Sobolev_space_def} and (\ref{Sob_norm_2}) -- result in 
\begin{multline}
\label{Sob_sum_product_approx_bound_analysis_3}
\|\bm{\Phi}_{D,\varepsilon}^{\tilde{\times}(\bm{w},\bm{x})}(\bm{w},\bm{x})-\bm{w}\bm{x}\|_{\bm{\mathcal{W}}^{1,\infty}((-D, D)^{2n}; dwdx)}\leq  2D^2\times  \\ \sum_{i=1}^n \sum_{u_i\in\{w_i, x_i, w_ix_i\}} \|\big(\bm{\Phi}_{\delta}^{\textnormal{sq}}-f\big)\circ t_{u_i}\|_{\bm{\mathcal{W}}^{1,\infty}((-D, D)^{2n})}.  
\end{multline}
Applying Corollary \ref{coro: Sobolev_function_composition} -- via (\ref{gof_Sob_norm}) -- to each summand of (\ref{Sob_sum_product_approx_bound_analysis_3}),  
\begin{multline}
\label{Sob_sum_product_approx_bound_analysis_4}
\hspace{-4mm}\|(\bm{\Phi}_{\delta}^{\textnormal{sq}}-f)\circ t_{u_i}\|_{\bm{\mathcal{W}}^{1,\infty}((-D, D)^{2n})}\leq  
C\max\{ \|\bm{\Phi}_{\delta}^{\textnormal{sq}}-f \|_{L^{\infty}((0, 1)^2)}  \\ , |\bm{\Phi}_{\delta}^{\textnormal{sq}}-f |_{\bm{\mathcal{W}}^{1,\infty}((0, 1)^2)} |  t_{u_i}|_{\bm{\mathcal{W}}^{1,\infty}((-D, D)^{2n})} \}.   
\end{multline}
Exploiting (\ref{Sob_semi_norm_2}) for $t_{u_i}\in\{t_{w_i}, t_{x_i}, t_{w_ix_i}\}$,  
\begin{equation}
\label{Sob_norm_values_nn_1}
|  t_{u_i} |_{\bm{\mathcal{W}}^{1,\infty}((-D, D)^{2n})}=1/2D.
\end{equation}
Substituting (\ref{Sob_norm_values_nn_1}) into (\ref{Sob_sum_product_approx_bound_analysis_4}) and, in turn, into (\ref{Sob_sum_product_approx_bound_analysis_3}) lead to   
\begin{multline}
\label{Sob_sum_product_approx_bound_analysis_5}
\hspace{-3mm}\|\bm{\Phi}_{D,\varepsilon}^{\tilde{\times}(\bm{w},\bm{x})}(\bm{w},\bm{x})-\bm{w}\bm{x}\|_{\bm{\mathcal{W}}^{1,\infty}((-D, D)^{2n}; dwdx)}\leq 2D^2C \sum_{i=1}^n \\  \hspace{-2mm} \sum_{u_i\in\{w_i, x_i, w_ix_i\}}\max\big\{
 \|\bm{\Phi}_{\delta}^{\textnormal{sq}}-f \|_{L^{\infty}((0, 1)^2)},  \frac{|\bm{\Phi}_{\delta}^{\textnormal{sq}}-f  |_{\bm{\mathcal{W}}^{1,\infty}((0,1)^2)}}{2D} \\ \big\}          
\leq 6D^2Cn \|\bm{\Phi}_{\delta}^{\textnormal{sq}}(x)-x^2 \|_{L^{\infty}((0, 1)^2)} \stackrel{(a)}{\leq} (6D^2C\delta)n, 
\end{multline}
where $(a)$ follows from Proposition \ref{prop_approx_x^2}. W.r.t. (\ref{Sob_sum_product_approx_bound_analysis_5}) and the $\tilde{\varepsilon}$ constraint of (\ref{approx_limit_varepsilon}), $\tilde{\varepsilon}=6D^2C\delta$. As a result, setting $\varepsilon=\tilde{\varepsilon} n$, (\ref{Sob_sum_product_approx_bound_analysis_5}) simplifies to (\ref{N_product_and_sum_approx_NN_Sobolev_bound}). This completes the proof of (\ref{N_product_and_sum_approx_NN_Sobolev_bound}). 

Regarding the superposition of (\ref{N_product_and_N_sum_NN_Sobolev_bounds}) per Lemma \ref{lem: NN_superposition_parallelization}, $\bm{\Phi}_{D, \varepsilon}^{\tilde{\times}(\bm{w},\bm{x})}(\bm{w},\bm{x})$ is implemented by adding the outputs of $n$ $\bm{\Phi}_{D, \tilde{\varepsilon}}^{\tilde{\times}(w,x)}$s fed, respectively, with $(w_i,x_i)$. Thus, 
\begin{multline}
\label{M_product_limit_1}
\mathcal{M}\big(\bm{\Phi}_{D, \varepsilon}^{\tilde{\times}(\bm{w},\bm{x})}\big)\leq\sum_{i=1}^n \mathcal{M}\big(\bm{\Phi}_{D, \tilde{\varepsilon}}^{\tilde{\times}(w,x)}\big)+n\stackrel{(a)}{\leq} n[ C_1\log_2(n\varepsilon^{-1})\\ +(C_2+1) ] \stackrel{(b)}{=}\bar{C}_1\log_2(\varepsilon^{-1})+\bar{C}_2, 	
\end{multline}    
where $(a)$ follows from (\ref{L_M_N_limit}) and $\varepsilon=\tilde{\varepsilon} n$; $(b)$ is due to setting $\bar{C}_1=nC_1$ and $\bar{C}_2=n[C_1\log_2 n+C_2+1]$. Similarly,  
\begin{multline}
\label{NL_product_limit_1}
\mathcal{N}(\bm{\Phi}_{D, \varepsilon}^{\tilde{\times}(\bm{w},\bm{x})})=\sum_{i=1}^n \mathcal{N}(\bm{\Phi}_{D, \tilde{\varepsilon}}^{\tilde{\times}(w,x)})\stackrel{(a)}{\leq} n[ C_1\log_2(n\varepsilon^{-1})+C_2] \\  \leq\bar{C}_1\log_2(\varepsilon^{-1})+\bar{C}_2;   	
\mathcal{L}\big(\bm{\Phi}_{D, \varepsilon}^{\tilde{\times}(\bm{w},\bm{x})}\big)=\mathcal{L}\big(\bm{\Phi}_{D, \tilde{\varepsilon}}^{\tilde{\times}(w,x)}\big)\\ \stackrel{(a)}{\leq}  C_1\log_2(n\varepsilon^{-1}) +C_2 \leq \bar{C}_1\log_2(\varepsilon^{-1})+\bar{C}_2,
\end{multline}
where $(a)$ follows from (\ref{L_M_N_limit}) and the assignment $\varepsilon=\tilde{\varepsilon} n$. Thus, (\ref{M_product_limit_1}) and (\ref{NL_product_limit_1}) lead to (\ref{L_M_N_epsilon_limit}). Deploying Definition \ref{Sobolev_semi_norm_def} -- via (\ref{Sob_semi_norm_2}) -- and applying the properties of $L^{\infty}$-norm to (\ref{Sob_sum_wx_NN_composition}):    
\begin{multline}
\label{Sob_semi_norm_bm_w_bm_x_2}
|\bm{\Phi}_{D, \varepsilon}^{\tilde{\times}(\bm{w},\bm{x})}|_{\bm{\mathcal{W}}^{1,\infty}((-D, D)^{2n})} \leq 2D^2\sum_{i=1}^n\sum_{u_i\in \{w_i, x_i, w_ix_i\} }   \\ 
|  \bm{\Phi}_{\delta}^{\textnormal{sq}}(x_i)\circ t_{u_i}  |_{\bm{\mathcal{W}}^{1,\infty}((-D, D)^{2n})} \stackrel{(a)}{\leq} 2D^2C \sum_{i=1}^n\sum_{u_i\in \{w_i, x_i, w_ix_i\} }\\
 |  \bm{\Phi}_{\delta}^{\textnormal{sq}}(x) |_{\bm{\mathcal{W}}^{1,\infty}((0, 1))} | t_{u_i}  |_{\bm{\mathcal{W}}^{1,\infty}((-D, D)^{2n})} \stackrel{(b)}{\leq} 3nCC_4D,
\end{multline}
where $(a)$ follows from (\ref{gof_Sob_semi_norm}); $(b)$ is due to (\ref{Sob_semi_norm_bound_x^2}) and (\ref{Sob_norm_values_nn_1}). Setting $\bar{C}_3=3nCC_4>0$, (\ref{Sob_semi_norm_bm_w_bm_x_2}) asserts that $\big|\bm{\Phi}_{D, \varepsilon}^{\tilde{\times}(\bm{w},\bm{x})}\big|_{\bm{\mathcal{W}}^{1,\infty}((-D, D)^{2n})}\leq\bar{C}_3D$. This proves (\ref{Sob_semi_norm_bm_w_bm_x}).    \QEDclosed

Using Proposition \ref{Proposition_N_product_and_sum_NN_Sobolev_bounds} regarding an approximating FNN on a vector-vector product and Lemma \ref{lem: NN_parallelization} (parallelization of FNNs), we proceed to the developments detailed in the sequel. 

From (\ref{W_x_product}), $\bm{W}\bm{x}=[\bm{w}_1\bm{x}, \dots, \bm{w}_m\bm{x} ]^T\in\mathbb{R}^{m}$. Meanwhile, Proposition \ref{Proposition_N_product_and_sum_NN_Sobolev_bounds} affirms that $\bm{w}_i\bm{x}$ can be approximated via a ReLU FNN such that
\begin{equation}
\label{N_product_and_sum_approx_NN_Sobolev_bound_tilde_varepsion}
\| \bm{\Phi}_{D, \varepsilon}^{\tilde{\times}(\bm{w},\bm{x})}(\bm{w}_i,\bm{x})-\bm{w}_i\bm{x}\|_{\bm{\mathcal{W}}^{1,\infty}((-D, D)^{2n}; dwdx)}\leq \varepsilon,    
\end{equation}    
where the ReLU FNN constraints of (\ref{L_M_N_epsilon_limit}) are valid. Thus, $\bm{W}\bm{x}$ can be approximated via the outputs of $m$ ReLU FNNs as      
\begin{equation}
\label{paralleized_FNN_outputs}
[\bm{\Phi}_{D, \varepsilon}^{\tilde{\times}(\bm{w},\bm{x})}(\bm{w}_1,\bm{x}), \dots, \bm{\Phi}_{D, \varepsilon}^{\tilde{\times}(\bm{w},\bm{x})}(\bm{w}_m,\bm{x}) ]^T\approx \bm{W}\bm{x}, 
\end{equation}
where $(\bm{\Phi}_{D, \varepsilon}^{\tilde{\times}(\bm{w},\bm{x})}(\bm{w}_i,\bm{x}))^T=\bm{\Phi}_{D, \varepsilon}^{\tilde{\times}(\bm{w},\bm{x})}(\bm{w}_i,\bm{x})$, $i\in [m]$. Let us now consider a selection matrix $\bm{\pi}_i\eqdef \{0,1\}^{2n\times n(m+1)}$ equated as $(\bm{w}_i,\bm{x})\eqdef [\bm{w}_i^T, \bm{x}^T]^T=\bm{\pi}_i [ (\bm{W}(:))^T, \bm{x}^T]^T$. If we let $\bm{\Phi}_{D, \varepsilon}^{\tilde{\times}(\bm{w},\bm{x})}\eqdef\big[ [\bm{W}_1, \bm{b}_1], [\bm{W}_2, \bm{b}_2], \ldots, [\bm{W}_K, \bm{b}_K] \big]$ and $\tilde{\bm{\Phi}}_{D, \varepsilon}^{\tilde{\times}(\bm{w}_i,\bm{x})}\eqdef\big[ [\tilde{\bm{W}}_1^i, \tilde{\bm{b}}_1^i], [\tilde{\bm{W}}_2^i, \tilde{\bm{b}}_2^i], \ldots, [\tilde{\bm{W}}_K^i, \tilde{\bm{b}}_K^i] \big]$, then 
\begin{equation}
\label{FNN_equivalence_1}
\bm{\Phi}_{D, \varepsilon}^{\tilde{\times}(\bm{w},\bm{x})}(\bm{w}_i,\bm{x})=\tilde{\bm{\Phi}}_{D, \varepsilon}^{\tilde{\times}(\bm{w}_i,\bm{x})}(\bm{W},\bm{x}), \hspace{2mm} i\in [m]
\end{equation}     
provided that 
\begin{equation}
\label{FNN_equivalence_2}
\tilde{\bm{\Phi}}_{D, \varepsilon}^{\tilde{\times}(\bm{w}_i,\bm{x})}=\big[ [\bm{W}_1\bm{\pi}_i, \bm{b}_1], [\bm{W}_2, \bm{b}_2], \ldots, [\bm{W}_K, \bm{b}_K] \big].
\end{equation}
Choosing $m$ selection matrices that fulfill (\ref{FNN_equivalence_1}) and (\ref{FNN_equivalence_2}), 
\begin{equation}
\label{paralleized_FNN_outputs_1}
[\tilde{\bm{\Phi}}_{D, \varepsilon}^{\tilde{\times}(\bm{w}_1,\bm{x})}(\bm{W},\bm{x}), \dots, \tilde{\bm{\Phi}}_{D, \varepsilon}^{\tilde{\times}(\bm{w}_m,\bm{x})}(\bm{W},\bm{x}) ]^T\approx \bm{W}\bm{x}. 
\end{equation}
Therefore, w.r.t. the constraints of Lemma \ref{lem: NN_parallelization} and (\ref{FNN_equivalence_1}), (\ref{paralleized_FNN_outputs_1}) is also equated for $\tilde{\bm{\Phi}}\eqdef [\tilde{\bm{\Phi}}_{D, \varepsilon}^{\tilde{\times}(\bm{w}_1,\bm{x})}, \ldots, \tilde{\bm{\Phi}}_{D, \varepsilon}^{\tilde{\times}(\bm{w}_m,\bm{x})}]$ as 
\begin{multline}
\label{parallelized_output_1}
\mathcal{P}(\tilde{\bm{\Phi}})(\bm{W}, \bm{x})= [\tilde{\bm{\Phi}}_{D, \varepsilon}^{\tilde{\times}(\bm{w}_1,\bm{x})}(\bm{W},\bm{x}), \dots, \tilde{\bm{\Phi}}_{D, \varepsilon}^{\tilde{\times}(\bm{w}_m,\bm{x})}(\bm{W},\bm{x}) ]^T \\
=[\bm{\Phi}_{D, \varepsilon}^{\tilde{\times}(\bm{w},\bm{x})}(\bm{w}_1,\bm{x}), \dots, \bm{\Phi}_{D, \varepsilon}^{\tilde{\times}(\bm{w},\bm{x})}(\bm{w}_m,\bm{x})]^T, 
\end{multline}	
where $\mathcal{M}\big(\mathcal{P}(\tilde{\bm{\Phi}})\big)=\sum_{i=1}^m \mathcal{M}\big(\bm{\Phi}_{D, \varepsilon}^{\tilde{\times}(\bm{w},\bm{x})} \big)\leq m[\bar{C}_1 \log_2(\varepsilon^{-1})+\bar{C}_2]$; $\mathcal{N}\big(\mathcal{P}(\tilde{\bm{\Phi}})\big)\leq\sum_{i=1}^m \mathcal{N}\big(\bm{\Phi}_{D, \varepsilon}^{\tilde{\times}(\bm{w},\bm{x})} \big)\leq m[\bar{C}_1 \log_2(\varepsilon^{-1})+\bar{C}_2]$; and $\mathcal{L}(\mathcal{P}(\tilde{\bm{\Phi}}))=\mathcal{L}\big(\bm{\Phi}_{D, \varepsilon}^{\tilde{\times}(\bm{w},\bm{x})} \big)\leq [\bar{C}_1 \log_2(\varepsilon^{-1})+\bar{C}_2]\leq m[\bar{C}_1 \log_2(\varepsilon^{-1})+\bar{C}_2]$. 

Since $\bm{\Phi}_{D, \varepsilon}^{\Cross}$ takes inputs that are all the elements of $\bm{W}$ and $\bm{x}$ to produce their approximated product which is also realizable by the parallelized FNN of (\ref{parallelized_output_1}), $\bm{\Phi}_{D, \varepsilon}^{\Cross}(\bm{W},\bm{x})=\mathcal{P}(\tilde{\bm{\Phi}})(\bm{W},\bm{x})$. Thus, (\ref{parallelized_output_1}) corroborates w.r.t. every scalar output that 
\begin{equation}
\label{parallelized_output_2}
\bm{\Phi}_{D, \varepsilon}^{\Cross}\big(\bm{W},\bm{x}\big)=\big[\bm{\Phi}_{D, \varepsilon}^{\tilde{\times}(\bm{w},\bm{x})}(\bm{w}_1,\bm{x}), \dots,  \bm{\Phi}_{D, \varepsilon}^{\tilde{\times}(\bm{w},\bm{x})}(\bm{w}_m,\bm{x})\big]^T. 
\end{equation}       
As a result, $\bm{\Phi}_{D, \varepsilon}^{\Cross}\big(\bm{0},\bm{x}\big)=\bm{\Phi}_{D, \varepsilon}^{\Cross}\big(\bm{W},\bm{0}\big)=\bm{0}$. Therefore, substituting (\ref{parallelized_output_2}) and (\ref{W_x_product}) into (\ref{matrix_vector_product_Sob_approx_NN_bound}), collecting similar terms, and exploiting (\ref{Sob_norm_4}):   
\begin{multline}
\label{App_error_bm_W_bm_x_Sob_norm_2}
\| \bm{\Phi}_{D, \varepsilon}^{\Cross}(\bm{W},\bm{x})-\bm{W}\bm{x}\|_{\bm{\mathcal{W}}^{1,\infty}((-D, D)^{2n}; \mathbb{R}^{m})}=\max_{i=1, \ldots, m} \\ \| \bm{\Phi}_{D, \varepsilon}^{\tilde{\times}(\bm{w},\bm{x})}(\bm{w}_i,\bm{x})-\bm{w}_i\bm{x}  \|_{\bm{\mathcal{W}}^{1,\infty}((-D, D)^{2n}; dwdx)}\stackrel{(a)}{\leq}\varepsilon,
\end{multline} 
where $(a)$ follows from (\ref{N_product_and_sum_approx_NN_Sobolev_bound_tilde_varepsion}). This completes the proof of (\ref{matrix_vector_product_Sob_approx_NN_bound}).  

Because $\bm{\Phi}_{D, \varepsilon}^{\Cross}(\bm{W},\bm{x})=\mathcal{P}(\tilde{\bm{\Phi}})(\bm{W},\bm{x})$, the aforementioned network constraints of $\mathcal{P}(\tilde{\bm{\Phi}})$ become the constraints of $\bm{\Phi}_{D, \varepsilon}^{\Cross}$. As a result, letting $m\bar{C}_1=\bar{\bar{C}}_1$ and $m\bar{C}_2=\bar{\bar{C}}_2$ lead to the network constraints mentioned in Theorem \ref{Theorem_Sobolev_upper_approximation_bound}.   \QEDclosed

\section*{Acknowledgments}
The author acknowledges the US Department of Commerce and NIST for funding this work; the NIST support staff for facilitating his NIST Guest Researcher-ship; and the Editor and anonymous Reviewers for their thoughtful comments that have guided the significant improvement of his previously submitted manuscript.

\section*{Dedication}
Along with his NIST colleagues, the author mourned the sudden loss of Michael Souryal who was his thoughtful NIST mentor. Honoring his late NIST mentor, the author dedicates this paper to the memory of Michael Souryal.

\balance

\clearpage


\section*{Supplementary material}

# Error Bounds for a Matrix-Vector Product Approximation with Deep ReLU Neural Networks

Tilahun M. Getu, *Member, IEEE*

*Abstract*—This supplementary material comprises *Proof of Lemma 1*, *Proof of Lemma 2*, and *Proof of Proposition 2*.

*Notation*: scalars, vectors, and matrices are represented by italic letters, bold lowercase letters, and bold uppercase letters, respectively. $\mathbb{R}^n$ denotes the sets of $n$-dimensional vectors of real numbers. $\mathbb{N}_0$ denotes the sets of natural numbers including zero. A sequence of $K$ matrix-vector tuples is denoted as $[[\boldsymbol{W}_1, \boldsymbol{b}_1], \ldots, [\boldsymbol{W}_K, \boldsymbol{b}_K]]$. The Python® syntax `len(W)` stands for the number of rows of $\boldsymbol{W}$. :=, max, $(\cdot)^T$, diag$(\cdot)$, $\boldsymbol{I}_n$, $\boldsymbol{0}_{m \times n}$, and $\boldsymbol{0}$ denote equal by definition, maximum, transpose, a (block) diagonal matrix, an $n \times n$ identity matrix, an $m \times n$ zero matrix, and a zero vector (matrix) whose dimension will be clear in context, respectively. The class of feedforward neural networks (FNNs) $\boldsymbol{\Phi} : \mathbb{R}^{N_0} \to \mathbb{R}^{N_K}$ with no more than $K$ layers, connectivity no more than $M$, input dimension $N_0$, output dimension $N_K$, and activation function $\rho$ is denoted as $\mathcal{NN}_{K,M,\rho}^{N_0,N_K}$. For an FNN $\boldsymbol{\Phi}$, $\mathcal{L}(\boldsymbol{\Phi})$, $\mathcal{M}(\boldsymbol{\Phi})$, $\mathcal{W}(\boldsymbol{\Phi})$, and $\mathcal{B}(\boldsymbol{\Phi})$ stand for the depth of $\boldsymbol{\Phi}$, the connectivity of $\boldsymbol{\Phi}$, the maximum width of $\boldsymbol{\Phi}$, and the maximum absolute value of the weights of $\boldsymbol{\Phi}$, respectively. Regarding $L^p$ (Lebesgue) spaces and $f(x) : \mathbb{R}^d \to \mathbb{R}$, $\|f\|_{L^\infty(\Omega)} := \inf\{C \geq 0 : |f(x)| \leq C, \forall x \in \Omega | \Omega \subset \mathbb{R}^d\}$.

## I. *Proof of Lemma 1*

If $\boldsymbol{\Phi}_K^{\boldsymbol{I}_{N_0}}(\boldsymbol{x}) = \boldsymbol{x}$ for $\boldsymbol{x} \in \mathbb{R}^{N_0}$, then (1) leads to

$$\boldsymbol{A}_K(\rho(\boldsymbol{A}_{K-1}(\rho(\ldots\rho(\boldsymbol{A}_1(\boldsymbol{x})))))) = \boldsymbol{x}, \qquad (74)$$

where $\rho(\boldsymbol{x}) = [\max(0, x_1), \ldots, \max(0, x_{N_0})]^T$; $\boldsymbol{A}_k(\boldsymbol{x}_{k-1}) = \boldsymbol{W}_k \boldsymbol{x}_{k-1} + \boldsymbol{b}_k$ for $\boldsymbol{x}_{k-1} = \rho(\boldsymbol{A}_{k-1}(\boldsymbol{x}_{k-2}))$ and $\boldsymbol{x}_0 = \boldsymbol{x}$; and $[\boldsymbol{W}_k, \boldsymbol{b}_k]$ – comprising $\boldsymbol{\Phi}_K^{\boldsymbol{I}_{N_0}}$ – is to be inferred for $k \in [K]$. In the sequel, we are going to exploit these identities: $\boldsymbol{x} = \rho(\boldsymbol{x}) - \rho(-\boldsymbol{x})$ and $\rho(\rho(\boldsymbol{x})) = \rho(\boldsymbol{x})$.

For (74) to be valid, here is a rectified linear unit (ReLU) FNN construction: all layers except the last layer would have an output equal to $[(\rho(\boldsymbol{x}))^T, (\rho(-\boldsymbol{x}))^T]^T$; these two components would then be added – at the last layer – so that we will obtain $\rho(\boldsymbol{x}) - \rho(-\boldsymbol{x}) = \boldsymbol{x}$. With respect to (w.r.t.) this construction, $\boldsymbol{x}_1 = \rho(\boldsymbol{W}_1 \boldsymbol{x} + \boldsymbol{b}_1) = \rho(\boldsymbol{W}_1[\rho(\boldsymbol{x}) - \rho(-\boldsymbol{x})] + \boldsymbol{b}_1) = \rho(\boldsymbol{W}_1 \rho(\boldsymbol{x}) - \boldsymbol{W}_1 \rho(-\boldsymbol{x}) + \boldsymbol{b}_1) = [(\rho(\boldsymbol{x}))^T, (\rho(-\boldsymbol{x}))^T]^T$. W.r.t. the last equality, $\boldsymbol{W}_1 = [\boldsymbol{I}_{N_0}, -\boldsymbol{I}_{N_0}]^T$ and $\boldsymbol{b}_1 = \boldsymbol{0}_{2N_0 \times 1}$. Thus, $[\boldsymbol{W}_1, \boldsymbol{b}_1] = [[\boldsymbol{I}_{N_0}, -\boldsymbol{I}_{N_0}]^T, \boldsymbol{0}]$.

W.r.t. the aforementioned construction, $\boldsymbol{x}_k = [(\rho(\boldsymbol{x}))^T, (\rho(-\boldsymbol{x}))^T]^T$, $k = 2, \ldots, K-1$. Accordingly,

T. M. Getu is with the National Institute of Standards and Technology (NIST), 100 Bureau Drive, Gaithersburg, MD 20899, USA and also with the École de Technologie Supérieure (ÉTS), Montréal, QC H3C 1K3, Canada (e-mail: tilahun.getu@nist.gov).

$\boldsymbol{x}_2 = \rho(\boldsymbol{W}_2 \boldsymbol{x}_1 + \boldsymbol{b}_2) = \rho(\boldsymbol{W}_2[(\rho(\boldsymbol{x}))^T, (\rho(-\boldsymbol{x}))^T]^T + \boldsymbol{b}_2) = [(\rho(\boldsymbol{x}))^T, (\rho(-\boldsymbol{x}))^T]^T$. Concerning the last equality, $\boldsymbol{W}_2 = \boldsymbol{I}_{2N_0}$, $\boldsymbol{b}_2 = \boldsymbol{0}_{2N_0 \times 1}$, and hence $[\boldsymbol{W}_2, \boldsymbol{b}_2] = [\boldsymbol{I}_{2N_0}, \boldsymbol{0}]$. Repeating the same procedure for the $k$-th layer – $k \in \{3, \ldots, K-1\}$, $[\boldsymbol{W}_k, \boldsymbol{b}_k] = [\boldsymbol{I}_{2N_0}, \boldsymbol{0}]$. At the last layer, $\boldsymbol{x}_K$ should be $\boldsymbol{x}$ for $\boldsymbol{\Phi}_K^{\boldsymbol{I}_{N_0}}$ to be an I-FNN. Thus, $\boldsymbol{x}_K = \boldsymbol{W}_K \boldsymbol{x}_{K-1} + \boldsymbol{b}_K = \boldsymbol{W}_K[(\rho(\boldsymbol{x}))^T, (\rho(-\boldsymbol{x}))^T]^T + \boldsymbol{b}_K = \boldsymbol{x}$. From the last equality, $\boldsymbol{W}_K = [\boldsymbol{I}_{N_0}, -\boldsymbol{I}_{N_0}]$, $\boldsymbol{b}_K = \boldsymbol{0}_{N_0 \times 1}$, and hence $[\boldsymbol{W}_K, \boldsymbol{b}_K] = [[\boldsymbol{I}_{N_0}, -\boldsymbol{I}_{N_0}], \boldsymbol{0}]$. Therefore, recalling that an I-FNN is denoted by a sequence of matrix-vector tuples as in (5), employing the deduced $[\boldsymbol{W}_k, \boldsymbol{b}_k]$ for $k \in [K]$ leads to (5). For $K = 1$, (74) leads to $\boldsymbol{x}_1 = \boldsymbol{W}_1 \boldsymbol{x} + \boldsymbol{b}_1 = \boldsymbol{x}$. Consequently, $\boldsymbol{\Phi}_1^{\boldsymbol{I}_{N_0}} = [[\boldsymbol{I}_{N_0}, \boldsymbol{0}]]$. ∎

## II. *Proof of Lemma 2*

For $K, n \geq 2$, it follows via (3a) and (3b) that the first layer's output of the parallelized FNN becomes $\boldsymbol{x}_1^p = \rho(\tilde{\boldsymbol{W}}_1 \boldsymbol{x} + \tilde{\boldsymbol{b}}_1)$. For the $n$ FNNs, the respective first layer outputs are given by $\boldsymbol{x}_1^i = \rho(\boldsymbol{W}_1^i \boldsymbol{x} + \boldsymbol{b}_1^i)$, $i \in [n]$. If $\mathcal{P}(\boldsymbol{\Phi}^1, \ldots, \boldsymbol{\Phi}^n)(\boldsymbol{x}) = [(\boldsymbol{\Phi}^1(\boldsymbol{x}))^T, \ldots, (\boldsymbol{\Phi}^n(\boldsymbol{x}))^T]^T$, this equality should regress to all layers' outputs. Hence, $\rho(\tilde{\boldsymbol{W}}_1 \boldsymbol{x} + \tilde{\boldsymbol{b}}_1) = [(\rho(\boldsymbol{W}_1^1 \boldsymbol{x} + \boldsymbol{b}_1^1))^T, \ldots, (\rho(\boldsymbol{W}_1^n \boldsymbol{x} + \boldsymbol{b}_1^n))^T]^T$. Since $\rho$ is applied element-wise, the first parallelization conditions emanate from: $\rho(\tilde{\boldsymbol{W}}_1 \boldsymbol{x} + \tilde{\boldsymbol{b}}_1) = \rho([(\boldsymbol{W}_1^1)^T, \ldots, (\boldsymbol{W}_1^n)^T]^T \boldsymbol{x} + [(\boldsymbol{b}_1^1)^T, \ldots, (\boldsymbol{b}_1^n)^T]^T)$. Hence, $\tilde{\boldsymbol{W}}_1 = [(\boldsymbol{W}_1^1)^T, \ldots, (\boldsymbol{W}_1^n)^T]^T$ and $\tilde{\boldsymbol{b}}_1 = [(\boldsymbol{b}_1^1)^T, \ldots, (\boldsymbol{b}_1^n)^T]^T$. This proves (6a).

Under the first conditions, it follows that $\rho(\tilde{\boldsymbol{W}}_2 \boldsymbol{x}_1^p + \tilde{\boldsymbol{b}}_2) = [(\rho(\boldsymbol{W}_2^1 \boldsymbol{x}_1^1 + \boldsymbol{b}_2^1))^T, \ldots, (\rho(\boldsymbol{W}_2^n \boldsymbol{x}_1^n + \boldsymbol{b}_2^n))^T]^T$. Thus, $\rho(\tilde{\boldsymbol{W}}_2 \boldsymbol{x}_1^p + \tilde{\boldsymbol{b}}_2) = \rho(\text{diag}(\boldsymbol{W}_2^1, \ldots, \boldsymbol{W}_2^n)[(\boldsymbol{x}_1^1)^T, \ldots, (\boldsymbol{x}_1^n)^T]^T + [(\boldsymbol{b}_2^1)^T, \ldots, (\boldsymbol{b}_2^n)^T]^T)$. From the parallelization of the first layers' outputs of the $n$ FNNs, $\boldsymbol{x}_1^p = [(\boldsymbol{x}_1^1)^T, \ldots, (\boldsymbol{x}_1^n)^T]^T$. Accordingly, the second parallelization conditions are: $\tilde{\boldsymbol{W}}_2 = \text{diag}(\boldsymbol{W}_2^1, \ldots, \boldsymbol{W}_2^n)$ and $\tilde{\boldsymbol{b}}_2 = [(\boldsymbol{b}_2^1)^T, \ldots, (\boldsymbol{b}_2^n)^T]^T$. Moreover, employing the same procedure recursively leads to: $\tilde{\boldsymbol{W}}_k = \text{diag}(\boldsymbol{W}_k^1, \ldots, \boldsymbol{W}_k^n)$ and $\tilde{\boldsymbol{b}}_k = [(\boldsymbol{b}_k^1)^T, \ldots, (\boldsymbol{b}_k^n)^T]^T$, $3 \leq k \leq K$. This ends the proof of (6b).

As we are parallelizing $n$ equal-depth FNNs, $\mathcal{L}(\mathcal{P}(\tilde{\boldsymbol{\Phi}})) = K$ and $\mathcal{M}(\mathcal{P}(\tilde{\boldsymbol{\Phi}})) = \sum_{k=1}^{K}(\|\tilde{\boldsymbol{W}}_k\|_{\ell_0} + \|\tilde{\boldsymbol{b}}_k\|_{\ell_0}) = \sum_{k=1}^{K} \sum_{i=1}^{n}(\|\boldsymbol{W}_k^i\|_{\ell_0} + \|\boldsymbol{b}_k^i\|_{\ell_0}) = \sum_{i=1}^{n} \sum_{k=1}^{K}(\|\boldsymbol{W}_k^i\|_{\ell_0} + \|\boldsymbol{b}_k^i\|_{\ell_0})$. From Definition 1 (stated in Sec. II-A), $\mathcal{M}(\boldsymbol{\Phi}^i) = \sum_{k=1}^{K}(\|\boldsymbol{W}_k^i\|_{\ell_0} + \|\boldsymbol{b}_k^i\|_{\ell_0})$. Thus, $\mathcal{M}(\mathcal{P}(\tilde{\boldsymbol{\Phi}})) = \sum_{i=1}^{n} \mathcal{M}(\boldsymbol{\Phi}^i)$. From the conditions of Lemma 2, $\mathcal{N}(\mathcal{P}(\tilde{\boldsymbol{\Phi}})) = N_0 + \sum_{k=1}^{K} \text{len}(\tilde{\boldsymbol{W}}_k)$ and $\mathcal{N}(\boldsymbol{\Phi}^i) = N_0 + \sum_{k=1}^{K} \text{len}(\boldsymbol{W}_k^i)$. Noting that $\text{len}(\tilde{\boldsymbol{W}}_k) = \sum_{i=1}^{n} \text{len}(\boldsymbol{W}_k^i)$ and $\sum_{k=1}^{K} \text{len}(\boldsymbol{W}_k^i) = \mathcal{N}(\boldsymbol{\Phi}^i) - N_0$, $\mathcal{N}(\mathcal{P}(\tilde{\boldsymbol{\Phi}})) = N_0 + \sum_{i=1}^{n} \sum_{k=1}^{K} \text{len}(\boldsymbol{W}_k^i) =$

$N_0 + \sum_{i=1}^{n}(\mathcal{N}(\mathbf{\Phi}^i) - N_0) = \sum_{i=1}^{n} \mathcal{N}(\mathbf{\Phi}^i) - nN_0 + N_0 = \sum_{i=1}^{n} \mathcal{N}(\mathbf{\Phi}^i) - (n-1)N_0$. ∎

### III. *Proof of Proposition 2*

A similar proof is provided in [1, p. 6-8]. For the sake of completeness, clarity, and rigor, we hereby provide a detailed proof.

To begin with the "sawtooth" construction of [2] and [3], let us consider the following sawtooth function $g : [0,0] \to [0,1]$:

$$g(x) = \begin{cases} 2x & : \text{if } x < 1/2 \\ 2(1-x) & : \text{if } x \geq 1/2. \end{cases} \quad (75)$$

Based on (75), let us now consider the $s$-fold composition of $g$ for $s \geq 2$ as

$$g_s := \underbrace{g \circ g \circ \ldots \circ g}_{s \text{ times}}, \quad (76)$$

where

$$g_0(x) := x \text{ and } g_1(x) := g(x). \quad (77)$$

Telgarsky has shown (see [2, p. 4]) that $g_s$ – as defined in (76) – is a sawtooth function with $2^{s-1}$ uniformly distributed "teeth" [2, Lemma 2.4], i.e., each application of $g$ doubles the number of teeth [3]. Employing the linear combinations of $g_s$, we are going to be show that the function $f(x) = x^2$, $x \in [0,1]$, can be approximated.

Suppose $f_m$ be the piece-wise linear interpolation of $f$ with $2^m + 1$ uniformly distributed breakpoints ("knots") per

$$f_m\left(\frac{k}{2^m}\right) = \left(\frac{k}{2^m}\right)^2, \quad k = 0, \ldots, 2^m \text{ and } m \in \mathbb{N}_0. \quad (78)$$

Note that $f_m$ approximates $f$ with error $\varepsilon_m = 2^{-2m-2}$ [3] in the $L^\infty$-norm sense such that [1]

$$\|f_m(x) - x^2\|_{L^\infty[0,1]} \leq 2^{-2m-2}. \quad (79)$$

It is now worth underscoring that refining the interpolation from $f_{m-1}$ to $f_m$ boils down to modifying it by a function proportional to a sawtooth function. Accordingly [1], [3],

$$f_{m-1}(x) - f_m(x) = \frac{g_m(x)}{2^{2m}}. \quad (80)$$

Employing (78), one can infer that

$$f_0(k) = k^2 = k, \quad k \in \{0, 1\}. \quad (81)$$

Deploying a recursive substitution via (80) while using (81),

$$f_m(x) = x - \sum_{s=1}^{m} \frac{g_s(x)}{2^{2s}}. \quad (82)$$

Yarotsky [3] realized (82) via a (maximum) width-3 deep ReLU network whilst allowing connections between neurons (nodes) in non-consecutive layers (a.k.a. *skip connections*). We herein follow the lead of [1] to realize (82) by a (maximum) width-4 deep ReLU network (deep ReLU FNN) based on (75). To this end, note that the superposition of three ReLU – denoted by $\rho(\cdot)$ – functions realizes $g(x)$ defined in (75) as

$$g(x) = 2\rho(x) - 4\rho(x - 1/2) + 2\rho(x - 1). \quad (83)$$

With regard to (76) and (83), it follows that $g_m = \rho(g_m)$:

$$g_m = 2\rho(g_{m-1}) - 4\rho(g_{m-1} - 1/2) + 2\rho(g_{m-1} - 1) \quad (84)$$

$$g_m = 0 \times \rho(f_{m-1}) + 2\rho(g_{m-1}) - 4\rho(g_{m-1} - 1/2) + 2\rho(g_{m-1} - 1). \quad (85)$$

If we have to realize (85) via a 2-layer ReLU FNN representing the composition of affine linear maps, it follows from Definition 1 that

$$g_m = \mathbf{A}_2^{(1)}(\rho(\mathbf{A}_1([g_{m-1}^T, f_{m-1}^T]^T))), \quad (86)$$

where

$$\mathbf{A}_1(\mathbf{x}) = \begin{bmatrix} 1 & 0 \\ 1 & 0 \\ 1 & 0 \\ 0 & 1 \end{bmatrix} \begin{bmatrix} x_1 \\ x_2 \end{bmatrix} - \begin{bmatrix} 0 \\ 1/2 \\ 1 \\ 0 \end{bmatrix} \quad (87)$$

and

$$\mathbf{A}_2^{(1)}(\mathbf{x}) = \begin{bmatrix} 2 & -4 & 2 & 0 \end{bmatrix} \begin{bmatrix} x_1 \\ x_2 \\ x_3 \\ x_4 \end{bmatrix}. \quad (88)$$

To continue, because $f_m$ – defined in (78) – is a non-negative function, $f_m = \rho(f_m)$. Hence, applying ReLU to both sides of (80) gives

$$f_m = \rho(f_{m-1}) - 2^{-2m}\rho(g_m) \quad (89)$$

$$f_m \stackrel{(a)}{=} \rho(f_{m-1}) - 2^{-2m}\big[2\rho(g_{m-1}) - 4\rho(g_{m-1} - 1/2) + 2\rho(g_{m-1} - 1)\big], \quad (90)$$

where $(a)$ is due to (84). Similar to (86), (90) can also be implemented through a 2-layer ReLU FNN as

$$f_m = \mathbf{A}_2^{(2)}(\rho(\mathbf{A}_1([g_{m-1}^T, f_{m-1}^T]^T))), \quad (91)$$

where

$$\mathbf{A}_1(\mathbf{x}) = \begin{bmatrix} 1 & 0 \\ 1 & 0 \\ 1 & 0 \\ 0 & 1 \end{bmatrix} \begin{bmatrix} x_1 \\ x_2 \end{bmatrix} - \begin{bmatrix} 0 \\ 1/2 \\ 1 \\ 0 \end{bmatrix} \quad (92)$$

and

$$\mathbf{A}_2^{(2)}(\mathbf{x}) = \begin{bmatrix} -2^{-2m+1} & 2^{-2m+2} & -2^{-2m+1} & 1 \end{bmatrix} \begin{bmatrix} x_1 \\ x_2 \\ x_3 \\ x_4 \end{bmatrix}. \quad (93)$$

Since the ReLU FNNs of (86) and (91) are fed with the same inputs and have the same first layer weights as well as biases, they can be parallelized. Parallelization of (86) and (91) then leads to

$$\begin{bmatrix} g_m \\ f_m \end{bmatrix} = \mathbf{A}_2\left(\rho\left(\mathbf{A}_1\left(\begin{bmatrix} g_{m-1} \\ f_{m-1} \end{bmatrix}\right)\right)\right), \quad (94)$$

$$\boldsymbol{\Phi}^{\text{sq}}_\varepsilon(x) = \begin{bmatrix} g_m \\ f_m \end{bmatrix} = \boldsymbol{A}_2\bigg(\rho\bigg(\boldsymbol{A}_1\bigg(\boldsymbol{A}_2\bigg(\rho\bigg(\ldots\rho\bigg(\boldsymbol{A}_1\bigg(\boldsymbol{A}_2\bigg(\rho\bigg(\boldsymbol{A}_1\bigg(\begin{bmatrix} x \\ x \end{bmatrix}\bigg)\bigg)\bigg)\bigg)\bigg)\bigg)\bigg)\bigg)\bigg)\bigg). \quad (96)$$

where

$$\boldsymbol{A}_2(\boldsymbol{x}) = \begin{bmatrix} 2 & -4 & 2 & 0 \\ -2^{-2m+1} & 2^{-2m+2} & -2^{-2m+1} & 1 \end{bmatrix} \begin{bmatrix} x_1 \\ x_2 \\ x_3 \\ x_4 \end{bmatrix}. \quad (95)$$

Meanwhile, iterating the ReLU FNN of (94) recursively down to $(f_0, g_0) = (f_0(x), g_0(x)) = (x, x)$ – as asserted by (77) and (81) – produces the ReLU FNN expressed by (96), as shown at the top of this page.

According to (96), $f_m$ can be realized via $\boldsymbol{\Phi}^{\text{sq}}_\varepsilon(x)$ such that $\mathcal{L}(\boldsymbol{\Phi}^{\text{sq}}_\varepsilon) = m+1$; $\mathcal{M}(\boldsymbol{\Phi}^{\text{sq}}_\varepsilon) = 13m$ which is due to (92), (95), and (96); $\mathcal{W}(\boldsymbol{\Phi}^{\text{sq}}_\varepsilon) = 4$; and $\mathcal{B}(\boldsymbol{\Phi}^{\text{sq}}_\varepsilon) \leq 4$ that also follows from (92), (95), and (96). Meanwhile, $f_m$ can be realized through $\mathcal{NN}^{1,1}_{m+1, 13m, \rho}$ for $\rho(x) := \max(0, x)$ and it approximates $f$ with error $\varepsilon = 2^{-2m-2}$; cf. (79). Thus, $\log_2(\varepsilon^{-1}) = 2m + 2 = 2(m+1) \Leftrightarrow m+1 = \mathcal{L}(\boldsymbol{\Phi}^{\text{sq}}_\varepsilon) = C \log_2(\varepsilon^{-1})$ for $C = 0.5$. Since $\varepsilon \to 0$ as $m \to \infty$, $\mathcal{L}(\boldsymbol{\Phi}^{\text{sq}}_\varepsilon), \mathcal{M}(\boldsymbol{\Phi}^{\text{sq}}_\varepsilon) \to \infty$ as $m \to \infty$. Therefore, there exists $\boldsymbol{\Phi}^{\text{sq}}_\varepsilon \in \mathcal{NN}^{1,1}_{\infty, \infty, \rho}$ – with the aforementioned network constraints – such that $\big\|\boldsymbol{\Phi}^{\text{sq}}_\varepsilon(x) - x^2\big\|_{L^\infty([0,1])} \leq \varepsilon$. ∎



\end{document}